\pdfoutput=1
\documentclass{article}
\usepackage{iclr2027_conference,times}
\iclrfinalcopy   

\usepackage[T1]{fontenc}
\usepackage{microtype}
\usepackage{amsmath,amssymb,amsthm}
\usepackage{booktabs}
\usepackage{array}
\usepackage{enumitem}
\usepackage{xcolor}
\usepackage{graphicx}
\usepackage{tikz}
\usepackage{pgfplots}
\pgfplotsset{compat=1.17}
\usepackage{algorithm}
\usepackage{algpseudocode}
\usepackage{xurl}
\usepackage{titletoc}
\usepackage{hyperref}

\usepackage{amsmath,amsfonts,bm}

\def\eqref#1{equation~\ref{#1}}

\def\1{\bm{1}}

\DeclareMathAlphabet{\mathsfit}{\encodingdefault}{\sfdefault}{m}{sl}
\SetMathAlphabet{\mathsfit}{bold}{\encodingdefault}{\sfdefault}{bx}{n}

\renewcommand{\eqref}[1]{Eq.~\textup{(\ref{#1})}}

\theoremstyle{plain}
\newtheorem{theorem}{Theorem}
\newtheorem{proposition}{Proposition}
\newtheorem{lemma}{Lemma}
\newtheorem{corollary}{Corollary}
\theoremstyle{definition}
\newtheorem{definition}{Definition}
\newtheorem{assumption}{Assumption}

\definecolor{linkblue}{HTML}{2C5F8A}
\hypersetup{colorlinks=true, linkcolor=linkblue, citecolor=linkblue,
  urlcolor=linkblue,
  pdftitle={Top-k Retrieval Leaves Memory Value Undetermined},
  pdfauthor={Arman Behnam}}

\title{\centering Top-$k$ Retrieval Leaves\\ Memory Value Undetermined}

\author{\parbox{\dimexpr\textwidth-2\tabcolsep\relax}{\centering
  Arman Behnam \\
  \normalfont Department of Computer Science \\
  Illinois Institute of Technology \\
  Chicago, IL, USA \\
  \texttt{abehnam@hawk.illinoistech.edu}}}

\begin{document}
\maketitle
\lhead{Preprint}

\begin{abstract}
A memory-augmented agent must decide which memories to keep, merge and retrieve, and its only evidence of their value is its retrieval log: the memories it retrieved in each episode and whether it succeeded. Existing methods assign every memory a value from this log, even when the log cannot separate one memory's contribution from another's. We treat the log as an experiment that nobody designed and ask which memory values it determines. Two counts answer this before any outcome is observed: how often each memory was retrieved, and how often each pair was retrieved together. From these counts, we prove that top-$k$ retrieval leaves every memory value undetermined, however many episodes are logged, and that a question with its own memories, asked once, determines at most one of
their values. On fourteen standard memory benchmarks, no memory value is determined in any of $55$ configurations. Both causes are choices about retrieval. We introduce LEDGER, which estimates each determined Shapley value in closed form with a confidence interval and reports every other value as undetermined. It merges memories with identical histories without losing a determined value, and it retrieves so that the next log determines more. On pooled HotpotQA, LEDGER raises the fraction of memories determined from $0.001$ to $0.797$, at a cost of $2.6$ accuracy points on HotpotQA and with a gain of $2.7$ on LoCoMo. Code: \url{https://anonymous.4open.science/r/ledger-review-AF71/}.
\end{abstract}
\section{Introduction}
\label{sec:intro}

Agents that operate over long horizons keep a store of memories and must decide which to retain, which to merge and which to retrieve for each query~\citep{episodicmemory2025, memsurvey2025}. Each of these decisions depends on how much every memory contributes to the agent's success. A deployed agent can learn this only from its own history: in each episode it retrieved some memories and then succeeded or failed. This paper asks \emph{what that history can tell us about the value of each memory, and when it can tell us nothing}.

Existing systems follow three approaches. The first scores memories with proxies such as recency, access frequency or similarity to the query~\citep{fademem2026, forgetting2026, gam2026}. The second learns a value signal from downstream reward~\citep{rmm2025, memrl2026, romerl2026}. The third computes the Shapley value~\citep{shapley1953}, which divides the agent's success fairly among the memories it used~\citep{memlens2026, clustershapley2025, maxshapley2025}. Three challenges stand in the way. First, \emph{a deployed agent cannot experiment on its memory}: Shapley methods need outcomes for sets of memories that were never retrieved together, and the agent cannot go back and retrieve them. Second, memories are retrieved in groups, so \emph{the history may not separate the contribution of one memory from that of another}. When it does not, a method still returns a number, but the number is set by the method and not by the data, and nothing in the output reveals this. Third, \emph{making the history more informative can lower accuracy}, because it means changing what the agent retrieves.

We treat the history as \emph{an experiment that nobody designed}. Each episode is a trial in which a set of memories was retrieved and an outcome was observed, and the theory of experimental design states which quantities a set of trials determines~\citep{rao1973linear, pukelsheim2006design}. This view answers each challenge. First, it needs no new trials, because it asks only what the recorded ones determine. Second, it decides, for every memory, whether the history determines its value before any value is estimated, so a method can decline to report a value that the data do not support. Third, it identifies which changes to retrieval make the history informative, so their cost in accuracy can be measured.

The analysis needs only two counts that any retrieval log contains: how often each memory was retrieved, and how often each pair was retrieved together. From these counts alone, before any outcome is used, it decides which memory values the history determines. Applied to standard memory benchmarks~\citep{hotpotqa2018, locomo2024, longmemeval2025}, it finds that none is determined. One common convention is enough to cause this: retrieving the same number of memories for every query. A second, giving each question its own memories, limits each question to as many determined values as it has episodes. Both conventions are choices about retrieval, so the remedy changes retrieval: share one store across queries, vary how many memories are retrieved, and fill part of each retrieval with the relevant memories retrieved least often. On pooled HotpotQA most values then become determined, fewer on other stores, while accuracy falls slightly on one corpus and rises on another. Our method, LEDGER, estimates the determined values with confidence intervals and reports the rest as undetermined.


\begin{figure}[h]
  \centering
\definecolor{inkc}{HTML}{1A1A1A}\definecolor{mutedc}{HTML}{6B6B6B}
\definecolor{gridc}{HTML}{C8C8C8}\definecolor{basec}{HTML}{9E9E9E}
\definecolor{memA}{HTML}{0072B2}\definecolor{memB}{HTML}{E69F00}
\definecolor{memC}{HTML}{CC79A7}\definecolor{memD}{HTML}{6A994E}
\definecolor{badc}{HTML}{C0392B}\definecolor{goodc}{HTML}{1B7F5B}
\begin{tikzpicture}[font=\scriptsize, x=1cm, y=1cm,
  card/.style={draw=#1, fill=#1!18, line width=0.6pt, rounded corners=1pt,
               minimum width=0.26cm, minimum height=0.30cm, inner sep=0pt,
               font=\tiny\bfseries, text=#1},
  flow/.style={-latex, mutedc, line width=0.7pt}]
\def\mc#1{\ifcase#1\or memA\or memB\or memC\or memD\fi}
\node[font=\footnotesize\bfseries] at (0.64, 1.30) {what each query retrieved};
\node[font=\footnotesize\bfseries] at (3.55, 1.30) {how often};
\node[font=\footnotesize\bfseries] at (7.10, 1.30) {how much each memory helps};
\node[anchor=west, font=\footnotesize] at (-1.25, 0.80)
  {(a) every query retrieves $2$ memories};
\foreach \j in {1,...,4} \node[card=\mc{\j}] at (0.32*\j-0.16, 0.25) {\j};
\node[anchor=east, font=\tiny, text=mutedc] at (-0.08,0.25) {memories $\rightarrow$};
\node[rotate=90, anchor=south, font=\tiny, text=mutedc] at (-0.12,-0.96)
  {$\leftarrow$ queries};
\foreach \r/\a/\b in {1/1/2, 2/1/3, 3/1/4, 4/2/3, 5/2/4, 6/3/4} {
  \fill[\mc{\a}] (0.32*\a-0.32, -0.32*\r) rectangle ++(0.32, 0.32);
  \fill[\mc{\b}] (0.32*\b-0.32, -0.32*\r) rectangle ++(0.32, 0.32);
}
\draw[gridc, line width=0.3pt] (0,-1.92) grid[step=0.32] (1.28,0)
  (0,-1.92) rectangle (1.28,0);
\draw[flow] (1.55,-0.96) -- (2.25,-0.96);
\foreach \i/\x/\y in {1/2.95/-0.36, 2/4.15/-0.36, 3/4.15/-1.56, 4/2.95/-1.56}
  \coordinate (A\i) at (\x,\y);
\foreach \u/\v/\p in {1/2/0.5, 1/3/0.27, 1/4/0.5, 2/3/0.5, 2/4/0.27, 3/4/0.5}
  \draw[mutedc, line width=0.8pt] (A\u) -- node[fill=white, inner sep=0.4pt,
    pos=\p, text=mutedc] {1} (A\v);
\foreach \i in {1,...,4}
  \node[circle, fill=\mc{\i}, text=white, font=\scriptsize\bfseries,
        inner sep=0pt, minimum size=0.44cm] at (A\i) {3};
\draw[flow] (4.65,-0.96) -- (5.30,-0.96);
\draw[mutedc, line width=0.4pt] (5.50,-1.92) -- (8.72,-1.92);
\foreach \x/\h/\hh/\c in {5.85/1.25/0.65/basec, 6.50/0.90/1.20/memA,
    7.12/0.60/0.90/memB, 7.74/0.75/1.05/memC, 8.36/0.30/0.60/memD} {
  \fill[\c!55] (\x-0.21,-1.92) rectangle ++(0.42,\h);
  \draw[\c!80!black, dashed, line width=0.9pt] (\x-0.21,-1.92) rectangle ++(0.42,\hh);
}
\foreach \x/\h in {6.50/0.90, 7.12/0.60, 7.74/0.75, 8.36/0.30}
  \draw[-latex, badc, line width=0.9pt] (\x,-1.92+\h+0.02) -- ++(0,0.26);
\draw[-latex, badc, line width=0.9pt] (5.85,-1.92+1.23) -- ++(0,-0.54);
\node[font=\tiny, text=mutedc] at (5.85,-2.10) {none};
\foreach \j/\x in {1/6.50, 2/7.12, 3/7.74, 4/8.36}
  \node[card=\mc{\j}] at (\x,-2.12) {\j};
\node[circle, draw=badc, line width=1pt, text=badc,
      font=\large\bfseries, minimum size=0.75cm] at (9.70,-0.85) {?};
\node[text=badc, font=\footnotesize\bfseries] at (9.70,-1.50) {0/4};
\begin{scope}[yshift=-3.45cm]
\node[anchor=west, font=\footnotesize] at (-1.25, 0.70)
  {(b) queries retrieve $1$, $2$ or $3$ memories};
\foreach \j in {1,...,4} \node[card=\mc{\j}] at (0.32*\j-0.16, 0.25) {\j};
\node[anchor=east, font=\tiny, text=mutedc] at (-0.08,0.25) {memories $\rightarrow$};
\node[rotate=90, anchor=south, font=\tiny, text=mutedc] at (-0.12,-0.96)
  {$\leftarrow$ queries};
\foreach \r/\a/\b in {1/1/2, 2/1/3, 3/1/4, 4/2/3, 5/2/4, 6/3/4} {
  \fill[\mc{\a}] (0.32*\a-0.32, -0.32*\r) rectangle ++(0.32, 0.32);
  \fill[\mc{\b}] (0.32*\b-0.32, -0.32*\r) rectangle ++(0.32, 0.32);
}
\fill[\mc{1}] (0,-2.24) rectangle ++(0.32,0.32);
\foreach \j in {1,2,3} \fill[\mc{\j}] (0.32*\j-0.32,-2.56) rectangle ++(0.32,0.32);
\draw[gridc, line width=0.3pt] (0,-2.56) grid[step=0.32] (1.28,0)
  (0,-2.56) rectangle (1.28,0);
\draw[inkc, line width=0.9pt, rounded corners=1pt]
  (-0.04,-2.60) rectangle (1.32,-1.88);
\node[font=\footnotesize\bfseries] at (-0.22,-2.24) {$+$};
\draw[flow] (1.55,-1.28) -- (2.25,-1.28);
\foreach \i/\x/\y in {1/2.95/-0.68, 2/4.15/-0.68, 3/4.15/-1.88, 4/2.95/-1.88}
  \coordinate (B\i) at (\x,\y);
\foreach \u/\v/\w/\lw/\p in {1/2/2/1.5/0.5, 1/3/2/1.5/0.27, 1/4/1/0.8/0.5,
                             2/3/2/1.5/0.5, 2/4/1/0.8/0.27, 3/4/1/0.8/0.5}
  \draw[mutedc, line width=\lw pt] (B\u) -- node[fill=white, inner sep=0.4pt,
    pos=\p, text=mutedc] {\w} (B\v);
\foreach \i/\f in {1/5, 2/4, 3/4, 4/3}
  \node[circle, fill=\mc{\i}, text=white, font=\scriptsize\bfseries,
        inner sep=0pt, minimum size=0.44cm] at (B\i) {\f};
\draw[flow] (4.65,-1.28) -- (5.30,-1.28);
\draw[mutedc, line width=0.4pt] (5.50,-2.24) -- (8.72,-2.24);
\foreach \x/\h/\c in {5.85/1.25/basec, 6.50/0.90/memA, 7.12/0.60/memB,
                      7.74/0.75/memC, 8.36/0.30/memD} {
  \fill[\c!55] (\x-0.21,-2.24) rectangle ++(0.42,\h);
  \draw[\c!80!black, line width=0.9pt] (\x-0.21,-2.24) rectangle ++(0.42,\h);
  \node[text=goodc] at (\x,-2.24+\h+0.16) {$\checkmark$};
}
\node[font=\tiny, text=mutedc] at (5.85,-2.42) {none};
\foreach \j/\x in {1/6.50, 2/7.12, 3/7.74, 4/8.36}
  \node[card=\mc{\j}] at (\x,-2.44) {\j};
\node[circle, draw=goodc, line width=1pt, text=goodc,
      font=\large, minimum size=0.75cm] at (9.70,-1.12) {$\checkmark$};
\node[text=goodc, font=\footnotesize\bfseries] at (9.70,-1.77) {4/4};
\end{scope}
\end{tikzpicture}
\caption{A history that looks sufficient and determines nothing. Rows are episodes, labelled by the memories they retrieve; the column $|S_t|$ gives each retrieval size.
  \emph{Left:} each pair of four memories retrieved once.
  \emph{Right:} the same episodes plus retrievals of sizes $1$ and $3$.}
  \label{fig:example}
\end{figure}
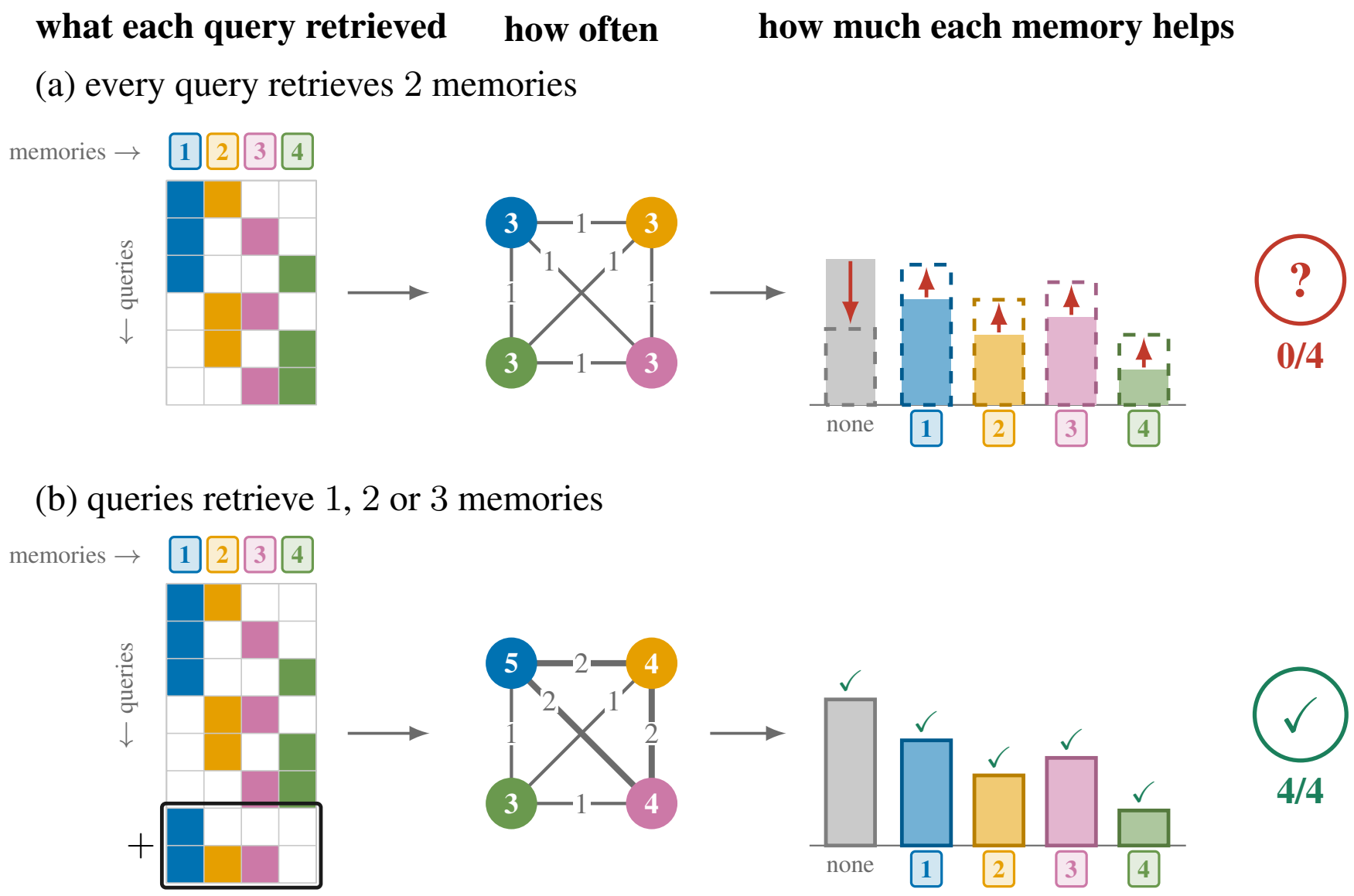

Figure~\ref{fig:example} shows the first convention on four memories. Each pair is retrieved once, giving six episodes for four unknown values, which looks sufficient. Yet every episode retrieves two memories, so adding the same amount to every value and removing twice that amount from the baseline leaves every predicted outcome unchanged: the history fixes only the differences between values. One episode of size one and one of size three determine all four. Appendix~\ref{app:discussion} discusses the scope, applications and limitations of these results, and Appendix~\ref{app:notation} collects the notation and terms.

This paper makes three contributions. First, it answers a question that prior work leaves open: which memory values a deployed agent's history determines. Proxy and learned methods report values without asking it, and Shapley methods avoid it by replaying sets of memories; we answer it exactly from the history, and show that a fixed retrieval size leaves every value undetermined however long the history grows (Section~\ref{sec:identifiability}). Second, it gives LEDGER, which computes each determined value in closed form with a confidence interval, reports every other value as undetermined, and merges memories with identical histories without losing any determined value (Section~\ref{sec:ledger}). Third, it shows that the conventions the field uses to evaluate memory leave every memory value undetermined, and that changing how memories are retrieved makes most values determined on pooled HotpotQA and fewer on other stores, at a measured cost in accuracy (Sections~\ref{sec:ledger} and~\ref{sec:experiments}). The same counts appear in slate off-policy evaluation, which sets the undetermined part aside because it does not affect the value of a policy~\citep{swaminathan2017slate}; for memory, the undetermined part is the finding.
\section{Background and Problem Formulation}
\label{sec:background}

\paragraph{Preliminaries.}We use three concepts from prior work. \emph{M\"obius dividends:} every set function $v$ on the subsets of $[n] = \{1, \dots, n\}$ can be written uniquely as $v(S) = \sum_{B \subseteq S}\theta_B$, where the dividend $\theta_B$ is the part of the value that the set $B$ creates jointly and none of its proper subsets creates~\citep{harsanyi1963, grabisch2000equivalent}. \emph{The Shapley value} of element $i$ is the unique attribution satisfying efficiency, symmetry, null player and additivity~\citep{shapley1953}; in dividends it is $\varphi_i = \sum_{B \ni i}\theta_B/|B|$, so each dividend is shared equally among its members~\citep{harsanyi1963}. \emph{Estimability:} in a linear model $\mathbf{r} = \mathbf{A}\boldsymbol{\beta} + \boldsymbol{\varepsilon}$, a combination $\mathbf{c}^{\!\top}\boldsymbol{\beta}$ is estimable when some linear function of $\mathbf{r}$ has expectation $\mathbf{c}^{\!\top}\boldsymbol{\beta}$ for every $\boldsymbol{\beta}$, which holds exactly when $\mathbf{c}$ lies in the row space of $\mathbf{A}$~\citep[Section~4a.2]{rao1973linear}; otherwise all least-squares solutions fit the data equally well and give $\mathbf{c}^{\!\top}\boldsymbol{\beta}$ every real value (Lemma~\ref{lem:minnorm}). Appendices~\ref{app:prelim} and~\ref{app:values} state these concepts formally.

\paragraph{Setting.}An agent holds a store of $n$ memories $m_1, \dots, m_n$, identified with $[n]$. In episode $t \in [T]$ it receives a query $q_t$ drawn from a distribution $\mathcal{D}_Q$, retrieves a non-empty set $S_t \subseteq [n]$, answers, and observes an outcome $r_t \in [0,1]$, for example whether the answer was correct. We make two assumptions; Appendix~\ref{app:assumptions} discusses what each requires and what is lost when it fails.

\begin{assumption}[Outcome model]
\label{ass:outcome}
For each query $q$ there is a set function $v_q : 2^{[n]} \to [0,1]$ such that $\mathbb{E}[r_t \mid \mathcal{F}_{t-1}, q_t, S_t] = v_{q_t}(S_t)$, where $\mathcal{F}_{t-1}$ is the history before episode $t$, and $\eta_t = r_t - v_{q_t}(S_t)$ is conditionally $\sigma$-sub Gaussian.
\end{assumption}

\begin{assumption}[Resolution $d$]
\label{ass:additive}
The query-averaged value $v(S) = \mathbb{E}_{q \sim \mathcal{D}_Q}[v_q(S)]$ has $\theta_B = 0$ whenever $|B| > d$.
\end{assumption}

The baseline $\theta_\emptyset = v(\emptyset)$ is the expected outcome with nothing retrieved, and the value of memory $i$ is its Shapley value $\varphi_i$ under $v$, which the baseline does not enter.

\paragraph{Input and output.}The method reads only what an agent already records: for each episode $t$, the retrieved set $S_t$ and the outcome $r_t$. We store the sets as a binary matrix $\mathbf{X} \in \{0,1\}^{T \times n}$, one row per episode and one column per memory, with $\mathbf{X}_{ti} = 1$ exactly when $m_i \in S_t$, and the outcomes as a vector $\mathbf{r}$. The product $\mathbf{X}^{\!\top}\mathbf{X}$ holds the two counts of the introduction,
\begin{equation}
\label{eq:identity}
  \bigl(\mathbf{X}^{\!\top}\mathbf{X}\bigr)_{ii} = \mathrm{freq}(m_i),
  \qquad
  \bigl(\mathbf{X}^{\!\top}\mathbf{X}\bigr)_{ij} = \mathrm{co}(m_i, m_j),
\end{equation}
how often $m_i$ was retrieved and how often $m_i$ and $m_j$ were retrieved together. By Assumption~\ref{ass:additive}, each outcome is the sum of the dividends of the retrieved sets of at most $d$ memories plus noise $\varepsilon_t = r_t - v(S_t)$. This gives the linear model derived in Appendix~\ref{app:design},
\begin{equation}
\label{eq:model}
  \mathbf{r} = \mathbf{X}_d\boldsymbol{\theta} + \boldsymbol{\varepsilon},
  \qquad
  (\mathbf{X}_d)_{tB} = \mathbf{1}[B \subseteq S_t],
  \quad |B| \le d,
\end{equation}
whose column for $B = \emptyset$ is all ones and carries the baseline; at $d = 1$, $\mathbf{X}_1 = [\mathbf{1}_T \mid \mathbf{X}]$. The output has one entry per memory: an estimate of $\varphi_i$ with a confidence interval when the log determines it, and \emph{undetermined} otherwise.

\paragraph{Problem.}Given this log, we first ask which memory values it determines.
Memory $i$ is \emph{identifiable} when the log determines $\varphi_i$: writing $\varphi_i = (\mathbf{c}^{(i)})^{\!\top}\boldsymbol{\theta}$ with $c^{(i)}_B = \mathbf{1}[i \in B]/|B|$, this holds exactly when $\mathbf{c}^{(i)}$ lies in the row space of $\mathbf{X}_d$ (Definition~\ref{def:identifiable}), and it is decided from $\mathbf{X}_d$ alone, before any outcome is seen. The determined values then guide how the agent manages its store: which memories to merge without losing a value, which to keep under a budget (Appendix~\ref{app:res-uses}), and how to retrieve so that the next log determines more, at a cost in accuracy that we measure.
\vspace{-3mm}
\section{What a Top-$k$ Log Determines}
\label{sec:identifiability}

The problem began with one decision for each memory: whether the log determines its value. This section makes that decision for the logs that agents actually produce. It first shows that whether $\mathbf{c}^{(i)}$ lies in the row space of $\mathbf{X}_d$ depends only on counts the agent already keeps, and then checks this condition under three common retrieval conventions: retrieving a fixed number of memories per query, giving each query its own memories, and retrieving some memories always together. For each convention, it identifies what the log cannot see and the change to retrieval that restores it. All proofs are in Appendix~\ref{app:proofs}.

\paragraph{What a log cannot see.} By Section~\ref{sec:background}, memory $i$ is identifiable exactly when $\mathbf{c}^{(i)}$ lies in the row space of $\mathbf{X}_d$, that is, when it is orthogonal to the kernel $\ker\mathbf{X}_d$. We call each $\mathbf{h} \in \ker\mathbf{X}_d$ an \emph{invisible change}: $\boldsymbol{\theta}$ and $\boldsymbol{\theta} + s\mathbf{h}$ predict the same expected outcome in every episode for every $s \in \mathbb{R}$, while $\varphi_i$ moves by $s\,(\mathbf{c}^{(i)})^{\!\top}\mathbf{h}$. Hence $\varphi_i$ is determined exactly when $(\mathbf{c}^{(i)})^{\!\top}\mathbf{h} = 0$ for every $\mathbf{h} \in \ker\mathbf{X}_d$. Since $\ker\mathbf{X}_d = \ker(\mathbf{X}_d^{\!\top}\mathbf{X}_d)$ (Appendix~\ref{app:proofs-counts}), the invisible changes depend only on $\mathbf{X}_d^{\!\top}\mathbf{X}_d$, which at $d = 1$ is
\begin{equation}
\label{eq:gram1}
  \mathbf{X}_1^{\!\top}\mathbf{X}_1 =
  \begin{bmatrix} T & \mathbf{f}^{\!\top} \\
                  \mathbf{f} & \mathbf{X}^{\!\top}\mathbf{X} \end{bmatrix},
  \qquad f_i = \mathrm{freq}(m_i).
\end{equation}
Whether each memory is identifiable is therefore decided by $T$, the retrieval frequencies and the co-retrieval counts alone, and at resolution $d$ by the number of episodes that retrieve each set of at most $2d$ memories (Lemma~\ref{lem:gram-d}). The retrieval convention that produces these counts therefore decides which memory values a log can determine, before any outcome is observed and whatever the memories contain.

\paragraph{A fixed retrieval size.}
\label{sec:fixedk}
The most common convention retrieves the $k$ memories most relevant to the query, so every episode retrieves the same number of memories; we call such a log a \emph{top-$k$ log}. In a top-$k$ log, an episode that retrieves a set $B$ of fewer than $k$ memories retrieves exactly $k - |B|$ other memories with it. In every episode, therefore, the columns of $\mathbf{X}_d$ for the sets $B \cup \{j\}$ with $j \notin B$ sum to $k - |B|$ times the column for $B$, which gives the invisible change
\begin{equation}
\label{eq:hB}
  \mathbf{h}_B \;=\; \sum_{j \notin B}\mathbf{e}_{B \cup \{j\}}
  \;-\; (k - |B|)\,\mathbf{e}_B \;\in\; \ker\mathbf{X}_d,
  \qquad |B| < d,
\end{equation}
where $\mathbf{e}_B$ is the unit vector for $\theta_B$. For $B = \emptyset$, $(\mathbf{c}^{(i)})^{\!\top}\mathbf{h}_\emptyset = 1$ for every memory $i$, so adding $\mathbf{h}_\emptyset$ changes every $\varphi_i$ without changing any predicted outcome.

\begin{theorem}[Top-$k$ leaves every value undetermined]
\label{thm:fixedk}
If every episode retrieves exactly $k$ memories, no memory is identifiable, at any resolution $d$ and any number of episodes $T$. If, in addition, outcomes are binary and each $v_q$ takes values in $(0,1)$, no estimator of any $\varphi_i$ is consistent.
\end{theorem}

At $d = 1$, the invisible change $\mathbf{h}_\emptyset$ raises every memory's value by $s$ and lowers the baseline by $ks$; this is the ambiguity of Figure~\ref{fig:example}. No estimator is consistent because adding a small multiple of $\mathbf{h}_\emptyset$ to the dividends of every $v_q$ leaves the distribution of the log unchanged, so two different values of $\varphi_i$ produce the same data (Appendix~\ref{app:proofs-fixedk}). More episodes do not help, because each new episode again retrieves exactly $k$ memories. The invisible changes disappear only if the retrieval size varies, and the next proposition gives how many sizes are needed.

\begin{proposition}[Sizes needed]
\label{prop:sizes}
If the retrieval sizes take at most $d$ distinct values, then $\ker\mathbf{X}_d$ contains a nonzero invisible change at every $T$.
\end{proposition}

At least $d + 1$ distinct sizes are therefore needed for the log to determine every dividend; at $d = 1$, two. With at most $d$ sizes, all below $n/2$, no memory is identifiable at all (Corollary~\ref{cor:sizes}). Under deterministic retrieval, where each query's sets are the top of one fixed ranking, the sets retrieved for a query at different sizes are nested, and nested sets can leave an invisible change of the same kind as $\mathbf{h}_\emptyset$, one that raises some memories' values and lowers the baseline (Lemma~\ref{lem:nesting}). Section~\ref{sec:ledger} removes both obstacles by varying the size and breaking the nesting, and Section~\ref{sec:exp-instrument} observes the nesting on real logs.

\paragraph{Private memories per query.} The second convention gives each query $b$ its own $n_b$ memories. The $T_b$ episodes of query $b$ then retrieve only from these memories, so $\mathbf{X}$ splits into one block per query, and the block of query $b$ has rank at most $T_b$.

\begin{proposition}[Private memories]
\label{prop:blocks}
If each query $b$ has its own $n_b$ memories and $T_b$ episodes, at least $\sum_b \max(n_b - T_b,\, 0)$ memories are not identifiable at $d = 1$.
\end{proposition}

A query asked once determines at most one of its $n_b$ memory values. Sharing one store across queries lifts this limit, but a fixed retrieval size leaves $\mathbf{h}_\emptyset$ in the log (Theorem~\ref{thm:fixedk}), so a shared store must be combined with varied retrieval sizes.

\paragraph{Memories always retrieved together.}
\label{sec:tied}
The third convention follows from ranking by relevance: memories that rank high for exactly the same queries are retrieved in the same episodes, and their columns of $\mathbf{X}$ are then identical. At $d = 1$, the invisible change $\mathbf{e}_{\{i\}} - \mathbf{e}_{\{j\}}$ raises $\varphi_i$ and lowers $\varphi_j$ by the same amount, leaving $\varphi_i + \varphi_j$ unchanged; this is aliasing~\citep{rosenbaum2024aliasing}. It accounts for the spurious credit that memory systems report for memories co-retrieved with useful ones~\citep{romerl2026, memoryr2_2026}. Because this change leaves the sum unchanged, Section~\ref{sec:ledger} merges such memories into one.

\paragraph{Retrieval decides what the log can determine.} Theorem~\ref{thm:fixedk} and Propositions~\ref{prop:sizes} and~\ref{prop:blocks} identify, from counts alone, which memory values a log leaves undetermined and why. Each undetermined value comes from an invisible change, and each invisible change comes from a retrieval convention (Table~\ref{tab:habits}): a fixed size gives $\mathbf{h}_B$, nested sets give a change like $\mathbf{h}_\emptyset$, private memories give changes inside each block, and tied memories give $\mathbf{e}_{\{i\}} - \mathbf{e}_{\{j\}}$. Every invisible change can be found from the log before any value is estimated, and a change to retrieval can remove it. Memory valuation thus becomes a question of design: which retrievals make the values an agent needs determinable.

\definecolor{memA}{HTML}{0072B2}\definecolor{memB}{HTML}{E69F00}
\definecolor{memC}{HTML}{CC79A7}\definecolor{memD}{HTML}{6A994E}
\definecolor{gridc}{HTML}{C8C8C8}
\newcommand{\logicon}[2]{
  \begin{tikzpicture}[baseline={([yshift=-0.6ex]current bounding box.center)}, x=0.17cm, y=0.17cm]
    \foreach \r/\c in {#2} {
      \fill[\ifcase\c\or memA\or memB\or memC\or memD\fi] (\c-1,-\r) rectangle ++(1,1);
    }
    \draw[gridc, line width=0.3pt] (0,-#1) grid (4,0) (0,-#1) rectangle (4,0);
  \end{tikzpicture}}
\begin{table}[h]
\centering
\footnotesize
\caption{Retrieval conventions and their invisible changes. Each icon is a small log: rows are episodes, columns are memories, and a filled cell marks a retrieval.}
\label{tab:habits}
\renewcommand{\arraystretch}{2.0}
\begin{tabular}{@{}c l l l l@{}}
\toprule
Log & Convention & Invisible change & Undetermined & Remedy (Section~\ref{sec:ledger}) \\
\midrule
\logicon{4}{1/1,1/2,2/1,2/3,3/2,3/4,4/3,4/4}
  & Fixed size $k$
  & $\mathbf{h}_B$, $|B| < d$
  & every value, at any $T$
  & $d + 1$ distinct sizes \\
\logicon{3}{1/1,2/1,2/2,3/1,3/2,3/3}
  & Nested top-$k$
  & like $\mathbf{h}_\emptyset$
  & the top-ranked memories' values
  & among equally relevant sets \\
\logicon{2}{1/1,1/2,2/3,2/4}
  & Private memories
  & inside each block
  & $\ge \sum_b \max(n_b - T_b, 0)$ values
  & share one store \\
\logicon{3}{1/1,1/2,1/3,2/2,2/3,2/4,3/2,3/3}
  & Tied memories
  & $\mathbf{e}_{\{i\}} - \mathbf{e}_{\{j\}}$
  & the split of $\varphi_i + \varphi_j$
  & merge, or retrieve apart \\
\bottomrule
\end{tabular}
\end{table}
\vspace{-3mm}
\section{LEDGER: Valuing Memory Where the Log Allows}
\label{sec:ledger}

\begin{figure}[h]
  \centering
\definecolor{inkc}{HTML}{1A1A1A}\definecolor{mutedc}{HTML}{6B6B6B}
\definecolor{gridc}{HTML}{C8C8C8}
\definecolor{memA}{HTML}{0072B2}\definecolor{memB}{HTML}{E69F00}
\definecolor{memC}{HTML}{CC79A7}\definecolor{memD}{HTML}{6A994E}
\definecolor{memE}{HTML}{7B4F9D}
\definecolor{badc}{HTML}{C0392B}\definecolor{goodc}{HTML}{1B7F5B}
\definecolor{loopc}{HTML}{2F6FAD}
\begin{tikzpicture}[font=\scriptsize, x=1cm, y=1cm,
  card/.style={draw=#1, fill=#1!18, line width=0.6pt, rounded corners=1pt,
               minimum width=0.28cm, minimum height=0.32cm, inner sep=0pt,
               font=\tiny\bfseries, text=#1},
  stage/.style={font=\footnotesize\bfseries, text=inkc},
  flow/.style={-latex, line width=0.9pt}]
\def\mc#1{\ifcase#1\or memA\or memB\or memC\or memD\or memE\fi}
\node[stage] at (0.8, 1.6) {the log};
\foreach \j in {1,...,5} \node[card=\mc{\j}] at (0.32*\j-0.16, 1.08) {\j};
\node[anchor=east, font=\tiny, text=mutedc] at (-0.06,1.08) {memories $\rightarrow$};
\node[rotate=90, anchor=south, font=\tiny, text=mutedc] at (-0.1,0)
  {$\leftarrow$ episodes};
\begin{scope}[shift={(0,0.8)}]
\foreach \r/\set in {1/{1,2,3}, 2/{2,3,4}, 3/{1,4}, 4/{2,3}, 5/{4}}
  \foreach \j in \set
    \fill[\mc{\j}] (0.32*\j-0.32, -0.32*\r) rectangle ++(0.32, 0.32);
\draw[gridc, line width=0.3pt] (0,-1.6) grid[step=0.32] (1.6,0)
  (0,-1.6) rectangle (1.6,0);
\end{scope}
\draw[flow, mutedc] (1.8,0) -- (2.55,0);
\node[stage] at (3.75, 1.6) {the test};
\node[draw=inkc, line width=0.8pt, rounded corners=3pt, align=center,
      minimum width=2.2cm, minimum height=1.3cm, fill=white] (test) at (3.75,0)
  {is $\varphi_i$ determined?\\[2pt]
   {\tiny\color{mutedc} does any invisible}\\[-1pt]
   {\tiny\color{mutedc} change move it?}};
\node[stage] at (8.3, 1.6) {what LEDGER does};
\draw[flow, goodc] (test.east) ++(0,0.25) -- (5.8,1.1);
\node[anchor=west, text=goodc, font=\scriptsize\bfseries] at (5.85,1.1) {yes};
\foreach \x/\h/\e/\c/\j in {6.75/0.42/0.10/memA/1, 7.15/0.30/0.07/memD/4} {
  \fill[\c!55] (\x-0.13,0.78) rectangle ++(0.26,\h);
  \draw[\c!70!black, line width=0.7pt] (\x,0.78+\h-\e) -- (\x,0.78+\h+\e)
    (\x-0.07,0.78+\h-\e) -- ++(0.14,0) (\x-0.07,0.78+\h+\e) -- ++(0.14,0);
  \node[card=\c] at (\x,0.6) {\j};
}
\node[anchor=west, align=left] at (7.5,1.1)
  {\textbf{estimate} $\hat{\varphi}_i$ with an interval};
\draw[flow, memB!80!black] (test.east) -- (5.8,0);
\node[anchor=west, text=memB!80!black, font=\scriptsize\bfseries] at (5.85,0)
  {tied};
\node[card=memB] at (6.65,0.16) {2};
\node[card=memC] at (6.65,-0.16) {3};
\draw[-latex, mutedc, line width=0.6pt] (6.84,0) -- (7.02,0);
\node[draw=inkc, line width=0.6pt, rounded corners=1pt, inner sep=1pt,
      minimum height=0.32cm, font=\tiny\bfseries,
      left color=memB!30, right color=memC!30] at (7.24,0) {2+3};
\node[anchor=west, align=left] at (7.5,0)
  {\textbf{merge}, then estimate the group value};
\draw[flow, badc] (test.east) ++(0,-0.25) -- (5.8,-1.0);
\node[anchor=west, text=badc, font=\scriptsize\bfseries] at (5.85,-1.0) {no};
\node[card=memE] at (6.75,-1.0) {5};
\node[circle, draw=badc, line width=0.8pt, text=badc, inner sep=0pt,
      minimum size=0.34cm, font=\scriptsize\bfseries] at (7.15,-1.0) {?};
\node[anchor=west, align=left] at (7.5,-1.0)
  {\textbf{report} \emph{undetermined}};
\draw[-latex, loopc, line width=1pt, rounded corners=6pt]
  (7.15,-1.18) -- (7.15,-1.75) -- (0.8,-1.75) -- (0.8,-0.95);
\node[fill=white, text=loopc, inner sep=1.5pt, font=\scriptsize]
  at (3.95,-1.75) {\textbf{retrieve differently}: vary $|S_t|$ and pick the
  most informative set that still answers};
\end{tikzpicture}
  \caption{LEDGER reads the kernel of the log from its counts and acts on
  each memory accordingly: it estimates the values the log determines,
  merges memories the log cannot separate, abstains on the rest.
  In the log shown, memories $2$ and $3$ are always retrieved together.}
  \label{fig:ledger}
\end{figure}
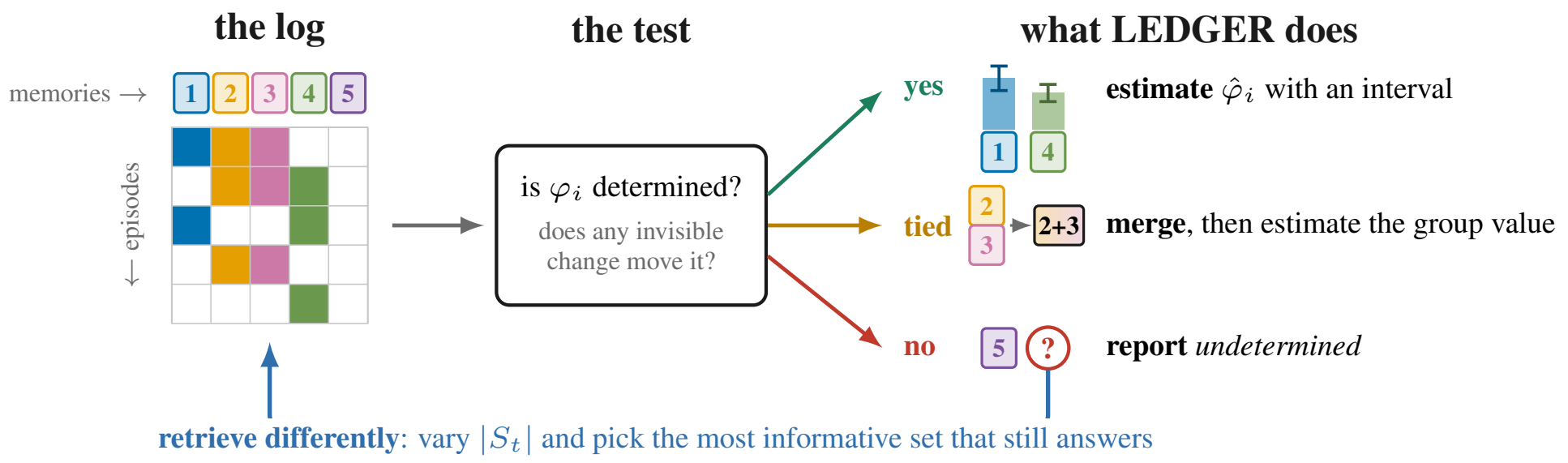

Section~\ref{sec:identifiability} showed that every undetermined value comes from an invisible change of the log, and that retrieval decides which invisible changes a log contains. This section presents LEDGER, a method built on one object, the invisible changes of the log: every value it reports is one that no invisible change moves, and each of its later actions removes invisible changes (Figure~\ref{fig:ledger}). With the log the agent already has, LEDGER first estimates each determined value with a confidence interval and reports every other value as undetermined. It then merges memories that are always retrieved together, which removes the invisible changes between them and keeps every determined value. For the logs the agent will collect, it removes the invisible changes that remain: it varies the retrieval size and fills part of each retrieval with the relevant memories retrieved least often. Appendix~\ref{app:algorithms} gives the three actions as algorithms.

\paragraph{Estimation.}
\label{sec:estimation}
LEDGER checks the condition of Section~\ref{sec:identifiability} numerically (Algorithm~\ref{alg:estimate}): with $\mathbf{U}$ an orthonormal basis of $\ker\mathbf{X}_d$, memory $i$ counts as determined when $\|\mathbf{U}^{\!\top}\mathbf{c}^{(i)}\| \le \tau\,\|\mathbf{c}^{(i)}\|$, with $\tau = 10^{-8}$. This check must come first, because a standard solver, such as least squares or ridge regression, returns a number for every memory, including undetermined ones (Lemma~\ref{lem:minnorm}). For an undetermined memory, that number estimates only the part of $\varphi_i$ that no invisible change moves. For each determined memory, LEDGER estimates $\varphi_i$ from a ridge fit with an interval,
\begin{equation}
\label{eq:estimate}
  \hat{\varphi}_i = (\mathbf{c}^{(i)})^{\!\top}\hat{\boldsymbol{\theta}},
  \quad
  \hat{\boldsymbol{\theta}} = (\mathbf{X}_d^{\!\top}\mathbf{X}_d
    + \lambda\mathbf{I})^{-1}\mathbf{X}_d^{\!\top}\mathbf{r},
  \qquad
  \hat{\varphi}_i \pm z_{1-\delta/(2n)}\,\hat{\sigma}\,
  \bigl\|\mathbf{X}_d(\mathbf{X}_d^{\!\top}\mathbf{X}_d
    + \lambda\mathbf{I})^{-1}\mathbf{c}^{(i)}\bigr\|,
\end{equation}
where $\lambda = 10^{-2}$ and $\hat{\sigma}$ is the residual standard deviation. The interval rests on the following:

\begin{proposition}[Intervals for determined values]
\label{prop:interval}
Suppose the noise $\varepsilon_t$ has mean zero and is $\sigma$-sub-Gaussian given the retrieved sets, and the retrieved sets do not depend on past outcomes. Then for each determined memory $i$, with probability at least $1 - \delta/n$,
\begin{equation}
\label{eq:interval}
  |\hat{\varphi}_i - \varphi_i|
  \;\le\; \sigma\sqrt{2\log(2n/\delta)}\,
  \bigl\|\mathbf{X}_d\mathbf{V}^{-1}\mathbf{c}^{(i)}\bigr\|
  \;+\; \lambda\,\bigl|(\mathbf{c}^{(i)})^{\!\top}
  \mathbf{V}^{-1}\boldsymbol{\theta}\bigr|,
  \qquad
  \mathbf{V} = \mathbf{X}_d^{\!\top}\mathbf{X}_d + \lambda\mathbf{I},
\end{equation}
so the bound holds for all $n$ memories at once with probability at least $1 - \delta$.
\end{proposition}

\vspace{-2mm}
The first term is the width of~\eqref{eq:estimate}, with the Gaussian quantile $z_{1-\delta/(2n)}$ in place of $\sqrt{2\log(2n/\delta)}$; the second is the ridge bias, which vanishes as $\lambda \to 0$ for determined values and does not for undetermined ones (Appendix~\ref{app:proofs-interval}). The retrieval rule of this section reads the past only through the retrieval counts, so it satisfies the second condition. Every other memory is reported as \emph{undetermined}, and only determined values guide which memories to keep. Memories that are always retrieved together are among the undetermined ones, and the next action handles them.
 
\paragraph{Merging.}
\label{sec:consolidation}
When memories $i$ and $j$ are always retrieved together, the invisible change $\mathbf{e}_{\{i\}} - \mathbf{e}_{\{j\}}$ raises $\varphi_i$ and lowers $\varphi_j$ by the same amount, so the log can determine at most the pair's total value. LEDGER treats the pair as one memory. It groups memories whose columns of $\mathbf{X}$ are identical, which hashing the columns finds in $O(nT)$ time (Algorithm~\ref{alg:merge}), and replaces each group $g$ by one memory whose column is the shared one and whose value is the sum $\bar{\theta}_g = \sum_{i \in g}\theta_{\{i\}}$; this gives a merged matrix $\bar{\mathbf{X}}$. Replacing memories changes the parameters, so the question is whether it changes what the log determines. Theorem~\ref{thm:quotient} shows that every determined value is kept.

\begin{theorem}[Merging keeps every determined value]
\label{thm:quotient}
At $d = 1$, $\mathrm{range}\,\mathbf{X}_1 = \mathrm{range}\,\bar{\mathbf{X}}_1$, so the two matrices give the same outcomes. A combination $\mathbf{c}^{\!\top}\boldsymbol{\theta}$ is determined by $\mathbf{X}_1$ exactly when $\mathbf{c}$ takes one value on each group and the corresponding combination of group values is determined by $\bar{\mathbf{X}}_1$.
\end{theorem}

\vspace{-2mm}
Every value determined before merging is determined after it, and every invisible change between tied memories is gone; the first action then estimates each group value that $\bar{\mathbf{X}}_1$ determines. The rule reads only retrieval histories: two unrelated memories with the same history are merged, and two near-duplicates retrieved in different episodes stay separate. Merging memories whose histories differ slightly is an approximation. For any grouping at $d = 1$, let a merged memory count as retrieved whenever one of its members is, let $D_g$ be the largest number of episodes in which group $g$ was retrieved while one of its members was not, let $s_g(\boldsymbol{\theta}) = \sum_{i \in g}\theta_{\{i\}}$, and let $\bar{\boldsymbol{\theta}} = (\theta_\emptyset, s_{g_1}(\boldsymbol{\theta}), \dots, s_{g_m}(\boldsymbol{\theta}))$.

\begin{theorem}[Merging memories with similar histories]
\label{thm:approx}
For every combination $\mathbf{c}$ in the row space of $\bar{\mathbf{X}}_1$, the value that the merged log determines for $\mathbf{c}^{\!\top}\bar{\boldsymbol{\theta}}$ differs from it by at most
\begin{equation}
\label{eq:approx}
  \frac{\|\mathbf{c}\|}{\sigma^{+}_{\min}(\bar{\mathbf{X}}_1)}
  \sum_{i=1}^{n} |\theta_{\{i\}}|\,\sqrt{D_{g(i)}},
\end{equation}
where $\sigma^{+}_{\min}$ is the smallest nonzero singular value and $g(i)$ is the group of memory $i$. The bound is zero when every $D_g = 0$, which is Theorem~\ref{thm:quotient}, and neither its dependence on $\sqrt{D_g}$ nor on $1/\sigma^{+}_{\min}$ can be improved.
\end{theorem}

Replacing $|\theta_{\{i\}}|$ by the upper end of its interval from~\eqref{eq:estimate} makes the bound computable.
 
\vspace{-3mm}
\paragraph{Retrieval.}
\label{sec:retrieval}
\label{sec:aopt}
The next log needs a store shared across queries, at least $d + 1$ distinct retrieval sizes (Proposition~\ref{prop:sizes}), and sets that are not nested (Table~\ref{tab:habits}). LEDGER keeps the store shared and draws the size $k_t$ of each retrieval from at least $d + 1$ values. Let $R_t$ be the $M$ memories most similar to $q_t$, and let $f_i(t)$ be the number of episodes before $t$ that retrieved memory $i$. LEDGER retrieves
\begin{equation}
\label{eq:rule}
  S_t = L_t \cup H_t,
\end{equation}
where $L_t$ holds the $\max(1, \mathrm{round}(\rho k_t))$ memories of $R_t$ with the smallest $f_i(t)$, and $H_t$ holds the $k_t - |L_t|$ memories of $R_t$ most similar to $q_t$ (Algorithm~\ref{alg:retrieve}). When $L_t$ and $H_t$ overlap, $|S_t| < k_t$. The memories in $H_t$ keep the answer within reach. The memories in $L_t$ target the least certain estimates: for small $\lambda$, the variance of $\hat{\theta}_{\{i\}}$ is at least $\sigma^2/(f_i(t) + \lambda)$, since $(\mathbf{V}^{-1})_{ii} \ge 1/V_{ii}$ for every positive definite $\mathbf{V}$ (Appendix~\ref{app:design-theory}). Because $f_i(t)$ changes from pass to pass, one query retrieves different sets over time, which breaks the nesting of top-$k$. The baseline test (Definition~\ref{def:intercept-test}) checks from the log whether a change like $\mathbf{h}_\emptyset$ remains. We use $\rho = 0.4$ and $M = 20$. Because this rule changes which memories the reader sees, it trades answer quality for determined values.
\vspace{-3mm}
\section{Experiments}
\label{sec:experiments}

Sections~\ref{sec:identifiability} and~\ref{sec:ledger} make four claims that only data can test. First, by Theorem~\ref{thm:fixedk} and Proposition~\ref{prop:blocks}, the logs of standard benchmarks leave every memory value undetermined; Section~\ref{sec:exp-protocol} checks this on fourteen benchmarks. Second, by Proposition~\ref{prop:sizes} and the retrieval rule~\eqref{eq:rule}, LEDGER's retrieval makes many more values
determined; Section~\ref{sec:exp-instrument} measures how many, and at what cost in accuracy. Third, by Proposition~\ref{prop:interval}, the interval in~\eqref{eq:estimate} contains each value LEDGER reports at the stated rate. Fourth, by Theorem~\ref{thm:quotient}, merging keeps every determined value and makes the total value of each group of tied memories determinable. Section~\ref{sec:exp-estimation} tests both.

\subsection{Setup}
\label{sec:setup}

The audit (Section~\ref{sec:exp-protocol} and Appendix~\ref{app:audit}) runs fourteen memory benchmarks, each with the retrieval it specifies. The retrieval experiments use three corpora with different structure: HotpotQA~\citep{hotpotqa2018} and LongMemEval~\citep{longmemeval2025}, whose questions we pool into one shared store per corpus, and LoCoMo~\citep{locomo2024}, whose questions already share a store within each conversation. Memories are turn- or paragraph-level spans, retrieved by dense similarity over sentence embeddings~\citep{reimers2019sbert}. Answers come from a hosted reader, \texttt{gemini-2.5-flash-lite}, and are scored by exact match, with the same model as judge otherwise; two further readers, \texttt{gemini-3.6-flash} and \texttt{gemini-3.7-flash}, measure the cost of pooling a store, and they and the use of determined values to decide which memories to keep appear in Appendix~\ref{app:results}. We compute every identifiability number from $\mathbf{X}$ alone, and compare retrieval rules on the same questions, reporting each accuracy difference with a $95\%$ interval and a paired permutation test. Coverage of the intervals is measured on synthetic logs (Appendix~\ref{app:setup-synthetic}). Table~\ref{tab:setup-runs} names the artifact behind each number below.

\subsection{What standard benchmarks determine}
\label{sec:exp-protocol}
 
For every memory, we checked whether $\mathbf{c}^{(i)}$ lies in the row space of $\mathbf{X}_1$, on the logs of fourteen memory benchmarks, each run with the retrieval it specifies and at every $k$ it uses (Table~\ref{tab:audit}). We ran fourteen benchmarks, each with its own retrieval at every $k$ it uses, and tested every memory at $d = 1$. Two conventions cause this. Within each block of the log, every episode retrieves the same number of memories (Theorem~\ref{thm:fixedk}). Most benchmarks also give each question or document its own store, which splits the log into many blocks (Proposition~\ref{prop:blocks}). A block is a group of memories never retrieved with a memory outside it, and even a shared store splits into blocks. Block counts follow the order of $k$, and ``all'' retrieves every memory of the context.

All $55$ configurations leave every memory value undetermined, for two reasons. Within each context, every episode retrieves the same number of memories, so a change like $\mathbf{h}_\emptyset$ is invisible in every log (Theorem~\ref{thm:fixedk}). Twelve of the fourteen benchmarks also give each question, document or conversation its own memories, which splits the log into blocks (Proposition~\ref{prop:blocks}). The two benchmarks with one shared store, MultiHopRAG and HELMET, leave every value undetermined because their retrieval size is fixed. LoCoMo shares a store within each conversation and splits its $1{,}542$ questions into $13$ blocks at $k = 5$, with the same result. A shared store is therefore necessary and not sufficient. Figure~\ref{fig:results}a compares LEDGER's retrieval rule~\eqref{eq:rule} with deterministic top-$k$ and with two random rules, $\varepsilon$-greedy and score perturbation, on pooled HotpotQA. The fraction of memories determined is measured on $300$ questions retrieved over $24$ passes ($n = 2{,}971$ memories, $T = 7{,}200$ episodes), and accuracy on $1{,}600$ held-out questions against a pooled store of $15{,}900$ memories.

\begin{table}[h]
\centering
\small
\caption{No memory benchmark determines a single memory value.}
\label{tab:audit}
\renewcommand{\arraystretch}{1.1}
\begin{tabular*}{\textwidth}{@{\extracolsep{\fill}}lllrrrr@{}}
\toprule
Benchmark & $k$ & Store & $T$ & $n$ & Blocks & Determined \\
\midrule
HotpotQA              & 3, 5, 10      & per question     & 300     & 2{,}971  & 300     & 0 \\
HotpotQA (1,200 q.)   & 3, 5, 10      & per question     & 1{,}200 & 11{,}912 & 1{,}200 & 0 \\
MuSiQue               & 3, 5, 10      & per question     & 300     & 5{,}999  & 300     & 0 \\
WikiHop               & 3, 5, 10      & per question     & 300     & 4{,}560  & 300     & 0 \\
TriviaQA              & 3, 5, 10      & per question     & 300     & 27{,}456 & 300     & 0 \\
LongMemEval           & 3, 5, 10      & per question     & 300     & 6{,}845  & 300     & 0 \\
\midrule
MSC-Self-Instruct     & 3, 5, 10      & per dialogue     & 300     & 18{,}916 & 300     & 0 \\
BABILong              & 3, 5, 10, all & per item         & 40      & 4{,}158  & 40      & 0 \\
InfiniteBench         & 3, 5, 10, all & per book         & 40      & 37{,}697 & 40      & 0 \\
NarrativeQA (summary) & 3, 5, 10, all & per document     & 300     & 115      & 11      & 0 \\
NarrativeQA (full)    & 3, 5, 10, all & per document     & 300     & 14{,}282 & 121, 60, 27, 11 & 0 \\
QuALITY               & 3, 5, 10, all & per article      & 219     & 1{,}495  & 47, 22, 12, 12  & 0 \\
LoCoMo                & 3, 5, 10      & per conversation & 1{,}542 & 5{,}882  & 37, 13, 11      & 0 \\
MemoryAgentBench      & 3, 5, 10      & per context      & 360     & 33{,}686 & 94, 65, 44      & 0 \\
\midrule
MultiHopRAG           & 3, 5, 6, 10   & shared           & 2{,}556 & 23{,}078 & 269, 98, 66, 15 & 0 \\
HELMET                & 3, 5, 10, 14  & shared           & 356     & 40{,}000 & 41      & 0 \\
\bottomrule
\end{tabular*}
\end{table}

Table~\ref{tab:audit} tests every memory on $55$ configurations of the 14 benchmarks, at $d = 1$ and therefore at every resolution (Lemma~\ref{lem:resolution}). The first reason is the retrieval size. In every log, all episodes of a context retrieve the same number of memories, $k$ or the whole context, so the log contains a change like $\mathbf{h}_\emptyset$ that reaches every memory (Theorem~\ref{thm:fixedk}, and Lemma~\ref{lem:audit} when the size differs between contexts). The second reason is the store. Twelve of the fourteen benchmarks give each question, document or conversation its own memories, which splits the log into blocks (Proposition~\ref{prop:blocks}). In HotpotQA, for example, each question has about ten memories of its own and is asked once, so varying the size would determine at most one of its ten values. A shared store also splits: MultiHopRAG's $2{,}556$ questions fall into $269$ blocks at $k = 3$ and $15$ at $k = 10$, and HELMET's $356$ into $41$. A shared store is necessary and not sufficient.

\vspace{-2mm}
\subsection{Retrieval and its cost}
\label{sec:exp-instrument}

\begin{figure}[h]
  \centering
\definecolor{inkc}{HTML}{1A1A1A}\definecolor{mutedc}{HTML}{8A8A8A}
\definecolor{ledgc}{HTML}{0072B2}\definecolor{randc}{HTML}{D55E00}
\definecolor{detc}{HTML}{5A5A5A}\definecolor{gatec}{HTML}{009E73}
\definecolor{goodc}{HTML}{EAF3FA}
\pgfplotsset{r3/.style={
  axis lines=left, axis line style={mutedc},
  tick label style={font=\scriptsize, text=inkc},
  label style={font=\scriptsize, text=inkc},
  title style={font=\scriptsize, text=inkc, align=center},
  every tick/.style={mutedc},
  extra y ticks={0}, extra y tick labels={},
  extra y tick style={grid=major, grid style={mutedc, densely dashed, line width=0.4pt}},
  clip=false}}
\begin{tikzpicture}
\begin{axis}[r3, name=pa, width=0.45\linewidth, height=5.2cm,
  xmin=0, xmax=1.05, ymin=-44, ymax=6,
  xtick={0,0.25,0.5,0.75,1}, xticklabels={0,.25,.5,.75,1},
  ytick={-40,-30,-20,-10,0},
  xlabel={fraction of memories determined},
  ylabel={accuracy change (points)},
  title={\textbf{(a)} Values determined and their cost}]
\draw[-stealth, mutedc, line width=0.6pt] (axis cs:0.30,-30) -- (axis cs:0.45,-18)
  node[pos=0.5, anchor=south east, font=\tiny, text=mutedc] {better};
\addplot[only marks, mark=triangle*, mark size=2.4pt, randc,
  error bars/.cd, y dir=both, y explicit, error bar style={line width=0.5pt}]
  coordinates {
    (0.001,-3.0)  -= (0,2.3) += (0,2.0)
    (0.989,-13.3) -= (0,4.0) += (0,3.6)
    (0.983,-14.0) -= (0,5.3) += (0,5.3)
    (1.000,-37.0) -= (0,6.0) += (0,6.3)};
\node[anchor=west, font=\tiny, text=randc] at (axis cs:0.03,-4.2) {$\varepsilon$-greedy, $\varepsilon{=}0.1$};
\node[anchor=east, font=\tiny, text=randc] at (axis cs:0.95,-11.6) {$\varepsilon$-greedy, $\varepsilon{=}0.3$};
\node[anchor=east, font=\tiny, text=randc] at (axis cs:0.95,-16.2) {perturbation, $0.05$};
\node[anchor=east, font=\tiny, text=randc] at (axis cs:0.95,-37.0) {perturbation, $0.10$};
\addplot[only marks, mark=square*, mark size=2.0pt, detc] coordinates {(0.001,0)};
\node[anchor=south west, font=\tiny, text=detc] at (axis cs:0.01,0.6) {top-$k$};
\addplot[only marks, mark=diamond*, mark size=2.8pt, gatec,
  error bars/.cd, y dir=both, y explicit, error bar style={line width=0.5pt}]
  coordinates {(0.509,0.25) +- (0,1.19)};
\node[anchor=south, font=\tiny, text=gatec] at (axis cs:0.509,1.7) {oracle (gold labels)};
\addplot[only marks, mark=*, mark size=2.6pt, ledgc,
  error bars/.cd, y dir=both, y explicit, error bar style={line width=0.7pt}]
  coordinates {(0.797,-2.56) -= (0,1.69) += (0,1.62)};
\node[anchor=north, font=\tiny\bfseries, text=ledgc] at (axis cs:0.797,-4.6) {LEDGER};
\end{axis}
\begin{axis}[r3, name=pb, at={(pa.east)}, anchor=west, xshift=0.55cm,
  width=0.29\linewidth, height=5.2cm,
  xmin=0.15, xmax=1.0, ymin=-6, ymax=6,
  xtick={0.25,0.5,0.75,1}, xticklabels={.25,.5,.75,1},
  ytick={-4,-2,0,2,4},
  xlabel={gold recall of top-$k$},
  title={\textbf{(b)} Cost across corpora}]
\addplot[domain=0.2:0.97, samples=2, mutedc, dotted, line width=0.6pt] {4.423-7.573*x};
\addplot[only marks, mark=*, mark size=2.4pt, ledgc,
  error bars/.cd, y dir=both, y explicit, error bar style={line width=0.7pt}]
  coordinates {
    (0.926,-2.56) -= (0,1.69) += (0,1.62)
    (0.461,0.84)  -= (0,2.72) += (0,2.92)
    (0.241,2.66)  -= (0,1.69) += (0,1.69)};
\node[anchor=east, font=\tiny, text=inkc] at (axis cs:0.90,-2.56) {HotpotQA};
\node[anchor=west, font=\tiny, text=inkc] at (axis cs:0.48,0.84) {LongMemEval};
\node[anchor=west, font=\tiny, text=inkc] at (axis cs:0.26,2.66) {LoCoMo};
\end{axis}
\begin{axis}[r3, name=pc, at={(pb.east)}, anchor=west, xshift=0.85cm,
  width=0.27\linewidth, height=5.2cm,
  ybar, bar width=0.45cm, enlarge x limits=0.45,
  ymin=-9.5, ymax=3, ytick={-6,-4,-2,0,2},
  symbolic x coords={all,missed}, xtick=data,
  xticklabels={{found\\all}, {missed\\some}},
  xticklabel style={font=\scriptsize, align=center},
  xlabel={top-$k$ on gold memories},
  title={\textbf{(c)} Where the cost arises}]
\addplot[fill=ledgc, draw=none] coordinates {(all,-6.9) (missed,0.7)};
\node[anchor=north, font=\tiny, text=inkc, align=center] at (axis cs:all,-7.0) {$0.242$ gold\\lost};
\node[anchor=south, font=\tiny, text=inkc, align=center] at (axis cs:missed,0.8) {$0.009$ gold\\lost};
\end{axis}
\end{tikzpicture}
  \caption{What LEDGER's retrieval determines, what it costs, and where the cost arises.}
  \label{fig:results}
\end{figure}
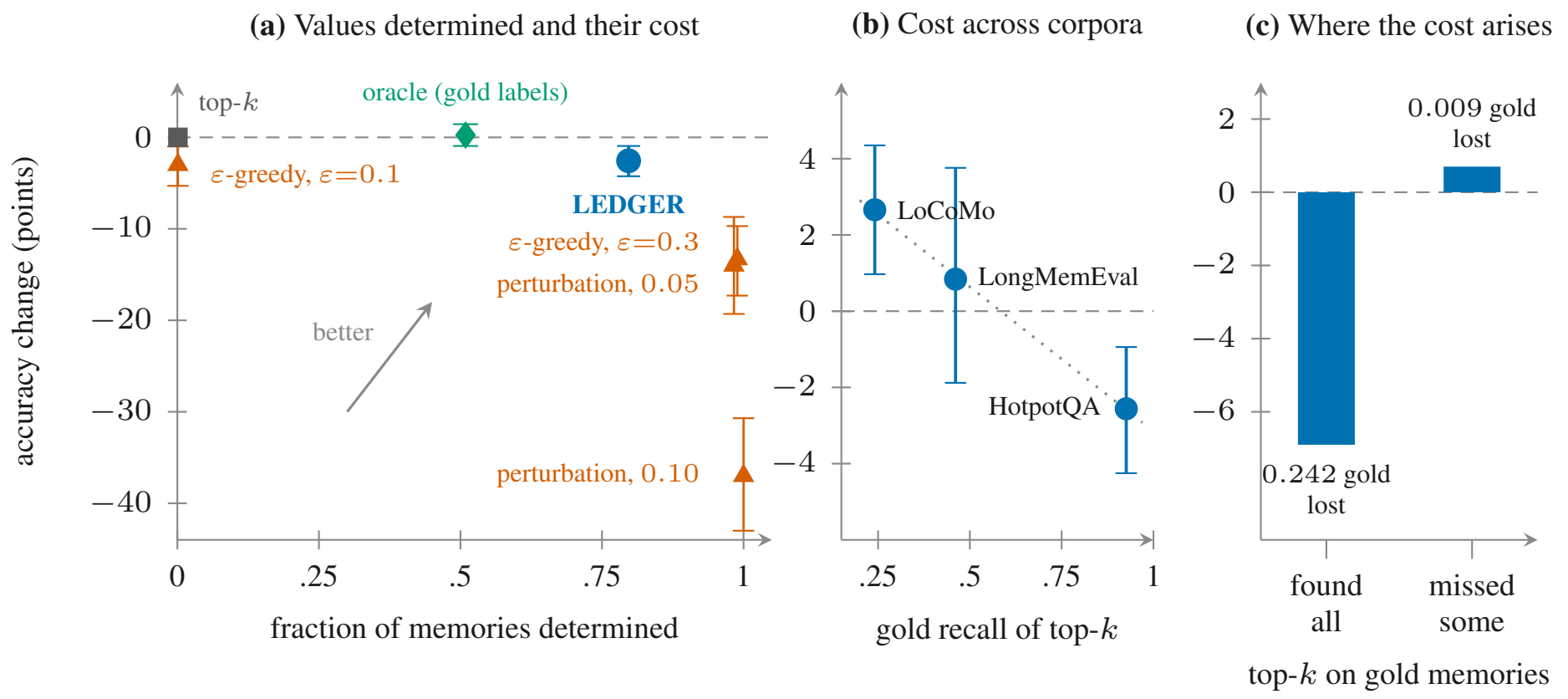

Hence, we vary the retrieval size on a shared store. On pooled HotpotQA, four rules draw the size from $\{3, 5, 8\}$ (Figure~\ref{fig:results}a): deterministic top-$k$, LEDGER's rule in~\eqref{eq:rule}, $\varepsilon$ greedy and score perturbation. LEDGER's sets fall below the drawn size in $6.7\%$ of episodes, when $L_t$ and $H_t$ overlap (Appendix~\ref{app:setup}). The fraction of memories determined is measured on $300$ questions retrieved over $24$ passes ($n = 2{,}971$ memories, $T = 7{,}200$
episodes). Accuracy is measured on $1{,}600$ held-out questions against a pooled store of $15{,}900$ memories. The accuracy of the random rules is measured on $300$ questions, with Holm corrected intervals. Deterministic top-$k$ determines $0.001$ of the memories, although its size varies. The reason is nesting: for each question, it retrieves the top of one ranking, so the set at each size contains the smaller ones. In this log, $\mathbf{X}\mathbf{w} = \mathbf{1}_T$ has a least-squares solution with residual $3.1 \times 10^{-15}$. Lowering the baseline by $s$ and raising each $\theta_{\{i\}}$ by
$s\,w_i$ then leaves every predicted outcome unchanged, an invisible change ($\mathbf{h}_\emptyset$). Pooling the store has its own cost, which this comparison does not include: on LongMemEval, answering from a pooled store costs $15.7$ points against each question's own store (Appendix~\ref{app:res-readers}).

LEDGER's rule fills part of each set with the relevant memories retrieved least often, and these change from pass to pass, which breaks the nesting. Its residual is $9.2 \times 10^{-2}$, so $\mathbf{1}_T$ lies outside the range of $\mathbf{X}$ and this change is gone. The fraction determined rises to $0.797$, at a cost of $2.6$ accuracy points ($95\%$ interval $[-4.2, -0.9]$, permutation $p = 0.004$). The random rules trade worse. At about LEDGER's cost, $\varepsilon$-greedy with $\varepsilon = 0.1$ loses $3.0$ points and determines $0.001$. To determine $0.98$ or more, the random rules lose at least $13.3$ points.

LEDGER's effect on accuracy depends on how much evidence top-$k$ already finds. We call a memory \emph{gold} when it holds evidence for the answer. On HotpotQA, LongMemEval and LoCoMo, top-$k$ retrieves $0.93$, $0.46$ and $0.24$ of the gold memories, and LEDGER changes accuracy by $-2.6$, $+0.8$ ($[-1.9, +3.8]$) and $+2.7$ points ($[+1.0, +4.4]$). Figure~\ref{fig:results}b shows the mechanism on HotpotQA. LEDGER replaces some of the similar memories with less-retrieved ones, which can displace gold memories that top-$k$ had found. On questions where top-$k$ retrieved every gold memory, LEDGER drops $0.242$ of them per question and loses $6.9$ points. On the other questions, it drops $0.009$ and gains $0.7$. An oracle that applies LEDGER only to the questions where top-$k$ missed a gold memory removes the cost ($[-0.94, +1.44]$). It determines $0.509$ of the memories, $64\%$ of LEDGER's gain.
 
\vspace{-2mm}
\subsection{Intervals and merging}
\label{sec:exp-estimation}
\label{sec:exp-consolidation}

Checking whether the intervals are valid requires the true values, so we use synthetic logs drawn from the model~\eqref{eq:model} with known $\boldsymbol{\theta}$, over $36$ settings of $n$, $k$, $T$ and $\lambda$ at each resolution, with $20$ random designs per setting (Appendix~\ref{app:setup}). A log counts as covered when the interval in~\eqref{eq:estimate}, with the noise scale known, contains $\varphi_i$ for every memory LEDGER reports, and at $\delta = 0.05$ at least $95\%$ of logs should be covered. At $d = 1$, every design has full rank, and at least $97.7\%$ of logs are covered in every setting. At $d = 2$, $24$ settings give rank-deficient designs, and at least $99.4\%$ of logs are covered in all $36$ settings, in a post hoc analysis (Appendix~\ref{app:corr-posthoc}). Counting every memory instead, coverage falls to $57.2\%$ in the worst setting, but every memory there is undetermined. The intervals are $1.52$ and $1.43$ times as wide as the pointwise Gaussian interval with the true $\sigma$ at $d = 1$ and $d = 2$. The confidence set of \citet{abbasi2011oful} is a median $6.6$ and $36$ times as wide at $\lambda = 1$, because it bounds every direction of $\boldsymbol{\theta}$, while LEDGER bounds only the $n$ directions $\mathbf{c}^{(i)}$ that define the memory values.

Merging replaces tied memories by groups. The value of each tied memory is undetermined, but the total value of its group can be determined, and by Theorem~\ref{thm:quotient} every other memory keeps its status. We therefore count units, meaning memories and groups. On pooled HotpotQA with $100$ questions and LEDGER's retrieval ($n = 992$ memories, $T = 2{,}400$), merging identical histories leaves $869$ units and raises the fraction determined from $0.815$ to $0.992$. Merging also helps on conversational memory. On two LoCoMo conversations ($n = 788$ memories), the standard log determines no value (Table~\ref{tab:audit}). Three passes of retrieval with score perturbation determine $79.0\%$ of their memories, and merging raises this to $95.7\%$ of the units.
\vspace{-3mm}
\section{Conclusion}
\label{sec:conclusion}

\vspace{-3mm}
A retrieval log determines a memory's value only if the retrievals
separate that memory from the others, and this can be checked from the
log before any outcome is seen.
Standard benchmarks fail this check: a fixed retrieval size and private
stores leave every memory value undetermined in all $55$ configurations
of fourteen benchmarks.
Both causes are choices about retrieval, so changing retrieval removes
them.
LEDGER raises the fraction of memories determined from $0.001$ to $0.797$
at a cost of $2.6$ accuracy points.
Every value LEDGER reports is determined by the log and comes with a
confidence interval, and every other value is reported as undetermined. Memory benchmarks should report which values their logs
determine alongside accuracy.
Finding at query time the questions where spreading retrieval costs
nothing is the open problem that remains.
\subsection*{AI use statement}

Generative AI tools helped design experiments
and choose their parameters, generate synthetic data, write code, and clean
datasets.
On the literature, they helped search, summarize, and analyze prior work
and format references.
On the paper, they helped create figures and tables, draft text, and edit it.
We reviewed all AI-assisted work and checked it in four ways.
Every empirical claim was pre-registered with a decision rule, and three
claims that failed their rule were withdrawn
(Appendix~\ref{app:corrections}).
Every reported number is reproduced from committed artifacts and was
reconciled against them in a separate audit.
Every theoretical claim was tested on synthetic designs with known ground
truth.
Every bibliography entry was verified against its source.
We take full responsibility for the content of this paper.

\subsection*{Ethics statement}

This work involves no human subjects, no new data collection, and no
release of new data, so no IRB review was required.
All experiments use public benchmarks under their published licenses;
their contents are encyclopedic text and synthetically generated
dialogue, and we neither collected nor inferred personal information from
them.
Two consequences of the method deserve explicit statement.
First, the instrumentation we recommend is not free to the end user: an
$A$-optimal tie-break deliberately selects among answer-preserving
retrieval sets by information gain, and we measure an accuracy cost whose
sign depends on the corpus, so a practitioner adopting it trades some
answer quality for the ability to value memory at all.
We report both signs, the mechanism that produces them
(Section~\ref{sec:aopt}), and a restriction that removes the cost on the corpus
where it is largest, so that this trade is made deliberately and not
inherited from a default.
Second, a framework that makes the value of an individual stored memory
identifiable thereby makes attribution to that memory identifiable.
The same quantity that justifies retaining or deleting a record supports
profiling the interactions the record was derived from.
We declare no conflicts of interest.
On research integrity, claims here are pre-registered, null and negative
results are reported alongside positive ones
(Appendix~\ref{app:results}), and corrections to
values that appeared in earlier drafts are recorded in the paper.

\subsection*{Reproducibility statement}

The two modeling assumptions are stated in Section~\ref{sec:background}
and their scope is discussed in Appendix~\ref{app:assumptions};
every theoretical claim in Sections~\ref{sec:identifiability} to~\ref{sec:consolidation} is proved in full in
Appendix~\ref{app:proofs}, with the merge-map construction and its
complexity given as explicit algorithms in
Appendix~\ref{app:algorithms}; the synthetic generator used to check
those claims against known ground truth is specified in
Appendix~\ref{app:setup}, and the tables reporting every headline
number are generated from committed artifacts and collected in
Appendix~\ref{app:results}.
Anonymous source code, the dated pre-registration documents, the exact
benchmark preparation and store-construction steps, and the artifacts
from which every table is generated are available at
\url{https://anonymous.4open.science/r/ledger-review-AF71/}.
The theoretical track (identifiability verdicts, ranks, the exact
quotient, and all synthetic experiments) is fully reproducible and
deterministic from the released code and a fixed seed, since it consults
no model.
The empirical track is reproducible in its design and not claimed to be
bit-reproducible in its outcomes: the retrieval logs, design matrices and
identifiability measurements regenerate byte-identically from the
released inputs, while the reader accuracies depend on hosted model
inference that we cannot pin, so we report paired comparisons with
intervals and the sample sizes they were computed at.
Appendix~\ref{app:corrections} documents the
defects we found in our own pipeline, the values they affected, and the
corrected values, so that a reader attempting to reproduce this work
knows which numbers changed and why.

\newpage
\bibliographystyle{iclr2027_conference}
\bibliography{references}

\clearpage
\part*{Appendix}
\appendix
\label{sec:appendix}
\counterwithin{theorem}{section}
\counterwithin{proposition}{section}
\counterwithin{lemma}{section}
\counterwithin{corollary}{section}
\counterwithin{definition}{section}

\startcontents[appendix]
\printcontents[appendix]{}{1}{
\setcounter{tocdepth}{2}}

\clearpage


\section{Preliminaries and Related Work}
\label{app:background}

This appendix collects the classical results the paper builds on and
fixes the quantities it reports.
Appendix~\ref{app:prelim} states when a linear model determines a linear
functional, Appendix~\ref{app:values} writes memory values as such
functionals, and Appendix~\ref{app:assumptions} discusses the scope of
the two assumptions.
Appendix~\ref{app:design} applies these results to a retrieval log,
Appendix~\ref{app:design-theory} gives the optimal-design and
confidence-set results that Sections~\ref{sec:ledger}
and~\ref{sec:experiments} use, and Appendix~\ref{app:related} relates
the work to the literature.
Notation and terms are collected in Appendix~\ref{app:notation}.

\subsection{Linear models and estimability}
\label{app:prelim}

Consider the linear model
$\mathbf{r} = \mathbf{A}\boldsymbol{\beta} + \boldsymbol{\varepsilon}$
with a known matrix $\mathbf{A} \in \mathbb{R}^{T \times p}$, an unknown
parameter $\boldsymbol{\beta} \in \mathbb{R}^p$ and
$\mathbb{E}\,\boldsymbol{\varepsilon} = \mathbf{0}$.
The question is which linear functionals
$\mathbf{c}^{\!\top}\boldsymbol{\beta}$ the data determine.

\begin{definition}[Estimable functional]
\label{def:estimable}
A linear functional $\mathbf{c}^{\!\top}\boldsymbol{\beta}$ is
\emph{estimable} if some $\mathbf{a} \in \mathbb{R}^T$ satisfies
$\mathbb{E}[\mathbf{a}^{\!\top}\mathbf{r}] =
\mathbf{c}^{\!\top}\boldsymbol{\beta}$ for every
$\boldsymbol{\beta} \in \mathbb{R}^p$.
\end{definition}

\begin{lemma}[\citealp{rao1973linear}, Section~4a.2]
\label{lem:estimable}
$\mathbf{c}^{\!\top}\boldsymbol{\beta}$ is estimable if and only if
$\mathbf{c} \in \mathrm{range}\,\mathbf{A}^{\!\top}$, equivalently
$\mathbf{c}^{\!\top}\mathbf{h} = 0$ for every
$\mathbf{h} \in \ker\mathbf{A}$.
If it is estimable, $\mathbf{c}^{\!\top}\hat{\boldsymbol{\beta}}$ takes
the same value for every least-squares solution
$\hat{\boldsymbol{\beta}}$, and when
$\mathrm{Cov}\,\boldsymbol{\varepsilon} = \sigma^2\mathbf{I}$ this value
is the best linear unbiased estimator of
$\mathbf{c}^{\!\top}\boldsymbol{\beta}$.
\end{lemma}

The first statement follows in one line:
$\mathbb{E}[\mathbf{a}^{\!\top}\mathbf{r}] =
(\mathbf{A}^{\!\top}\mathbf{a})^{\!\top}\boldsymbol{\beta}$ equals
$\mathbf{c}^{\!\top}\boldsymbol{\beta}$ for every $\boldsymbol{\beta}$
exactly when $\mathbf{A}^{\!\top}\mathbf{a} = \mathbf{c}$.
The equivalence with the kernel condition is
$\mathrm{range}\,\mathbf{A}^{\!\top} = (\ker\mathbf{A})^{\perp}$.
Two further facts are used throughout.
First, $\ker\mathbf{A} = \ker(\mathbf{A}^{\!\top}\mathbf{A})$: if
$\mathbf{A}^{\!\top}\mathbf{A}\mathbf{h} = \mathbf{0}$, then
$\|\mathbf{A}\mathbf{h}\|^2 = \mathbf{h}^{\!\top}\mathbf{A}^{\!\top}
\mathbf{A}\mathbf{h} = 0$, and the converse is immediate.
Estimability is therefore decided by the Gram matrix
$\mathbf{A}^{\!\top}\mathbf{A}$ alone.
Second, a vector $\mathbf{h} \in \ker\mathbf{A}$ is invisible in the
data: $\boldsymbol{\beta}$ and $\boldsymbol{\beta} + s\mathbf{h}$ give
the same distribution of $\mathbf{r}$ for every $s \in \mathbb{R}$,
while $\mathbf{c}^{\!\top}\boldsymbol{\beta}$ changes by
$s\,\mathbf{c}^{\!\top}\mathbf{h}$.
The next lemma states what a solver reports for a functional that is not
estimable.

\begin{lemma}[What a solver reports]
\label{lem:minnorm}
Write $\mathbf{c} = \mathbf{c}_1 + \mathbf{c}_2$ with
$\mathbf{c}_1 \in \mathrm{range}\,\mathbf{A}^{\!\top}$ and
$\mathbf{c}_2 \in \ker\mathbf{A}$.
The least-squares solutions are
$\mathbf{A}^{\dagger}\mathbf{r} + \ker\mathbf{A}$, where
$\mathbf{A}^{\dagger}$ is the Moore--Penrose pseudoinverse.
If $\mathbf{c}_2 \neq \mathbf{0}$, then
$\mathbf{c}^{\!\top}\hat{\boldsymbol{\beta}}$ takes every real value as
$\hat{\boldsymbol{\beta}}$ ranges over these solutions.
The minimum-norm solution and the ridge solution
$(\mathbf{A}^{\!\top}\mathbf{A} + \lambda\mathbf{I})^{-1}
\mathbf{A}^{\!\top}\mathbf{r}$, for every $\lambda > 0$, lie in
$\mathrm{range}\,\mathbf{A}^{\!\top}$, so each reports
$\mathbf{c}_1^{\!\top}\hat{\boldsymbol{\beta}}$: an estimate of
$\mathbf{c}_1^{\!\top}\boldsymbol{\beta}$, which differs from
$\mathbf{c}^{\!\top}\boldsymbol{\beta}$ by
$\mathbf{c}_2^{\!\top}\boldsymbol{\beta}$.
\end{lemma}

\begin{proof}
If $\hat{\boldsymbol{\beta}}$ is a least-squares solution, so is
$\hat{\boldsymbol{\beta}} + s\mathbf{c}_2$ for every $s$, and
$\mathbf{c}^{\!\top}(\hat{\boldsymbol{\beta}} + s\mathbf{c}_2) =
\mathbf{c}^{\!\top}\hat{\boldsymbol{\beta}} + s\|\mathbf{c}_2\|^2$
takes every real value.
The minimum-norm solution $\mathbf{A}^{\dagger}\mathbf{r}$ lies in
$\mathrm{range}\,\mathbf{A}^{\!\top}$ by definition of the pseudoinverse,
and the ridge solution equals
$\mathbf{A}^{\!\top}(\mathbf{A}\mathbf{A}^{\!\top} +
\lambda\mathbf{I})^{-1}\mathbf{r}$, which lies there as well.
For any $\hat{\boldsymbol{\beta}} \in \mathrm{range}\,\mathbf{A}^{\!\top}$,
$\mathbf{c}_2^{\!\top}\hat{\boldsymbol{\beta}} = 0$.
\end{proof}

The reported number and its standard error look the same whether or not
$\mathbf{c}_2 = \mathbf{0}$, so nothing in the output shows that the part
$\mathbf{c}_2^{\!\top}\boldsymbol{\beta}$ was set to zero by the solver.
LEDGER therefore tests $\mathbf{c}_2 = \mathbf{0}$ before it estimates
(Section~\ref{sec:estimation}).

\begin{lemma}[Appending a column]
\label{lem:append}
For $\mathbf{A} \in \mathbb{R}^{T \times p}$ and
$\mathbf{a} \in \mathbb{R}^T$,
$\mathrm{rank}\,[\mathbf{a} \mid \mathbf{A}] = \mathrm{rank}\,\mathbf{A}
+ \mathbf{1}[\mathbf{a} \notin \mathrm{range}\,\mathbf{A}]$.
Appending columns to $\mathbf{A}$ never makes a functional of the
original parameters estimable that was not estimable before.
\end{lemma}

\begin{proof}
The first statement is the definition of rank.
For the second, let $\mathbf{h} \in \ker\mathbf{A}$ with
$\mathbf{c}^{\!\top}\mathbf{h} \neq 0$.
Then $(\mathbf{h}, \mathbf{0}) \in \ker[\mathbf{A} \mid \mathbf{B}]$ for
any appended block $\mathbf{B}$, and its inner product with
$(\mathbf{c}, \mathbf{0})$ is $\mathbf{c}^{\!\top}\mathbf{h} \neq 0$, so
$(\mathbf{c}, \mathbf{0})$ is not estimable by
Lemma~\ref{lem:estimable}.
\end{proof}

\subsection{Set functions, dividends and values}
\label{app:values}

The results of Appendix~\ref{app:prelim} apply to memory values once
each value is a linear functional of a parameter vector.
M\"obius dividends provide that vector.

\begin{definition}[M\"obius dividends;
\citealp{harsanyi1963, grabisch2000equivalent}]
\label{def:mobius}
For $v : 2^{[n]} \to \mathbb{R}$, the M\"obius dividend of
$B \subseteq [n]$ is
$\theta_B = \sum_{A \subseteq B}(-1)^{|B \setminus A|}\,v(A)$.
Then $v(S) = \sum_{B \subseteq S}\theta_B$ for every $S \subseteq [n]$,
and the dividends are the unique numbers with this property.
We call $v$ \emph{$d$-additive} if $\theta_B = 0$ whenever $|B| > d$.
\end{definition}

The dividend $\theta_B$ is the part of $v$ that the members of $B$
create jointly and no proper subset of $B$ creates.
The dividend of the empty set is $\theta_\emptyset = v(\emptyset)$, the
baseline.
The Shapley value~\citep{shapley1953} of element $i$ is
\begin{equation*}
  \varphi_i(v) = \sum_{S \subseteq [n] \setminus \{i\}}
  \frac{|S|!\,(n - |S| - 1)!}{n!}
  \bigl[ v(S \cup \{i\}) - v(S) \bigr],
\end{equation*}
the average marginal contribution of $i$ over all orders in which the
elements can be added.

\begin{lemma}[\citealp{harsanyi1963}]
\label{lem:shapley}
$\varphi_i = \sum_{B \ni i}\theta_B/|B|$.
If $v$ is $d$-additive, then
$\varphi_i = (\mathbf{c}^{(i)})^{\!\top}\boldsymbol{\theta}$, where
$\boldsymbol{\theta} = (\theta_B)_{|B| \le d}$ and
$c^{(i)}_B = \mathbf{1}[i \in B]/|B|$.
The baseline $\theta_\emptyset$ does not enter $\varphi_i$.
\end{lemma}

Each dividend is thus shared equally among its members.
Other values are linear in the dividends with other coefficients.
The Banzhaf value~\citep{banzhaf1964weighted}, which averages the
marginal contribution of $i$ over all subsets with equal weight, is
$\sum_{B \ni i}2^{1-|B|}\theta_B$.
More generally, every probabilistic
value~\citep{weber1988probabilistic} averages the marginal contributions
$v(S \cup \{i\}) - v(S) = \sum_{A \subseteq S}\theta_{A \cup \{i\}}$ with
some weights, so it is linear in $\boldsymbol{\theta}$, and the
estimability test of Appendix~\ref{app:prelim} applies to it with its own
coefficient vector.
At $d = 1$ every marginal contribution equals $\theta_{\{i\}}$, so every
probabilistic value of $i$, the Shapley value included, equals
$\theta_{\{i\}}$.

\subsection{Assumptions and their scope}
\label{app:assumptions}

Section~\ref{sec:background} states two assumptions.
This subsection says what each requires, what is lost when it fails, and
the further condition that estimation needs.

\paragraph{Assumption~\ref{ass:outcome}.}
The assumption restricts neither the language model, the encoder nor the
query distribution.
It states that the expected outcome of an episode is a function of the
query and the retrieved set, which is what it means for retrieval to
matter, and that the deviations from this function are
conditionally sub-Gaussian, which holds for every bounded outcome.
The outcome may be binary or graded.
Decisions about the whole store, such as what to keep and what to merge,
need the query-averaged value
\begin{equation}
\label{eq:vbar}
  v(S) \;=\; \mathbb{E}_{q \sim \mathcal{D}_Q}\bigl[ v_q(S) \bigr],
\end{equation}
and the paper values memories under this $v$.

\paragraph{Assumption~\ref{ass:additive}.}
The Shapley value is defined for every $v$, and
Lemma~\ref{lem:shapley} writes it in dividends of every order.
The resolution $d$ keeps the dividends of order at most $d$, and the next
proposition gives exactly what the others contribute.

\begin{proposition}[Truncation error]
\label{prop:trunc}
For any $v$ and any $d$, let
$v^{(d)}(S) = \sum_{B \subseteq S,\, |B| \le d}\theta_B$.
Then
\begin{equation}
\label{eq:trunc}
  \varphi_i(v) - \varphi_i\bigl(v^{(d)}\bigr)
  \;=\; \sum_{B \ni i,\; |B| > d} \frac{\theta_B}{|B|},
  \qquad\text{so}\qquad
  \bigl| \varphi_i(v) - \varphi_i(v^{(d)}) \bigr| \le
  \frac{1}{d+1}\sum_{B \ni i,\, |B| > d} |\theta_B|.
\end{equation}
\end{proposition}

\begin{proof}
The dividends of $v^{(d)}$ are those of $v$ of order at most $d$ and zero
otherwise, by uniqueness in Definition~\ref{def:mobius}.
Subtracting the two expressions of Lemma~\ref{lem:shapley} gives the
identity, and $|B| \ge d + 1$ gives the bound.
\end{proof}

The proposition needs no assumption on $v$.
When Assumption~\ref{ass:additive} fails, the omitted dividends also
enter the noise $\varepsilon_t$, and the estimates at resolution $d$ are
biased wherever these dividends correlate with the design.
A memory undetermined at resolution $d$ remains undetermined at every
higher resolution (Appendix~\ref{app:interpret}), so the negative results
of Section~\ref{sec:experiments} hold whatever the true order of $v$.

\paragraph{Exogeneity.}
\label{sec:exogeneity}
Write $v_q = v + \delta_q$, so that $\mathbb{E}_q\,\delta_q(S) = 0$ for
every $S$.
The noise of~\eqref{eq:model} is then
$\varepsilon_t = \eta_t + \delta_{q_t}(S_t)$.
Retrieval reads $q_t$ when it chooses $S_t$, so $\delta_{q_t}(S_t)$ can
correlate with the design.

\begin{definition}[Exogeneity]
\label{def:exogeneity}
A log is \emph{exogenous} at resolution $d$ if
$\mathbb{E}\bigl[\mathbf{X}_d^{\!\top}\boldsymbol{\delta}\bigr] =
\mathbf{0}$, where $\boldsymbol{\delta}_t = \delta_{q_t}(S_t)$.
\end{definition}

A log is exogenous, for example, when $S_t$ is independent of $q_t$:
then $\mathbb{E}[\mathbf{1}[B \subseteq S_t]\,\delta_{q_t}(S_t)] =
\mathbb{E}_{S_t}\bigl[\mathbf{1}[B \subseteq S_t]\,
\mathbb{E}_q\,\delta_q(S_t)\bigr] = 0$.
Retrieval by relevance makes $S_t$ depend on $q_t$, so the logs of
Section~\ref{sec:experiments} need not be exogenous, and the values
LEDGER reports on them describe the association between retrieval and
outcome in the log.
Identifiability involves $\mathbf{X}_d$ alone, so it is unaffected;
only estimation is.
Appending query features $\psi(q_t)$ as columns of the design removes
the part of $\boldsymbol{\delta}$ in their span, and by
Lemma~\ref{lem:append} every result on identifiability in
Section~\ref{sec:identifiability} holds for the augmented design.
Estimation needs one further condition, on adaptivity:
Proposition~\ref{prop:interval} requires that the retrieved sets do not
depend on past outcomes.
LEDGER's rule~\eqref{eq:rule} depends on the past only through
$\mathbf{X}$, so it satisfies this condition.

\subsection{The design of a retrieval log}
\label{app:design}

\paragraph{The linear model.}
Under Assumptions~\ref{ass:outcome} and~\ref{ass:additive},
$r_t = v(S_t) + \varepsilon_t$ with $v$ $d$-additive.
By Definition~\ref{def:mobius},
\begin{equation*}
  v(S_t) = \sum_{|B| \le d}\theta_B\,\mathbf{1}[B \subseteq S_t],
  \qquad\text{so}\qquad
  \mathbf{r} = \mathbf{X}_d\boldsymbol{\theta} + \boldsymbol{\varepsilon},
\end{equation*}
which is~\eqref{eq:model}.
The design $\mathbf{X}_d$ has
$p_d = \sum_{\ell = 0}^{d}\binom{n}{\ell}$ columns, one for each set of
at most $d$ memories, and its column for $\emptyset$ is $\mathbf{1}_T$.
At $d = 1$, $\mathbf{X}_1 = [\mathbf{1}_T \mid \mathbf{X}]$, and its
Gram matrix is~\eqref{eq:gram1}: the number of episodes $T$, the
retrieval frequencies $\mathbf{f}$ and the co-retrieval counts
$\mathbf{X}^{\!\top}\mathbf{X}$ of~\eqref{eq:identity}.
By Appendix~\ref{app:prelim}, these counts decide every question of
estimability at $d = 1$.

\paragraph{Determined values.}
The paper uses the following definition throughout.

\begin{definition}[Determined value, fraction determined]
\label{def:identifiable}
At resolution $d$, memory $i$ is \emph{identifiable}, and its value is
\emph{determined}, if $\varphi_i$ is estimable in
$\mathbf{r} = \mathbf{X}_d\boldsymbol{\theta} + \boldsymbol{\varepsilon}$,
that is, if $\mathbf{c}^{(i)} \in \mathrm{range}\,\mathbf{X}_d^{\!\top}$.
Otherwise its value is \emph{undetermined}.
The fraction determined is the number of identifiable memories divided by
$n$; after merging, it is the number of determined units divided by the
number of units.
\end{definition}

At $d = 1$, $\mathbf{c}^{(i)} = (0, \mathbf{e}_i)$, so memory $i$ is
identifiable exactly when $(0, \mathbf{e}_i)$ is orthogonal to
$\ker\mathbf{X}_1$.
Numerically, LEDGER computes an orthonormal basis $\mathbf{U}$ of
$\ker\mathbf{X}_d$ and declares memory $i$ determined when
$\|\mathbf{U}^{\!\top}\mathbf{c}^{(i)}\| \le
\tau\|\mathbf{c}^{(i)}\|$ (Section~\ref{sec:estimation}).
Unless stated otherwise, every fraction in the paper is computed at
$d = 1$ with the baseline column included.

\paragraph{Blocks.}
The co-retrieval graph has one vertex per memory and an edge between
$m_i$ and $m_j$ whenever $\mathrm{co}(m_i, m_j) > 0$.

\begin{definition}[Blocks]
\label{def:blocks}
The blocks of a log are the connected components of the co-retrieval
graph restricted to the memories retrieved at least once.
Block $b$ contains $n_b$ memories and the $T_b$ episodes whose retrieved
sets lie in it.
\end{definition}

The members of one retrieved set are pairwise adjacent, so every
episode belongs to exactly one block.

\begin{lemma}[Block rank]
\label{lem:blocks}
After permuting rows and columns, the columns of $\mathbf{X}$ for the
retrieved memories form a block-diagonal matrix with diagonal blocks
$\mathbf{X}^{(b)} \in \{0,1\}^{T_b \times n_b}$, and the columns for
memories never retrieved are zero.
Hence $\mathrm{rank}\,\mathbf{X} = \sum_b \mathrm{rank}\,\mathbf{X}^{(b)}
\le \sum_b \min(T_b, n_b)$.
\end{lemma}

The baseline column $\mathbf{1}_T$ meets every block, so
$\mathbf{X}_1$ is not block-diagonal, and
$\mathrm{rank}\,\mathbf{X}_1 \le \mathrm{rank}\,\mathbf{X} + 1$ by
Lemma~\ref{lem:append}.
A memory that is never retrieved has a zero column, so $\mathbf{e}_i$ lies
in $\ker\mathbf{X}_1$ and its value is undetermined.

\paragraph{Tied memories.}
If memories $i$ and $j$ are retrieved in exactly the same episodes, their
columns of $\mathbf{X}$ are equal, and
$\mathbf{e}_{\{i\}} - \mathbf{e}_{\{j\}} \in \ker\mathbf{X}_1$.
Its inner product with $\mathbf{c}^{(i)}$ is $1$, so neither value is
determined, while the sum $\theta_{\{i\}} + \theta_{\{j\}}$ is
orthogonal to this change.
Merging (Section~\ref{sec:consolidation}) replaces each such group by
one unit valued by this sum.

\paragraph{The baseline test.}
Whether a log contains a change of the kind of $\mathbf{h}_\emptyset$ is
decided by one linear system.

\begin{definition}[Baseline test]
\label{def:intercept-test}
The baseline test solves $\mathbf{X}\mathbf{w} = \mathbf{1}_T$ by least
squares and reports the residual
$\|\mathbf{X}\hat{\mathbf{w}} - \mathbf{1}_T\|$.
\end{definition}

A zero residual means $\mathbf{1}_T \in \mathrm{range}\,\mathbf{X}$.
Then $(-1, \hat{\mathbf{w}}) \in \ker\mathbf{X}_1$: lowering the baseline
by $s$ and raising each $\theta_{\{i\}}$ by $s\hat{w}_i$ leaves every
predicted outcome unchanged, and every memory with $\hat{w}_i \neq 0$ is
undetermined.
A nonzero residual means that no change of this form exists, and by
Lemma~\ref{lem:append} it is equivalent to
$\mathrm{rank}\,\mathbf{X}_1 = \mathrm{rank}\,\mathbf{X} + 1$.
The test says something about retrieval only when
$\mathrm{rank}\,\mathbf{X} < T$: when $\mathrm{rank}\,\mathbf{X} = T$,
$\mathbf{1}_T$ lies in $\mathrm{range}\,\mathbf{X} = \mathbb{R}^T$
whatever the retrieval.
In a top-$k$ log, $\hat{\mathbf{w}} = \mathbf{1}_n/k$ solves the system
exactly, which gives $\mathbf{h}_\emptyset$; under deterministic top-$k$
at several sizes, nesting can make the residual zero as well, and
Section~\ref{sec:exp-instrument} measures it.

\paragraph{The example of Figure~\ref{fig:example}.}
In the left log, each of the $\binom{4}{2} = 6$ pairs of four memories is
retrieved once.
Every row of $\mathbf{X}$ has two ones, so $\mathbf{X}\mathbf{w} =
\mathbf{1}_6$ with $\mathbf{w} = \tfrac{1}{2}\mathbf{1}_4$, and
$\ker\mathbf{X}_1$ is spanned by
$\mathbf{h}_\emptyset = (-2, 1, 1, 1, 1)$, since
$\mathrm{rank}\,\mathbf{X}_1 = \mathrm{rank}\,\mathbf{X} = 4$ for five
parameters.
Each $(0, \mathbf{e}_i)$ has inner product $1$ with
$\mathbf{h}_\emptyset$, so no memory is identifiable, while each
difference $\theta_{\{i\}} - \theta_{\{j\}}$ is orthogonal to it and is
determined.
The right log adds one episode of size $1$ and one of size $3$.
Their rows break $\mathbf{X}\mathbf{w} = \mathbf{1}_8$: a solution would
need $w_i = \tfrac{1}{2}$ for every $i$ from the pairs, and then the
episode of size $1$ has row sum $\tfrac{1}{2} \neq 1$.
Hence $\mathrm{rank}\,\mathbf{X}_1 = 5$, the kernel is trivial, and all
four memories are identifiable.

\subsection{Optimal design and confidence sets}
\label{app:design-theory}

\paragraph{The variance of an estimate.}
With $\mathbf{V} = \mathbf{X}_d^{\!\top}\mathbf{X}_d + \lambda\mathbf{I}$
and noise covariance $\sigma^2\mathbf{I}$, the ridge estimate
of~\eqref{eq:estimate} satisfies
\begin{equation*}
  \mathrm{Var}\bigl(\mathbf{c}^{\!\top}\hat{\boldsymbol{\theta}}\bigr)
  = \sigma^2\,\mathbf{c}^{\!\top}\mathbf{V}^{-1}
    \mathbf{X}_d^{\!\top}\mathbf{X}_d\mathbf{V}^{-1}\mathbf{c}
  = \sigma^2\bigl(\mathbf{c}^{\!\top}\mathbf{V}^{-1}\mathbf{c}
    - \lambda\,\|\mathbf{V}^{-1}\mathbf{c}\|^2\bigr),
\end{equation*}
since $\mathbf{X}_d^{\!\top}\mathbf{X}_d = \mathbf{V} - \lambda\mathbf{I}$.
The first term dominates for small $\lambda$, so
$\mathrm{tr}(\mathbf{V}^{-1})$ is, up to $\sigma^2$, the total variance
of the coordinate estimates.
$A$-optimal design chooses the design to minimise
it~\citep{pukelsheim2006design}.
For invertible $\mathbf{V}$, the Sherman--Morrison identity
\begin{equation*}
  (\mathbf{V} + \mathbf{x}\mathbf{x}^{\!\top})^{-1}
  = \mathbf{V}^{-1}
  - \frac{\mathbf{V}^{-1}\mathbf{x}\mathbf{x}^{\!\top}\mathbf{V}^{-1}}
         {1 + \mathbf{x}^{\!\top}\mathbf{V}^{-1}\mathbf{x}}
\end{equation*}
shows that adding an episode with row $\mathbf{x}$ reduces
$\mathrm{tr}(\mathbf{V}^{-1})$ by
$\|\mathbf{V}^{-1}\mathbf{x}\|^2/(1 + \mathbf{x}^{\!\top}\mathbf{V}^{-1}
\mathbf{x})$, and updates $\mathbf{V}^{-1}$ in $O(p_d^2)$ time.

\paragraph{The bound behind LEDGER's retrieval rule.}
For every positive definite $\mathbf{V}$ and every $i$,
\begin{equation*}
  (\mathbf{V}^{-1})_{ii} \;\ge\; \frac{1}{V_{ii}},
\end{equation*}
by the Cauchy--Schwarz inequality:
$1 = (\mathbf{e}_i^{\!\top}\mathbf{V}^{1/2}\mathbf{V}^{-1/2}\mathbf{e}_i)^2
\le (\mathbf{e}_i^{\!\top}\mathbf{V}\mathbf{e}_i)
(\mathbf{e}_i^{\!\top}\mathbf{V}^{-1}\mathbf{e}_i)$.
At $d = 1$, the diagonal entry of $\mathbf{V}$ for memory $i$ is
$f_i + \lambda$, so the variance of $\hat{\theta}_{\{i\}}$ is, for small
$\lambda$, at least $\sigma^2/(f_i + \lambda)$.
Memories retrieved rarely therefore have the largest lower bound on
their variance, and LEDGER's rule~\eqref{eq:rule} retrieves them first
among the candidates $R_t$.
The rule is motivated by $A$-optimal design and optimises no criterion:
the bound is a lower bound, and the rule uses only the counts $f_i(t)$.

\paragraph{Confidence sets.}
For ridge regression with conditionally $\sigma$-sub-Gaussian noise, an
adaptively chosen design and $\|\boldsymbol{\theta}\| \le S$,
\citet{abbasi2011oful} show that, with probability at least
$1 - \delta$,
\begin{equation*}
  \|\hat{\boldsymbol{\theta}} - \boldsymbol{\theta}\|_{\mathbf{V}}
  \le \sigma\sqrt{2\log\!\Bigl(
      \frac{\det(\mathbf{V})^{1/2}\det(\lambda\mathbf{I})^{-1/2}}{\delta}
      \Bigr)}
  + \lambda^{1/2}S,
\end{equation*}
where $\|\mathbf{z}\|_{\mathbf{V}} =
(\mathbf{z}^{\!\top}\mathbf{V}\mathbf{z})^{1/2}$.
By the Cauchy--Schwarz inequality this bounds
$|\mathbf{c}^{\!\top}(\hat{\boldsymbol{\theta}} - \boldsymbol{\theta})|$
by $\|\mathbf{c}\|_{\mathbf{V}^{-1}}$ times the right-hand side, for
every $\mathbf{c}$ at once.
The set remains valid when the design depends on past outcomes, and its
log-determinant grows with $p_d$.
LEDGER's interval bounds only the $n$ functionals $\mathbf{c}^{(i)}$ and
requires a design that ignores past outcomes
(Proposition~\ref{prop:interval}); Section~\ref{sec:exp-estimation}
compares the two widths.

\subsection{Related work}
\label{app:related}

This work meets six lines of research: memory management by proxy
statistics, memory value learned from outcomes, cooperative-game
attribution, budgeted selection, graph-structured retrieval, and the
classical theory of estimation and design.
Each has produced strong systems, and this paper adopts much of their
machinery.
It asks a question none of them poses: which memory values a retrieval
log determines.

\paragraph{Memory management by proxy.}
Persistent memory is a primary determinant of long-horizon agent
performance~\citep{episodicmemory2025, memsurvey2025}, and a body of
work decides what to retain.
FadeMem~\citep{fademem2026} applies differential decay across a
two-layer hierarchy, modulating an exponential forgetting rate by
semantic relevance, access frequency and recency.
Budget-curated forgetting~\citep{forgetting2026} formulates retention as
bounded optimisation over similar statistics, and ACT-R-inspired
architectures~\citep{actrmemory2025} import the activation equation of a
model of human memory.
SSGM~\citep{ssgm2026} decouples memory evolution from execution with
consistency checks, temporal decay and access control, and
Aeon~\citep{aeon2026} targets the throughput of the same operations.
These systems differ in the statistic they use and share its status: it
is a function of content and access, fixed before any outcome is
observed, and validated by downstream accuracy.
Two memories with the same content and access pattern therefore receive
the same score whatever they contribute.
Access frequency is the diagonal of $\mathbf{X}^{\!\top}\mathbf{X}$
in~\eqref{eq:identity}, so these statistics record how often a memory is
used, which decides whether its contribution can be learned and does not
measure it (Appendix~\ref{app:applications}).

\paragraph{Memory value learned from outcomes.}
A second line learns a value signal from downstream reward.
RMM~\citep{rmm2025} refines retrieval online by reinforcement learning
from the evidence the model cites.
MemRL~\citep{memrl2026} applies reinforcement learning to episodic memory
updates, with a two-phase retrieval that filters for high-utility
entries.
AttriMem~\citep{attrimem2026} uses token-level attribution to the final
answer as a local reward for a memory-construction policy, and
Memory-R1~\citep{yan2026memory} and Memory-R2~\citep{memoryr2_2026}
train memory operations from outcome rewards, the latter with explicit
credit assignment over operations.
RoMeRL~\citep{romerl2026} names the memory-reward trap: irrelevant
memories accrue positive credit because they are retrieved together with
useful ones.
These methods learn a signal and show that it helps; none characterises
when the signal is recoverable.
The memory-reward trap is the invisible change
$\mathbf{e}_{\{i\}} - \mathbf{e}_{\{j\}}$ of tied memories
(Section~\ref{sec:tied}): when two memories share a retrieval history, no
procedure that reads the log can separate their credit, and the answer
consistent with the data is to report the pair as undetermined and merge
it.
When every rollout retrieves the same number of memories, the credit of
every memory is undetermined (Theorem~\ref{thm:fixedk}).
Both statements constrain every method in this line, LEDGER included.

\paragraph{Cooperative-game attribution.}
The Shapley value~\citep{shapley1953} is the canonical answer to what a
participant contributes to a joint outcome, and Data
Shapley~\citep{ghorbani2019datashapley} brought it to machine learning.
It has since been applied to retrieved content.
\citet{clustershapley2025} value source documents in summaries with a
clustering approximation and accuracy bounds.
MaxShapley~\citep{maxshapley2025} obtains exact values in polynomial
time from a decomposable max-sum utility.
Shapley context pruning~\citep{shapleyprune2026} and
RepoShapley~\citep{reposhapley2026} apply the same machinery to passage
reranking and to repository-level code context, and
MemLens~\citep{memlens2026} defines a Memory Shapley value and uses it
to drive retention.
General estimators approximate interactions of any order by sampling
coalitions~\citep{faithshap2023, shapiq2023, spex2025}, and context
attribution fits a surrogate to random ablations of the
prompt~\citep{cohen2024contextcite}.
We adopt the same notion of value.
The difference is access.
Each method above evaluates $v(S)$ on coalitions of its choosing, which
requires replaying the system on subsets that never occurred.
A deployed agent has only the coalitions that did occur, and LEDGER asks
which values those determine.
Because the Shapley value is a fixed linear functional of the M\"obius
dividends~\citep{harsanyi1963, grabisch2000equivalent}
(Lemma~\ref{lem:shapley}), the answer is exact and requires no sampling,
and the error of truncating to resolution $d$ is written in closed form
(Proposition~\ref{prop:trunc}).

\paragraph{Budgeted selection.}
Retention under a budget is a selection problem, and submodular
maximisation supplies its standard
guarantees~\citep{nemhauser1978, conforti1984curvature,
buchbinder2014submodular}.
PACMS~\citep{pacms2026} applies a facility-location objective with lazy
greedy selection to pooled memory entries, turns and tool outputs.
These guarantees hold for a proxy objective defined in embedding space,
and coverage in that space is assumed to track usefulness.
LEDGER's retention objective (Appendix~\ref{app:results}) is the
estimated value itself.
We do not assume that it is submodular, and we claim no guarantee that
would require it.

\paragraph{Graph-structured retrieval and coarsening.}
Retrieval-augmented generation~\citep{lewis2020rag} conditions generation
on a retrieved set, and graphs structure what is retrieved.
HippoRAG~\citep{hipporag2024} combines a knowledge graph with
Personalized PageRank, G-Retriever~\citep{gretriever2024} retrieves
subgraphs as a prize-collecting Steiner tree, and
RAGraph~\citep{ragraph2024} provides a general framework for
retrieval-augmented graph learning.
For multi-hop reasoning, RoG~\citep{rog2024} generates relation paths as
plans, ToG-2~\citep{tog2025} alternates knowledge-guided and
context-enhanced retrieval, StructRAG~\citep{structrag2025} structures
information at inference time, and HopRAG~\citep{hoprag2025} traverses
passages with logic-aware hops.
GAM~\citep{gam2026} and TierMem~\citep{tiermem2026} organise agent
memory hierarchically.
The graphs in these systems encode relations between contents.
The graph in this paper encodes usage: an edge joins two memories
retrieved in the same episode, and by~\eqref{eq:identity} its weighted
adjacency matrix is the Gram matrix of the valuation problem.
Merging is likewise distinct from graph
coarsening~\citep{graphcoarsening2024, topologycoarsening2024}, whose
guarantees state that a coarsened graph approximates the original within
a constant.
Merging memories with identical retrieval histories keeps every
determined value exactly (Theorem~\ref{thm:quotient}), because the only
quantities it removes are the splits of each group's total, which the
log never determined.

\paragraph{Estimation, design and bandits.}
Estimability of linear functionals under a rank-deficient design is
classical~\citep{rao1973linear}, and so is the aliasing of effects that
never vary apart~\citep{rosenbaum2024aliasing}.
Our contribution here is the identity~\eqref{eq:identity}: the design
matrix of memory valuation is determined by counts that memory systems
already record, so these results apply to them.
LEDGER's intervals are per-functional; the self-normalised confidence
set of \citet{abbasi2011oful} bounds every direction at once and remains
valid under adaptive designs, at a width we compare in
Section~\ref{sec:exp-estimation}.
Because only a set-level outcome is observed, the setting has the
full-bandit feedback of combinatorial and top-$k$
bandits~\citep{chen2013combinatorial, rejwan2020topk}, and this is why
the design, and not the reward model, limits what can be learned.
LEDGER's retrieval rule is motivated by optimal
design~\citep{pukelsheim2006design}, with one constraint that classical
design does not have: it may choose only among memories relevant to the
query, which is the source of the accuracy cost measured in
Section~\ref{sec:exp-instrument}.

\paragraph{Slate evaluation and intervention harvesting.}
Two neighbours are close enough to state the difference precisely.
Slate off-policy evaluation~\citep{swaminathan2017slate} builds the same
Gram matrix from logged set-valued actions.
Its estimand is the value of a policy, for which the kernel can be
projected out; ours is the value of each memory, and the kernel decides
which of those the log determines.
Recent work there reduces variance by learning an abstraction of the
slate space~\citep{lips2024}, which resembles merging; merging is forced
by Theorem~\ref{thm:quotient} and is exact.
Intervention harvesting~\citep{agarwal2019interventionharvesting}
recovers position bias from variation that occurs naturally across
deployed rankers.
LEDGER addresses the complementary case, in which that variation does not
occur, and measures what it costs to create it
(Section~\ref{sec:exp-instrument}).

\paragraph{Evaluation protocols.}
Benchmarks for long-horizon memory and multi-hop retrieval, including
LoCoMo~\citep{locomo2024}, LongMemEval~\citep{longmemeval2025} and
HotpotQA~\citep{hotpotqa2018}, measure answer accuracy under a fixed
retrieval protocol, and they measure it well.
Their protocols give each question its own memories or retrieve a fixed
number of items, and under either protocol the log determines no memory
value (Section~\ref{sec:exp-protocol}).
Appendix~\ref{app:audit} lists the fourteen benchmarks we audit and the
protocol of each.

\section{Proofs}
\label{app:proofs}

This appendix proves every result of Sections~\ref{sec:identifiability}
and~\ref{sec:ledger}, in the order in which they appear, together with
four supporting results: the form of the Gram matrix at every resolution
(Lemma~\ref{lem:gram-d}), a sharper form of Proposition~\ref{prop:sizes}
(Corollary~\ref{cor:sizes}), a sufficient condition for nested sets to
leave a change like $\mathbf{h}_\emptyset$ (Lemma~\ref{lem:nesting}), and
the monotonicity of identifiability in the resolution
(Lemma~\ref{lem:resolution}).
Throughout, $1 \le d \le n$, every retrieved set is non-empty, and
$\mathbf{e}_B$ denotes the unit vector of $\mathbb{R}^{p_d}$ for the
coordinate $\theta_B$.
We use the facts of Appendix~\ref{app:prelim}: $\varphi_i$ is determined
exactly when $(\mathbf{c}^{(i)})^{\!\top}\mathbf{h} = 0$ for every
$\mathbf{h} \in \ker\mathbf{X}_d$ (Lemma~\ref{lem:estimable}), and
$\ker\mathbf{X}_d = \ker(\mathbf{X}_d^{\!\top}\mathbf{X}_d)$.

\subsection{Counts decide identifiability}
\label{app:proofs-counts}

\begin{lemma}[Gram matrix at resolution $d$]
\label{lem:gram-d}
For all sets $B, B'$ of at most $d$ memories,
$(\mathbf{X}_d^{\!\top}\mathbf{X}_d)_{BB'} =
|\{t \in [T] : B \cup B' \subseteq S_t\}|$.
At $d = 1$, $\mathbf{X}_1^{\!\top}\mathbf{X}_1$ is the matrix
of~\eqref{eq:gram1}.
\end{lemma}

\begin{proof}
$(\mathbf{X}_d^{\!\top}\mathbf{X}_d)_{BB'} =
\sum_t \mathbf{1}[B \subseteq S_t]\,\mathbf{1}[B' \subseteq S_t] =
\sum_t \mathbf{1}[B \cup B' \subseteq S_t]$.
At $d = 1$, the pairs $(\emptyset, \emptyset)$, $(\emptyset, \{i\})$ and
$(\{i\}, \{j\})$ give $T$, $f_i$ and $\mathrm{co}(m_i, m_j)$, with
$\mathrm{co}(m_i, m_i) = f_i$.
\end{proof}

Since $\ker\mathbf{X}_d = \ker(\mathbf{X}_d^{\!\top}\mathbf{X}_d)$,
the set of determined values depends on the log only through these
counts: at $d = 1$ through $T$, the frequencies and the co-retrieval
counts, and at resolution $d$ through the number of episodes that
retrieve each set of at most $2d$ memories.

\subsection{A fixed retrieval size}
\label{app:proofs-fixedk}

\begin{proof}[Proof of the kernel identity~\eqref{eq:hB}]
Let $|B| < d$, so that every set $B \cup \{j\}$ indexes a column of
$\mathbf{X}_d$.
The entry of $\mathbf{X}_d\mathbf{h}_B$ for episode $t$ is
\begin{equation*}
  \sum_{j \notin B}\mathbf{1}[B \cup \{j\} \subseteq S_t]
  - (k - |B|)\,\mathbf{1}[B \subseteq S_t].
\end{equation*}
If $B \not\subseteq S_t$, both terms vanish.
If $B \subseteq S_t$, the sum counts the members of $S_t \setminus B$,
which number $|S_t| - |B| = k - |B|$, so the entry is $0$.
\end{proof}

\begin{proof}[Proof of Theorem~\ref{thm:fixedk}]
\emph{Identifiability.}
Take $B = \emptyset$, which satisfies $|B| < d$ because $d \ge 1$:
$\mathbf{h}_\emptyset = \sum_{j = 1}^{n}\mathbf{e}_{\{j\}} -
k\,\mathbf{e}_\emptyset \in \ker\mathbf{X}_d$.
Since $c^{(i)}_\emptyset = 0$ and $c^{(i)}_{\{j\}} = \mathbf{1}[j = i]$,
\begin{equation*}
  (\mathbf{c}^{(i)})^{\!\top}\mathbf{h}_\emptyset
  = \sum_{j=1}^{n} c^{(i)}_{\{j\}} = 1 \neq 0
\end{equation*}
for every memory $i$.
By Lemma~\ref{lem:estimable}, no $\varphi_i$ is estimable.
The argument uses only that every row of $\mathbf{X}$ has $k$ ones, so it
holds for every $d \ge 1$ and every $T$.

\emph{Consistency.}
Let outcomes be binary with
$\Pr(r_t = 1 \mid \mathcal{F}_{t-1}, q_t, S_t) = v_{q_t}(S_t)$, where
each $v_q$ takes values in $(0,1)$, and let the queries and the retrieval
rule be fixed.
The law of the log $(q_t, S_t, r_t)_{t \le T}$ depends on the functions
$v_q$ only through their values on sets of size $k$, since no other set
is ever retrieved.
Define $u(S) = |S| - k$ and, for $s \in \mathbb{R}$,
$v_q^{s} = v_q + s\,u$.
Because each $v_q$ takes finitely many values, all in $(0,1)$, there is
$\epsilon > 0$ with $\epsilon \le v_q(S) \le 1 - \epsilon$ for all $q$
and $S$; for $|s| < \epsilon/n$, every $v_q^{s}$ takes values in
$(0,1)$, so the perturbed model is valid.
Moreover $v_q^{s}(S) = v_q(S)$ whenever $|S| = k$, so the law of the log
is the same under $(v_q)$ and $(v_q^{s})$ for every $T$.
The dividends of $u$ are $-k$ on $\emptyset$, $1$ on each singleton and
$0$ otherwise, that is, $u$ has dividend vector $\mathbf{h}_\emptyset$.
Averaging over queries, $v^{s} = v + s\,u$ has dividends
$\boldsymbol{\theta} + s\,\mathbf{h}_\emptyset$, which is $d$-additive
whenever $\boldsymbol{\theta}$ is, and by
Lemma~\ref{lem:shapley} $\varphi_i(v^{s}) = \varphi_i(v) + s$.
Suppose an estimator $\hat{\varphi}_{i,T}$, a function of the first $T$
episodes, converged in probability to $\varphi_i$ under every valid
model.
Fix $0 < s < \epsilon/n$.
Under $(v_q)$ it converges to $\varphi_i(v)$, and under $(v_q^{s})$,
which gives the same law for every $T$, it converges to
$\varphi_i(v) + s$.
A sequence of random variables with a fixed law for each $T$ cannot
converge in probability to two different constants, a contradiction.
\end{proof}

\subsection{Sizes needed}
\label{app:proofs-sizes}

\begin{proof}[Proof of Proposition~\ref{prop:sizes}]
Let $K$ be the set of retrieval sizes, with $|K| \le d$.
For a polynomial $P$ of degree at most $d$, write
$P(x) = \sum_{\ell=0}^{d} a_\ell \binom{x}{\ell}$; the binomial
polynomials $\binom{x}{0}, \dots, \binom{x}{d}$ form a basis of the
polynomials of degree at most $d$, so the coefficients $a_\ell$ exist and
are unique.
Define $\mathbf{h}^{P} \in \mathbb{R}^{p_d}$ by
$h^{P}_B = a_{|B|}$.
For every episode $t$,
\begin{equation*}
  (\mathbf{X}_d\mathbf{h}^{P})_t
  = \sum_{B \subseteq S_t,\, |B| \le d} a_{|B|}
  = \sum_{\ell = 0}^{d}\binom{|S_t|}{\ell}a_\ell
  = P(|S_t|).
\end{equation*}
Take $P(x) = \prod_{k \in K}(x - k)$, a nonzero polynomial of degree
$|K| \le d$ that vanishes at every retrieval size.
Then $\mathbf{X}_d\mathbf{h}^{P} = \mathbf{0}$, and
$\mathbf{h}^{P} \neq \mathbf{0}$ because $P \neq 0$ and $d \le n$, so
every coefficient $a_\ell$ with $\ell \le d$ is carried by at least one
coordinate.
The construction does not depend on $T$.
\end{proof}

If every dividend were determined, $\ker\mathbf{X}_d$ would be trivial,
so the proposition implies that a log with at most $d$ distinct sizes
leaves some dividend undetermined.
The same construction shows more: under a mild condition on the sizes,
it leaves every memory value undetermined.

\begin{corollary}[Few sizes leave every value undetermined]
\label{cor:sizes}
If the retrieval sizes take at most $d$ distinct values and every size
is smaller than $n/2$, no memory is identifiable at resolution $d$.
\end{corollary}

\begin{proof}
For the vector $\mathbf{h}^{P}$ of the previous proof,
\begin{equation*}
  (\mathbf{c}^{(i)})^{\!\top}\mathbf{h}^{P}
  = \sum_{\ell=1}^{d}\binom{n-1}{\ell-1}\frac{a_\ell}{\ell}
  = \frac{1}{n}\sum_{\ell=1}^{d}\binom{n}{\ell}a_\ell
  = \frac{P(n) - P(0)}{n},
\end{equation*}
using $\binom{n-1}{\ell-1}/\ell = \binom{n}{\ell}/n$ and
$a_0 = P(0)$.
With $P(x) = \prod_{k \in K}(x - k)$ and $1 \le k < n/2$ for every
$k \in K$, we have $n - k > k$, so
$|P(n)| = \prod_{k}(n - k) > \prod_{k} k = |P(0)|$, hence
$P(n) \neq P(0)$, and the inner product is nonzero for every $i$.
\end{proof}

At $d = 1$ with one size $k$, $P(x) = x - k$ gives
$\mathbf{h}^{P} = \mathbf{h}_\emptyset$ and
$(P(n) - P(0))/n = 1$, recovering Theorem~\ref{thm:fixedk} without the
condition $k < n/2$.
The condition cannot be dropped in general: with $n = 6$, $d = 2$ and
every set of sizes $2$ and $4$ retrieved, $P(6) = P(0) = 8$, and every
memory is identifiable although $\ker\mathbf{X}_2 \neq \{\mathbf{0}\}$.
Conversely, $d + 1$ distinct sizes are necessary for every dividend to be
determined, and they are not sufficient: which values are determined
still depends on which sets are retrieved.

\subsection{Nested sets}
\label{app:proofs-nesting}

Under deterministic top-$k$, each query $q$ has a fixed ranking of the
memories, and every set retrieved for $q$ is a prefix of it.
Let $k_1$ be the smallest retrieval size, let $A_q$ be the $k_1$ memories
ranked highest for $q$, and let $C = \bigcup_q A_q$.

\begin{lemma}[Nesting leaves a change like $\mathbf{h}_\emptyset$]
\label{lem:nesting}
Suppose every set $S$ retrieved for any query $q$ satisfies
$S \cap C = A_q$.
Then $\mathbf{w} = \mathbf{1}_C/k_1$ solves
$\mathbf{X}\mathbf{w} = \mathbf{1}_T$, the vector
$(-1, \mathbf{w})$ lies in $\ker\mathbf{X}_1$, and no memory in $C$ is
identifiable at $d = 1$.
\end{lemma}

\begin{proof}
Every retrieved set contains $A_q$, because the sets for $q$ are
prefixes of one ranking of length at least $k_1$.
By hypothesis it contains no other member of $C$, so
$(\mathbf{X}\mathbf{w})_t = |S_t \cap C|/k_1 = |A_{q_t}|/k_1 = 1$.
Then $\mathbf{X}_1(-1, \mathbf{w}) = -\mathbf{1}_T +
\mathbf{X}\mathbf{w} = \mathbf{0}$, and for $i \in C$ the inner product
of $(0, \mathbf{e}_i)$ with $(-1, \mathbf{w})$ is $1/k_1 \neq 0$.
\end{proof}

The hypothesis holds whenever the queries have private stores, and in a
shared store it holds when no set retrieved for one query contains a
memory ranked in the top $k_1$ for another.
The lemma does not require a single retrieval size, so varying the size
does not remove this change.
The baseline test (Definition~\ref{def:intercept-test}) detects the
resulting change whether or not the hypothesis holds exactly: under
deterministic top-$k$ on pooled HotpotQA its residual is
$3.1 \times 10^{-15}$, zero up to rounding
(Section~\ref{sec:exp-instrument}).

\subsection{Private memories}
\label{app:proofs-blocks}

\begin{proof}[Proof of Proposition~\ref{prop:blocks}]
Order the memories by query and the episodes by query.
Because query $b$ retrieves only from its own $n_b$ memories,
$\mathbf{X}$ is block-diagonal with blocks
$\mathbf{X}^{(b)} \in \{0,1\}^{T_b \times n_b}$
(Lemma~\ref{lem:blocks}).
For any $\mathbf{g} \in \ker\mathbf{X}^{(b)}$, the vector that equals
$\mathbf{g}$ on the memories of $b$ and $0$ on the baseline and on every
other memory lies in $\ker\mathbf{X}_1$.
If memory $i$ of query $b$ is identifiable, $(0, \mathbf{e}_i)$ is
orthogonal to all these vectors, so $\mathbf{e}_i$, restricted to the
memories of $b$, lies in
$(\ker\mathbf{X}^{(b)})^{\perp} = \mathrm{range}\,\mathbf{X}^{(b)\top}$.
The identifiable memories of $b$ thus give linearly independent vectors in
a space of dimension $\mathrm{rank}\,\mathbf{X}^{(b)} \le
\min(T_b, n_b)$, so at most $T_b$ of them are identifiable and at least
$\max(n_b - T_b, 0)$ are not.
Summing over queries gives the bound.
\end{proof}

\subsection{Resolution}
\label{app:proofs-resolution}

\begin{lemma}[Undetermined values stay undetermined]
\label{lem:resolution}
If memory $i$ is not identifiable at resolution $d$, it is not
identifiable at any resolution $d' > d$.
\end{lemma}

\begin{proof}
It suffices to take $d' = d + 1 \le n$.
The columns of $\mathbf{X}_{d+1}$ are those of $\mathbf{X}_d$ followed by
the columns for sets of size $d + 1$.
Let $\mathbf{h} \in \ker\mathbf{X}_d$ with
$(\mathbf{c}^{(i)})^{\!\top}\mathbf{h} \neq 0$, and let
$\tilde{\mathbf{h}}$ extend $\mathbf{h}$ by zeros.
Then $\mathbf{X}_{d+1}\tilde{\mathbf{h}} = \mathbf{X}_d\mathbf{h} =
\mathbf{0}$.
The coefficient vector of $\varphi_i$ at resolution $d + 1$ agrees with
$\mathbf{c}^{(i)}$ on every set of size at most $d$, since both equal
$\mathbf{1}[i \in B]/|B|$, so its inner product with
$\tilde{\mathbf{h}}$ is $(\mathbf{c}^{(i)})^{\!\top}\mathbf{h} \neq 0$.
\end{proof}

\subsection{Intervals}
\label{app:proofs-interval}

We state the conditions of Proposition~\ref{prop:interval} formally.
Let $\mathcal{S} = \sigma(S_1, \dots, S_T)$.
The proposition assumes that for every $t$ and every $s \in \mathbb{R}$,
\begin{equation}
\label{eq:subg-cond}
  \mathbb{E}\bigl[\varepsilon_t \mid \mathcal{S}, \varepsilon_{1:t-1}\bigr]
  = 0,
  \qquad
  \mathbb{E}\bigl[e^{s\varepsilon_t} \mid \mathcal{S},
    \varepsilon_{1:t-1}\bigr] \le e^{s^2\sigma^2/2}.
\end{equation}
Conditioning on all sets, including later ones, is what requires the
sets not to depend on past outcomes: if $S_{t+1}$ depended on $r_t$, it
would carry information about $\varepsilon_t$ and~\eqref{eq:subg-cond}
would fail in general.
The mean-zero part of~\eqref{eq:subg-cond} also requires the
query-specific part $\delta_{q_t}(S_t)$ of the noise to average out given
the sets, which is the exogeneity of Appendix~\ref{app:assumptions}.

\begin{proof}[Proof of Proposition~\ref{prop:interval}]
With $\mathbf{r} = \mathbf{X}_d\boldsymbol{\theta} +
\boldsymbol{\varepsilon}$ and
$\mathbf{X}_d^{\!\top}\mathbf{X}_d = \mathbf{V} - \lambda\mathbf{I}$,
\begin{equation*}
  \hat{\boldsymbol{\theta}} - \boldsymbol{\theta}
  = \mathbf{V}^{-1}(\mathbf{V} - \lambda\mathbf{I})\boldsymbol{\theta}
    + \mathbf{V}^{-1}\mathbf{X}_d^{\!\top}\boldsymbol{\varepsilon}
    - \boldsymbol{\theta}
  = -\lambda\mathbf{V}^{-1}\boldsymbol{\theta}
    + \mathbf{V}^{-1}\mathbf{X}_d^{\!\top}\boldsymbol{\varepsilon}.
\end{equation*}
Fix $\mathbf{c} = \mathbf{c}^{(i)}$ and let
$\mathbf{a} = \mathbf{X}_d\mathbf{V}^{-1}\mathbf{c}$, so that
$\hat{\varphi}_i - \varphi_i =
-\lambda\,\mathbf{c}^{\!\top}\mathbf{V}^{-1}\boldsymbol{\theta}
+ \mathbf{a}^{\!\top}\boldsymbol{\varepsilon}$.
The vector $\mathbf{a}$ is a function of the sets, hence
$\mathcal{S}$-measurable.
Conditioning successively on $\mathcal{S}$ and
$\varepsilon_{1:T-1}, \dots, \varepsilon_{1:0}$ and
applying~\eqref{eq:subg-cond} at each step gives, for every
$s \in \mathbb{R}$,
\begin{equation*}
  \mathbb{E}\bigl[e^{s\,\mathbf{a}^{\!\top}\boldsymbol{\varepsilon}}
    \mid \mathcal{S}\bigr]
  \le \prod_{t=1}^{T} e^{s^2 a_t^2\sigma^2/2}
  = e^{s^2\sigma^2\|\mathbf{a}\|^2/2}.
\end{equation*}
The Chernoff bound then gives, with
$u = \sigma\|\mathbf{a}\|\sqrt{2\log(2n/\delta)}$,
\begin{equation*}
  \Pr\bigl(|\mathbf{a}^{\!\top}\boldsymbol{\varepsilon}| > u
    \mid \mathcal{S}\bigr)
  \le 2\exp\!\Bigl(-\frac{u^2}{2\sigma^2\|\mathbf{a}\|^2}\Bigr)
  = \frac{\delta}{n}.
\end{equation*}
On the complementary event,
$|\hat{\varphi}_i - \varphi_i| \le u +
\lambda|\mathbf{c}^{\!\top}\mathbf{V}^{-1}\boldsymbol{\theta}|$,
which is~\eqref{eq:interval}.
A union bound over the $n$ memories gives the simultaneous statement with
probability at least $1 - \delta$.
\end{proof}

The proof does not use that memory $i$ is determined; the bias term
shows why the proposition is stated for determined memories.
Write $\mathbf{c} = \mathbf{c}_1 + \mathbf{c}_2$ with
$\mathbf{c}_1 \in \mathrm{range}\,\mathbf{X}_d^{\!\top}$ and
$\mathbf{c}_2 \in \ker\mathbf{X}_d$.
Since $\mathbf{V}\mathbf{c}_2 = \lambda\mathbf{c}_2$,
$\lambda\,\mathbf{c}^{\!\top}\mathbf{V}^{-1}\boldsymbol{\theta} =
\lambda\,\mathbf{c}_1^{\!\top}\mathbf{V}^{-1}\boldsymbol{\theta} +
\mathbf{c}_2^{\!\top}\boldsymbol{\theta}$.
For a determined memory, $\mathbf{c}_2 = \mathbf{0}$, and because
$\mathbf{V}^{-1}\mathbf{c}_1$ lies in the row space, where the
eigenvalues of $\mathbf{V}^{-1}$ are at most
$1/(\sigma^{+}_{\min}(\mathbf{X}_d)^2 + \lambda)$,
\begin{equation*}
  \lambda\,|\mathbf{c}^{\!\top}\mathbf{V}^{-1}\boldsymbol{\theta}|
  \le \frac{\lambda\,\|\mathbf{c}\|\,\|\boldsymbol{\theta}\|}
           {\sigma^{+}_{\min}(\mathbf{X}_d)^2 + \lambda}
  \;\longrightarrow\; 0
  \quad (\lambda \to 0).
\end{equation*}
For an undetermined memory, the bias contains
$\mathbf{c}_2^{\!\top}\boldsymbol{\theta}$, which does not shrink with
$\lambda$ or with $T$; this is the undercoverage observed on
undetermined values in Section~\ref{sec:exp-estimation}.
The interval~\eqref{eq:estimate} replaces $\sigma$ by $\hat{\sigma}$ and
$\sqrt{2\log(2n/\delta)}$ by the Gaussian quantile $z_{1-\delta/(2n)}$,
which is smaller, and omits the bias term; its coverage is therefore
measured empirically (Section~\ref{sec:exp-estimation}).

\subsection{Merging}
\label{app:proofs-merging}

Let the groups $g_1, \dots, g_m$ partition $[n]$, and let
$\mathbf{G} \in \{0,1\}^{n \times m}$ have $G_{ig} = \mathbf{1}[i \in g]$.
Write $\bar{\mathbf{G}} = \mathrm{diag}(1, \mathbf{G}) \in
\{0,1\}^{(1+n) \times (1+m)}$ for its extension by the baseline
coordinate.
Its columns have disjoint non-empty supports, so $\bar{\mathbf{G}}$ has
full column rank and $\bar{\mathbf{G}}^{\!\top}$ is surjective.
For $\boldsymbol{\theta} = (\theta_\emptyset, \theta_{\{1\}}, \dots,
\theta_{\{n\}})$, the vector
$\bar{\boldsymbol{\theta}} = \bar{\mathbf{G}}^{\!\top}\boldsymbol{\theta}
= (\theta_\emptyset, s_{g_1}(\boldsymbol{\theta}), \dots,
s_{g_m}(\boldsymbol{\theta}))$ collects the baseline and the group values;
Theorem~\ref{thm:approx} writes $\mathbf{s}(\boldsymbol{\theta})$ for it.

\begin{proof}[Proof of Theorem~\ref{thm:quotient}]
When each group consists of memories with identical columns, the column
of $\mathbf{X}$ for memory $i$ equals the column of $\bar{\mathbf{X}}$
for its group, so $\mathbf{X} = \bar{\mathbf{X}}\mathbf{G}^{\!\top}$ and
$\mathbf{X}_1 = \bar{\mathbf{X}}_1\bar{\mathbf{G}}^{\!\top}$.
Because $\bar{\mathbf{G}}^{\!\top}$ is surjective,
$\mathrm{range}\,\mathbf{X}_1 = \bar{\mathbf{X}}_1\,
\mathrm{range}\,\bar{\mathbf{G}}^{\!\top} =
\mathrm{range}\,\bar{\mathbf{X}}_1$, and the least-squares fitted
outcomes, the projection of $\mathbf{r}$ onto this common range, agree.
For the row spaces,
$\mathrm{range}\,\mathbf{X}_1^{\!\top} = \bar{\mathbf{G}}\,
\mathrm{range}\,\bar{\mathbf{X}}_1^{\!\top}$.
Hence $\mathbf{c} \in \mathrm{range}\,\mathbf{X}_1^{\!\top}$ exactly when
$\mathbf{c} = \bar{\mathbf{G}}\bar{\mathbf{c}}$ for some
$\bar{\mathbf{c}} \in \mathrm{range}\,\bar{\mathbf{X}}_1^{\!\top}$.
The condition $\mathbf{c} = \bar{\mathbf{G}}\bar{\mathbf{c}}$ says that
$c_{\{i\}} = \bar{c}_{g(i)}$ for every $i$ and $c_\emptyset =
\bar{c}_\emptyset$, that is, $\mathbf{c}$ takes one value on each group;
$\bar{\mathbf{c}}$ is then unique because $\bar{\mathbf{G}}$ has full
column rank.
In that case
$\mathbf{c}^{\!\top}\boldsymbol{\theta} =
\bar{\mathbf{c}}^{\!\top}\bar{\mathbf{G}}^{\!\top}\boldsymbol{\theta} =
\bar{\mathbf{c}}^{\!\top}\bar{\boldsymbol{\theta}}$, the corresponding
combination of group values.
\end{proof}

In particular, for a memory $i$ in a group with at least two members,
$(0, \mathbf{e}_i)$ does not take one value on that group, so $\varphi_i$
is undetermined, while the group value $s_{g(i)}(\boldsymbol{\theta})$ is
determined exactly when $\bar{\mathbf{X}}_1$ determines it.

For an arbitrary grouping, the merged column of group $g$ is
$\bar{\mathbf{x}}_g = \max_{i \in g}\mathbf{x}_i$ entrywise, where
$\mathbf{x}_i$ is the column of $\mathbf{X}$ for memory $i$.
Let $\boldsymbol{\mu} = \mathbf{X}_1\boldsymbol{\theta}$ be the vector of
expected outcomes.
The value that the merged log determines for a combination
$\bar{\mathbf{c}}^{\!\top}\bar{\boldsymbol{\theta}}$ with
$\bar{\mathbf{c}} \in \mathrm{range}\,\bar{\mathbf{X}}_1^{\!\top}$ is
$\bar{\mathbf{c}}^{\!\top}\bar{\mathbf{X}}_1^{\dagger}\boldsymbol{\mu}$,
the common value of $\bar{\mathbf{c}}^{\!\top}\boldsymbol{\beta}$ over
all least-squares solutions $\boldsymbol{\beta}$ of
$\bar{\mathbf{X}}_1\boldsymbol{\beta} \approx \boldsymbol{\mu}$
(Lemma~\ref{lem:estimable}); it is the limit of the merged estimate as
the noise vanishes.

\begin{proof}[Proof of Theorem~\ref{thm:approx}]
\emph{The bound.}
Since $\bar{\mathbf{X}}_1\bar{\boldsymbol{\theta}} =
\theta_\emptyset\mathbf{1}_T + \sum_g s_g(\boldsymbol{\theta})
\bar{\mathbf{x}}_g = \theta_\emptyset\mathbf{1}_T + \sum_i
\theta_{\{i\}}\bar{\mathbf{x}}_{g(i)}$,
\begin{equation*}
  \boldsymbol{\mu} = \bar{\mathbf{X}}_1\bar{\boldsymbol{\theta}}
  + \mathbf{e},
  \qquad
  \mathbf{e} = \sum_{i=1}^{n}\theta_{\{i\}}
  \bigl(\mathbf{x}_i - \bar{\mathbf{x}}_{g(i)}\bigr).
\end{equation*}
For $\bar{\mathbf{c}} \in \mathrm{range}\,\bar{\mathbf{X}}_1^{\!\top}$,
$\bar{\mathbf{c}}^{\!\top}\bar{\mathbf{X}}_1^{\dagger}\bar{\mathbf{X}}_1
= \bar{\mathbf{c}}^{\!\top}$, because
$\bar{\mathbf{X}}_1^{\dagger}\bar{\mathbf{X}}_1$ is the orthogonal
projection onto the row space.
Hence
\begin{equation*}
  \bar{\mathbf{c}}^{\!\top}\bar{\mathbf{X}}_1^{\dagger}\boldsymbol{\mu}
  - \bar{\mathbf{c}}^{\!\top}\bar{\boldsymbol{\theta}}
  = \bar{\mathbf{c}}^{\!\top}\bar{\mathbf{X}}_1^{\dagger}\mathbf{e},
  \qquad
  \bigl|\bar{\mathbf{c}}^{\!\top}\bar{\mathbf{X}}_1^{\dagger}\mathbf{e}\bigr|
  \le \|\bar{\mathbf{c}}\|\,\|\bar{\mathbf{X}}_1^{\dagger}\|\,
    \|\mathbf{e}\|
  = \frac{\|\bar{\mathbf{c}}\|\,\|\mathbf{e}\|}
         {\sigma^{+}_{\min}(\bar{\mathbf{X}}_1)}.
\end{equation*}
Each vector $\mathbf{x}_i - \bar{\mathbf{x}}_{g(i)}$ has entries in
$\{-1, 0\}$, with $-1$ exactly in the episodes where group $g(i)$ was
retrieved and memory $i$ was not; there are at most $D_{g(i)}$ of them, so
$\|\mathbf{x}_i - \bar{\mathbf{x}}_{g(i)}\| \le \sqrt{D_{g(i)}}$.
The triangle inequality gives
$\|\mathbf{e}\| \le \sum_i |\theta_{\{i\}}|\sqrt{D_{g(i)}}$, which
proves~\eqref{eq:approx}.
When every $D_g = 0$, $\mathbf{e} = \mathbf{0}$ and the merged log
determines every $\bar{\mathbf{c}}^{\!\top}\bar{\boldsymbol{\theta}}$
exactly, as in Theorem~\ref{thm:quotient}.

\emph{Sharpness.}
Fix integers $1 \le a < T$.
Take two memories forming one group, where memory $2$ is retrieved in
episodes $1, \dots, a$ and memory $1$ is never retrieved, and let
$\boldsymbol{\theta} = (0, \theta, 0)$ with $\theta \neq 0$.
Then $D_g = a$, $\bar{\mathbf{X}}_1 = [\mathbf{1}_T \mid
\mathbf{1}_{[a]}]$ has full column rank, and
$\boldsymbol{\mu} = \mathbf{X}_1\boldsymbol{\theta} = \mathbf{0}$, so for
$\bar{\mathbf{c}} = (0, 1)$ the merged log determines the value $0$ for
the group value $s_g = \theta$: the error is $|\theta|$.
The bound~\eqref{eq:approx} equals $|\theta|\sqrt{a}/
\sigma^{+}_{\min}(\bar{\mathbf{X}}_1)$, where
\begin{equation*}
  \sigma^{+}_{\min}(\bar{\mathbf{X}}_1)^2
  = \lambda_{\min}\!\begin{bmatrix} T & a \\ a & a \end{bmatrix}
  = \frac{T + a - \sqrt{(T - a)^2 + 4a^2}}{2}
  \;\longrightarrow\; a
  \quad (T \to \infty).
\end{equation*}
The ratio of the error to the bound therefore tends to $1$.
Consider any bound of the form
$\|\bar{\mathbf{c}}\|\,\gamma(\sigma^{+}_{\min})\sum_i
|\theta_{\{i\}}|\,\rho(D_{g(i)})$.
With $\gamma(\sigma) = 1/\sigma$, the family forces
$\rho(a) \ge \sigma^{+}_{\min}(\bar{\mathbf{X}}_1) \to \sqrt{a}$ for
every integer $a$, so $\rho(D) \ge \sqrt{D}$.
With $\rho(D) = \sqrt{D}$, it forces
$\sigma^{+}_{\min}\,\gamma(\sigma^{+}_{\min}) \ge
\sigma^{+}_{\min}/\sqrt{a}$, and the right-hand side tends to $1$, so no
bound of this form improves $1/\sigma^{+}_{\min}$ by a constant factor
below one on these logs.
Neither factor can be improved while the other is kept.
\end{proof}

\section{Algorithms}
\label{app:algorithms}

This appendix gives LEDGER's three actions as algorithms, in the order of Section~\ref{sec:ledger}: the test and the estimates (Algorithm~\ref{alg:estimate}), merging (Algorithm~\ref{alg:merge}) and retrieval (Algorithm~\ref{alg:retrieve}), followed by the baseline test and the cost of each step. The released code implements them as stated; its names for the parameters are given where they differ from the paper's.

\subsection{The test and the estimates}
\label{app:alg-estimate}

Algorithm~\ref{alg:estimate} decides, for each memory, whether the log determines its value, and estimates only the values it determines. The test comes first, because the ridge solution returns a number for every memory (Lemma~\ref{lem:minnorm}). It uses the kernel of the Gram matrix, which equals $\ker\mathbf{X}_d$ (Appendix~\ref{app:prelim}) and has $p_d$ rows whatever $T$ is.

\begin{algorithm}[h]
\caption{Test and estimate}
\label{alg:estimate}
\begin{algorithmic}[1]
\Require retrieved sets $S_1, \dots, S_T$; outcomes $\mathbf{r}$;
  resolution $d$; ridge parameter $\lambda$; tolerance $\tau$; level
  $\delta$
\Ensure for each memory $i$: an estimate with an interval, or
  \emph{undetermined}
\State build $\mathbf{X}_d$ with $(\mathbf{X}_d)_{tB} =
  \mathbf{1}[B \subseteq S_t]$ for $|B| \le d$
\State $\mathbf{M} \gets \mathbf{X}_d^{\!\top}\mathbf{X}_d$
  \Comment{counts only, Lemma~\ref{lem:gram-d}}
\State $\mathbf{U} \gets$ orthonormal eigenvectors of $\mathbf{M}$ with
  eigenvalue at most $\epsilon_{\mathrm{rank}}\,\|\mathbf{M}\|$
  \Comment{basis of $\ker\mathbf{X}_d$}
\State $\mathbf{V} \gets \mathbf{M} + \lambda\mathbf{I}$;\quad
  $\hat{\boldsymbol{\theta}} \gets \mathbf{V}^{-1}\mathbf{X}_d^{\!\top}
  \mathbf{r}$
\State $\hat{\sigma} \gets \|\mathbf{r} - \mathbf{X}_d
  \hat{\boldsymbol{\theta}}\| / \sqrt{T - \mathrm{tr}(\mathbf{V}^{-1}\mathbf{M})}$
\For{$i = 1, \dots, n$}
  \State $c^{(i)}_B \gets \mathbf{1}[i \in B]/|B|$ for $|B| \le d$
  \If{$\|\mathbf{U}^{\!\top}\mathbf{c}^{(i)}\| \le
      \tau\,\|\mathbf{c}^{(i)}\|$}
    \State $\hat{\varphi}_i \gets (\mathbf{c}^{(i)})^{\!\top}
      \hat{\boldsymbol{\theta}}$;\quad
      $w_i \gets z_{1-\delta/(2n)}\,\hat{\sigma}\,
      \|\mathbf{X}_d\mathbf{V}^{-1}\mathbf{c}^{(i)}\|$
    \State \textbf{report} $\hat{\varphi}_i \pm w_i$
  \Else
    \State \textbf{report} \emph{undetermined}
  \EndIf
\EndFor
\end{algorithmic}
\end{algorithm}

In all experiments, $\lambda = 10^{-2}$, $\tau = 10^{-8}$ and $\delta = 0.05$. The basis $\mathbf{U}$ keeps the eigenvectors of $\mathbf{M}$ with eigenvalue at most $p_d\,\epsilon_{\mathrm{mach}}\max(\lambda_{\max}(\mathbf{M}), 1)$, and $\hat{\sigma}$ uses the effective degrees of freedom $\mathrm{tr}(\mathbf{V}^{-1}\mathbf{M})$. For $p_d > 4{,}000$, the test solves one least squares problem per memory by LSMR instead of forming $\mathbf{U}$. The width uses $\|\mathbf{X}_d\mathbf{V}^{-1}\mathbf{c}\|^2 = \mathbf{c}^{\!\top}\mathbf{V}^{-1}\mathbf{M}\mathbf{V}^{-1}\mathbf{c}$, so it needs $\mathbf{M}$ and not $\mathbf{X}_d$. At $d = 1$, $\mathbf{c}^{(i)} = (0, \mathbf{e}_i)$, so the test reads the $i$-th row of $\mathbf{U}$ and the estimate is $\hat{\theta}_{\{i\}}$.

\subsection{Merging}
\label{app:alg-merge}

Algorithm~\ref{alg:merge} groups memories with identical columns of $\mathbf{X}$ and builds the merged matrix of Theorem~\ref{thm:quotient}. Hashing makes it linear in the size of the log.

\begin{algorithm}[h]
\caption{Merge memories with identical histories}
\label{alg:merge}
\begin{algorithmic}[1]
\Require the matrix $\mathbf{X} \in \{0,1\}^{T \times n}$
\Ensure groups $g_1, \dots, g_m$ and the merged matrix
  $\bar{\mathbf{X}} \in \{0,1\}^{T \times m}$
\State $H \gets$ empty hash table from column keys to groups
\For{$i = 1, \dots, n$}
  \State $\kappa \gets$ the set $\{t : \mathbf{X}_{ti} = 1\}$, as a
    sorted tuple
  \Comment{the retrieval history of $m_i$}
  \State append $i$ to the group $H[\kappa]$, creating it if absent
\EndFor
\State $g_1, \dots, g_m \gets$ the groups of $H$
\State $\bar{\mathbf{X}}_{\cdot g} \gets$ the column of any member of
  $g$, for each group $g$
\State \Return $g_1, \dots, g_m$ and $\bar{\mathbf{X}}$
\end{algorithmic}
\end{algorithm}

The memories never retrieved share the empty history and form one group with a zero column, which stays undetermined. LEDGER then applies Algorithm~\ref{alg:estimate} to $\bar{\mathbf{X}}$, with one unit per group; by Theorem~\ref{thm:quotient}, every value determined before merging is determined after it.

\paragraph{Merging similar histories.} Theorem~\ref{thm:approx} bounds the error of merging memories whose histories differ. Theorem~\ref{thm:approx} bounds the error of merging memories whose histories differ. LEDGER finds candidate pairs of units by locality-sensitive hashing of their columns and keeps those whose columns differ in at most $\delta_{\max} = 8$ episodes. It accepts a candidate unconditionally when the merged unit's total is undetermined before the merge and determined after it, and otherwise when the bound~\eqref{eq:approx}, with each $|\theta_{\{i\}}|$ replaced by $|\hat{\theta}_{\{i\}}| + 1.96\,\mathrm{se}_i$, gives a positive gain in determined values. On the pooled HotpotQA log of Section~\ref{sec:exp-estimation}, it accepts two merges, both of the first kind.

\subsection{Retrieval}
\label{app:alg-retrieve}

Algorithm~\ref{alg:retrieve} implements the rule~\eqref{eq:rule}.
It keeps one counter per memory, and its only state across episodes is
the vector of counts $\mathbf{f}(t)$.

\begin{algorithm}[h]
\caption{LEDGER's retrieval (one episode)}
\label{alg:retrieve}
\begin{algorithmic}[1]
\Require query $q_t$; similarities $\mathrm{sim}(q_t, m_i)$ for all $i$;
  counts $\mathbf{f}(t)$; sizes $\mathcal{K}$ with at least $d + 1$
  values; window $M$; share $\rho$
\Ensure retrieved set $S_t$ and updated counts
\State draw $k_t$ from $\mathcal{K}$
\State $R_t \gets$ the $M$ memories most similar to $q_t$
\State $\ell_t \gets \max(1, \mathrm{round}(\rho k_t))$
\State $L_t \gets$ the $\ell_t$ members of $R_t$ with the smallest
  $f_i(t)$
\State $H_t \gets$ the $k_t - \ell_t$ members of $R_t$ most similar to
  $q_t$
\State $S_t \gets L_t \cup H_t$
  \Comment{$|S_t| < k_t$ when $L_t$ and $H_t$ overlap}
\State $f_i(t + 1) \gets f_i(t) + \mathbf{1}[i \in S_t]$ for every $i$
\State \Return $S_t$
\end{algorithmic}
\end{algorithm}

In the experiments, $\mathcal{K} = \{3, 5, 8\}$, $M = 20$ and $\rho = 0.4$, so $\ell_t = 1, 2, 3$ at $k_t = 3, 5, 8$; the code calls the rule \texttt{aopt0.4} and the window \texttt{M\_CAND}. The size $k_t$ is drawn uniformly from $\mathcal{K}$ by a generator seeded with the seed, the question and the pass, shared by every rule compared in Section~\ref{sec:exp-instrument}, so the rules differ only in which memories they retrieve. Ties in $f_i(t)$ are broken in a fixed order that the sort determines. The rule reads the past only through $\mathbf{f}(t)$, a function of $\mathbf{X}$, so it satisfies the condition of Proposition~\ref{prop:interval}. Because the counts change from pass to pass, one query retrieves different sets over time, which breaks the nesting of Lemma~\ref{lem:nesting}.

\subsection{The baseline test and costs}
\label{app:alg-cost}

The baseline test (Definition~\ref{def:intercept-test}) solves
$\mathbf{X}\mathbf{w} = \mathbf{1}_T$ by least squares and reports the
residual $\|\mathbf{X}\hat{\mathbf{w}} - \mathbf{1}_T\|$.
A residual at the level of rounding, such as $3.1 \times 10^{-15}$ on
pooled HotpotQA under deterministic top-$k$, means that
$\mathbf{1}_T$ lies in $\mathrm{range}\,\mathbf{X}$ and the log contains
a change like $\mathbf{h}_\emptyset$.

Table~\ref{tab:costs} gives the cost of each step, with
$\bar{k} = \frac{1}{T}\sum_t |S_t|$ and
$p_d = \sum_{\ell \le d}\binom{n}{\ell}$.
Building $\mathbf{M}$ touches each pair of memories in each retrieved set
once, which is why its cost depends on $\bar{k}$ and not on $n$.

\begin{table}[h]
\centering
\small
\caption{Cost of each step at $d = 1$, where $p_1 = n + 1$.}
\label{tab:costs}
\renewcommand{\arraystretch}{1.15}
\begin{tabular}{@{}lll@{}}
\toprule
Step & Time & Memory \\
\midrule
Gram matrix $\mathbf{M}$ & $O(T\bar{k}^2 + n^2)$ & $O(n^2)$ \\
Kernel basis $\mathbf{U}$ and $\mathbf{V}^{-1}$ & $O(n^3)$ & $O(n^2)$ \\
Test for all memories & $O(n \cdot \dim\ker\mathbf{X}_1)$ & $O(n^2)$ \\
Estimates and intervals & $O(n^3)$, once & $O(n^2)$ \\
Merging (Algorithm~\ref{alg:merge}) & $O(T\bar{k} + n)$ expected & $O(T\bar{k} + n)$ \\
Retrieval, one episode & $O(n + M\log M)$ & $O(n)$ \\
Rank-one update of $\mathbf{V}^{-1}$ & $O(n^2)$ & $O(n^2)$ \\
\bottomrule
\end{tabular}
\end{table}

The cubic step dominates.
On pooled HotpotQA, with $n = 2{,}971$, it is one eigendecomposition of a
$2{,}972 \times 2{,}972$ matrix.
Merging reads each retrieval once, since the history of $m_i$ is the list
of episodes that retrieved it; this is the $O(nT)$ bound of
Section~\ref{sec:consolidation} for a dense representation of
$\mathbf{X}$, and $O(T\bar{k} + n)$ for the sparse one.
\providecommand{\verify}[1]{\textcolor{red}{[#1]}}

\section{Experimental Setup}
\label{app:setup}

This appendix specifies every experiment of Section~\ref{sec:experiments}: the corpora and stores (Appendix~\ref{app:setup-data}), retrieval and reading (Appendix~\ref{app:setup-reading}), the retrieval rules (Appendix~\ref{app:setup-rules}), the synthetic logs (Appendix~\ref{app:setup-synthetic}), and the statistics (Appendix~\ref{app:setup-stats}). Table~\ref{tab:setup-runs} maps each number in the main text to the run that produces it. The audit of fourteen benchmarks has its own protocol, given in Appendix~\ref{app:audit}.

\subsection{Corpora and stores}
\label{app:setup-data}

The retrieval experiments use three corpora with different structure.

\paragraph{HotpotQA.} Each question of HotpotQA~\citep{hotpotqa2018} comes with ten paragraphs, two of which support the answer. We use the paragraphs as memories and pool the paragraphs of many questions into one shared store, so that a paragraph of one question can be retrieved for another. A few validation items ship fewer than ten paragraphs, and we do not de-duplicate, so $300$ questions give $n = 2{,}971$ memories. The gold memories of a question are its supporting paragraphs.

\paragraph{LongMemEval.} Each question of LongMemEval~\citep{longmemeval2025} comes with a chat history. We use its turns as memories and pool the histories of the questions into one shared store. We use the oracle split. The gold memories of a question are its turns flagged \texttt{has answer}; the accuracy comparison uses the $479$ of $500$ questions that have at least one such turn.

\paragraph{LoCoMo.} LoCoMo~\citep{locomo2024} contains long conversations, each with its own questions. We use the turns of a conversation as memories, and all questions of a conversation share its store; retrieval for a question is restricted to its conversation. The gold memories of a question are the turns cited as its evidence. 9 of the $1{,}542$ questions cite no evidence and are excluded from gold recall.

\subsection{Retrieval and reading}
\label{app:setup-reading}

Every memory and every question is embedded with the sentence encoder \texttt{all-MiniLM-L6 v2}~\citep{reimers2019sbert}, normalised to unit length, so similarity is the cosine of the embeddings. All rules rank memories by this similarity; they differ in which memories of the ranking they retrieve (Appendix~\ref{app:setup-rules}). The reader is \texttt{gemini-2.5-flash-lite} at temperature $0$, prompted with the question and the retrieved memories in retrieval order, up to $24{,}000$ characters. An answer that matches a gold answer exactly after normalization is correct; otherwise the same model, with a fixed judge prompt, decides. Both prompts are released with the code. Two further readers, \texttt{gemini-3.6-flash} and \texttt{gemini-3.7-flash}, are used only to measure the cost of pooling a store on LongMemEval (Appendix~\ref{app:res-readers}). Hosted models cannot be pinned, so accuracies are reproducible in design and not bit for bit; the retrieval logs and every identifiability number regenerate byte for byte from the released inputs.

\subsection{Retrieval rules}
\label{app:setup-rules}

Four rules are compared in Section~\ref{sec:exp-instrument}.
All four draw the size $k_t$ of each retrieval from
$\mathcal{K} = \{3, 5, 8\}$ with the same seeded function of the question
and the pass, so they differ only in which memories they retrieve.

\begin{itemize}
  \item \emph{Deterministic top-$k$} retrieves the $k_t$ memories most
    similar to the query.
  \item \emph{LEDGER} applies the rule~\eqref{eq:rule} with $M = 20$ and
    $\rho = 0.4$ (Algorithm~\ref{alg:retrieve}); the code calls it
    \texttt{aopt0.4}.
  \item \emph{$\varepsilon$-greedy} replaces, with probability
    $\varepsilon$, the top-$k_t$ set by a uniformly random $k_t$-subset of
    the whole store; we use $\varepsilon \in \{0.1, 0.3\}$.
  \item \emph{Score perturbation} adds independent Gumbel noise of scale
    $\beta$ to each similarity and retrieves the top $k_t$ of the
    perturbed scores; we use $\beta \in \{0.05, 0.10\}$.
\end{itemize}

LEDGER's sets fall below the drawn size when $L_t$ and $H_t$ overlap. On pooled HotpotQA this happens in $1.0\%$, $3.3\%$ and $15.9\%$ of episodes at $k_t = 3$, $5$ and $8$, for a mean size of $5.25$ against $5.33$ for the other rules, whose sets always have size $k_t$. The realized sizes take at least three distinct values, so Proposition~\ref{prop:sizes} is satisfied at $d = 1$ and $d = 2$. The LoCoMo merging run of Section~\ref{sec:exp-estimation} uses score perturbation with $\beta = 0.08$.

\subsection{Synthetic logs}
\label{app:setup-synthetic}

Coverage requires the true values, so the intervals are tested on synthetic logs drawn from the model~\eqref{eq:model}. A setting fixes the number of memories $n \in \{10, 30\}$, the retrieval size $k \in \{3, 5\}$, the ratio $T/n \in \{2, 10, 50\}$ and the ridge parameter $\lambda \in \{10^{-4}, 10^{-2}, 10^{-1}\}$, which gives $36$ settings at each resolution $d \in \{1, 2\}$. For each setting we draw $20$ random designs and, for each design, $100$ noise vectors, and we record whether every interval that Algorithm~\ref{alg:estimate} reports contains its true value. Joint coverage is the fraction of the $2{,}000$ draws in which all reported intervals do. Each design draws, for every episode, a size uniformly from $\{1, \dots, k\}$ and a uniformly random set of that size, so $k$ is the largest size. The dividends are independent standard normal, and the noise is standard normal with $\sigma = 1$, which the interval is given. At $d = 1$ every design has full rank; at $d = 2$, $24$ of the $36$
settings give rank-deficient designs. The restriction of coverage to reported values (Section~\ref{sec:exp-estimation}) is a post hoc analysis of the same draws; Appendix~\ref{app:corrections} records it as such.

The baselines of the width comparison are the pointwise Gaussian interval, which uses the true $\sigma$ and no multiplicity correction, and the self-normalised confidence set of \citet{abbasi2011oful}, evaluated at $\lambda = 1$ on a separate grid of $n$, $k$ and $T$ (Appendix~\ref{app:results}).

\subsection{Statistics}
\label{app:setup-stats}

Every accuracy comparison is paired: two rules answer the same questions, and we report the mean difference in accuracy with a $95\%$ interval computed by $10{,}000$ paired bootstrap resamples over questions (percentile interval) and a two-sided paired permutation test with $10{,}000$ sign flips. The four comparisons against the random rules share a family and are Holm-corrected. Fractions determined on pooled HotpotQA are averaged over five seeds ($0$ to $4$), with a range across seeds reported in Appendix~\ref{app:results}. Every claim with a decision rule was pre-registered before its measurement; Appendix~\ref{app:corrections} lists the pre-registered claims, the three that were withdrawn, and the one post hoc analysis.

\subsection{Runs behind the main text}
\label{app:setup-runs}

\begin{table}[h]
\centering
\small
\caption{The run behind each result in Section~\ref{sec:experiments}. Artifacts are named as in the released code.}
\label{tab:setup-runs}
\renewcommand{\arraystretch}{1.15}
\begin{tabular}{@{}>{\raggedright\arraybackslash}p{0.29\linewidth}>{\raggedright\arraybackslash}p{0.36\linewidth}>{\raggedright\arraybackslash}p{0.27\linewidth}@{}}
\toprule
Result & Design & Artifact \\
\midrule
Audit, $55$ configurations (Table~\ref{tab:audit})
  & fourteen benchmarks, one pass, their own retrieval
  & \texttt{audit\_stageB.csv} \\
Fraction determined, four rules (Figure~\ref{fig:results}a)
  & pooled HotpotQA, $300$ questions $\times$ $24$ passes,
    $n = 2{,}971$, $T = 7{,}200$, seeds $0$--$4$
  & \texttt{e1\_}\\ && \texttt{identifiability.csv} \\
Baseline-test residuals
  & same log as above
  & \texttt{e1\_ident\_}\\ && \texttt{both\_bases.csv} \\
Accuracy, LEDGER against top-$k$
  & pooled HotpotQA, $1{,}600$ held-out questions, store of
    $15{,}900$ memories
  & \texttt{e1\_primary.csv} \\
Accuracy, random rules
  & pooled HotpotQA, $300$ questions, one pass, Holm-corrected
  & \texttt{e1\_frontier\_}\\ && \texttt{table.csv} \\
Accuracy on LoCoMo and LongMemEval (Figure~\ref{fig:results}b)
  & $1{,}542$ and $479$ questions, deterministic against
    \texttt{aopt0.4}
  & \texttt{e1\_replicate\_}\\ && \texttt{summary.csv} \\
Mechanism and oracle (Figure~\ref{fig:results}c)
  & pooled HotpotQA, split by gold recall of top-$k$
  & \texttt{e1\_mechanism.csv},\\ && \texttt{e1\_gated.csv} \\
Coverage and widths
  & $36$ settings $\times$ $20$ designs $\times$ $100$ draws per $d$
  & \texttt{v14\_recommended.csv},\\ && \texttt{v12ab\_all\_cells.csv} \\
Coverage on reported values (post hoc)
  & same draws as above
  & \texttt{posthoc/v14\_guard\_}\\ && \texttt{coverage.csv} \\
Merging on HotpotQA
  & pooled HotpotQA, $100$ questions $\times$ $24$ passes, $n = 992$,
    $T = 2{,}400$, \texttt{aopt0.4}
  & \texttt{x1\_merge.csv} \\
Merging on LoCoMo
  & two conversations, $304$ questions, $n = 788$, $T = 912$, three passes, score perturbation $\beta = 0.08$, $300$ sampled columns
  & \texttt{locomo\_}\\ && \texttt{identifiability.csv} \\
\bottomrule
\end{tabular}
\end{table}

\paragraph{Compute.}
All identifiability computations, merging and synthetic experiments run on a CPU; the largest step is one eigendecomposition of a $2{,}972 \times 2{,}972$ matrix per log. Reader and judge calls go to hosted APIs; the number of calls per experiment is not logged per run; the per-row records account for at least $16{,}742$ reader calls.
\providecommand{\verify}[1]{\textcolor{red}{[#1]}}

\section{Benchmark Audit}
\label{app:audit}

This appendix gives the protocol of the audit in Section~\ref{sec:exp-protocol} (Appendix~\ref{app:audit-protocol}), proves that every one of its $55$ configurations determines no memory value (Appendix~\ref{app:audit-why}), and describes each benchmark (Appendix~\ref{app:audit-benchmarks}).

\subsection{Protocol}
\label{app:audit-protocol}

For each benchmark, we build its store as the benchmark specifies and run its retrieval once over its questions, one episode per question, at each retrieval size it uses. We add the sizes $3$, $5$ and $10$, at the size it specifies where that differs, and at all for the four long-context benchmarks. Each configuration gives a log $S_1, \dots, S_T$, from which we build $\mathbf{X}_1$ and test every memory with Algorithm~\ref{alg:estimate} at $d = 1$, with the baseline column included. We also compute the blocks of the log (Definition~\ref{def:blocks}), the mean retrieval size, and the number of memories never retrieved. The audit uses the retrieved sets and no outcome, so it calls no reader: by Appendix~\ref{app:prelim}, whether a value is determined does not depend on the outcomes. By Lemma~\ref{lem:resolution}, a memory undetermined at $d = 1$ is undetermined at every $d$, so the audit covers every resolution.

\subsection{Why every configuration determines nothing}
\label{app:audit-why}

The audit finds no determined value in any of the $55$ configurations (Table~\ref{tab:audit}). The following lemma explains all of them at once. It extends Theorem~\ref{thm:fixedk} from one retrieval size to one size per block.

\begin{lemma}[Constant size within each block]
\label{lem:audit}
Suppose that, within each block $b$ of a log, every episode retrieves the same number $k_b$ of memories. Then no memory is identifiable at any resolution.
\end{lemma}

\begin{proof}
Define $\mathbf{w} \in \mathbb{R}^n$ by $w_i = 1/k_b$ for every memory $i$ of block $b$, and $w_i = 0$ for memories never retrieved. An episode of block $b$ retrieves $k_b$ memories, all in block $b$, so $(\mathbf{X}\mathbf{w})_t = k_b \cdot (1/k_b) = 1$ for every $t$. Hence $(-1, \mathbf{w}) \in \ker\mathbf{X}_1$, and for every retrieved memory $i$ the inner product of $(0, \mathbf{e}_i)$ with this vector is $w_i \neq 0$. A memory never retrieved has a zero column, so $(0, \mathbf{e}_i) \in \ker\mathbf{X}_1$ and its inner product with itself is $1$. No memory is identifiable at $d = 1$, and by Lemma~\ref{lem:resolution} at no higher resolution.
\end{proof}
We verified its hypothesis on every one of the $55$ configurations. The hypothesis holds in every configuration of the audit. At a fixed $k$, every episode of the log retrieves $k$ memories. At ``all'', every episode of a context retrieves the whole context, so the size is constant within each context, and each block lies within one context. The vector $(-1, \mathbf{w})$ is a change like $\mathbf{h}_\emptyset$: it lowers the baseline and raises each memory's value by the inverse of its block's retrieval size.

Blocks limit the log further, and would limit it even if the sizes varied. In the six benchmarks with a store per question and one episode per question, every block has $T_b = 1$, so by Proposition~\ref{prop:blocks} at most one value per question could be determined, whatever the retrieval. The larger HotpotQA run, with $1{,}200$ questions, confirms that more questions do not help: the number of blocks grows with the number of questions, and every block still has one episode. The two benchmarks with a shared store, MultiHopRAG and HELMET, split into fewer and larger blocks: MultiHopRAG from $269$ blocks at $k = 3$ to $15$ at $k = 10$, and HELMET into $41$ blocks of $8.7$ episodes on average at every $k$. Their blocks have several episodes each, and the fixed size alone leaves every value undetermined. A shared store is therefore necessary and not sufficient, as Section~\ref{sec:exp-protocol} states.

\subsection{The benchmarks}
\label{app:audit-benchmarks}

Table~\ref{tab:audit-benchmarks} gives, for each benchmark, the unit we use as a memory, the scope of its store and the retrieval sizes of the audit. The size the benchmark itself specifies is in bold. The counts $T$, $n$ and the blocks are in Table~\ref{tab:audit}.

\begin{table}[h]
\centering
\small
\caption{The fourteen audited benchmarks. ``All'' retrieves every memory of the context; bold marks the size the benchmark specifies, where it specifies one.}
\label{tab:audit-benchmarks}
\renewcommand{\arraystretch}{1.15}
\begin{tabular}{@{}>{\raggedright\arraybackslash}p{0.26\linewidth}
                >{\raggedright\arraybackslash}p{0.27\linewidth}
                >{\raggedright\arraybackslash}p{0.18\linewidth}
                >{\raggedright\arraybackslash}p{0.17\linewidth}@{}}
\toprule
Benchmark & Memory & Store & Sizes \\
\midrule
HotpotQA~\citep{hotpotqa2018} & paragraph & per question & 3, 5, 10 \\
MuSiQue~\citep{trivedi2022musique} & paragraph & per question & 3, 5, 10 \\
WikiHop~\citep{welbl2018constructing} & supporting document & per question & 3, 5, 10 \\
TriviaQA~\citep{joshi2017triviaqa} & paragraph of the Wikipedia contex & per question & 3, 5, 10 \\
LongMemEval~\citep{longmemeval2025} & turn & per question & 3, 5, 10 \\
MSC-Self-Instruct~\citep{xu2021beyond} & turn& per dialogue & 3, 5, 10 \\
BABILong~\citep{kuratov2024babilong} & sentence & per item & 3, 5, 10\\
InfiniteBench~\citep{zhang2024bench} & paragraph, first 1,000 per book & per book & 3, 5, 10 \\
NarrativeQA, summary~\citep{kovcisky2018narrativeqa} & paragraph of the full text & per document & 3, 5, 10 \\
NarrativeQA, full text & chunk of the full text & per document & 3, 5, 10 \\
QuALITY~\citep{pang2022quality} & paragraph & per article & 3, 5, 10 \\
LoCoMo~\citep{locomo2024} & turn & per conversation & 3, 5, 10 \\
MemoryAgentBench~\citep{hu2026evaluating} & paragraph & per context & 3, 5, 10 \\
MultiHopRAG~\citep{tang2024multihop} & paragraph & shared & 3, 5, \textbf{6}, 10 \\
HELMET~\citep{yen2025helmet} & passage & shared, $40{,}000$ & 3, 5, 10, \textbf{14} \\
\bottomrule
\end{tabular}
\end{table}

Three details affect the counts. In HotpotQA, $n = 2{,}971$ for $300$ questions because a few validation items ship fewer than ten paragraphs; at $k = 10$, every paragraph of a question is retrieved. In the ``all'' configurations, the log keeps $T$ equal to the number of questions and changes only the retrieval size, whose mean is $104.0$ for BABILong, $942.4$ for InfiniteBench, $125.5$ for QuALITY and $1{,}372.5$ for NarrativeQA on the full text. The HELMET store is capped at $40{,}000$ passages by us, for compute.
\providecommand{\verify}[1]{\textcolor{red}{[#1]}}

\section{Additional Results}
\label{app:results}

This appendix reports the full results behind Section~\ref{sec:experiments}, in its order: retrieval on pooled HotpotQA (Appendix~\ref{app:res-frontier}), the replications on LoCoMo and LongMemEval (Appendix~\ref{app:res-replicate}), the mechanism of the accuracy cost (Appendix~\ref{app:res-mechanism}), additional readers (Appendix~\ref{app:res-readers}), intervals (Appendix~\ref{app:res-intervals}), merging (Appendix~\ref{app:res-merging}), and the uses of the determined values (Appendix~\ref{app:res-uses}). Appendix~\ref{app:setup} specifies every run, and Table~\ref{tab:setup-runs} names its artifact.

\subsection{Retrieval on pooled HotpotQA}
\label{app:res-frontier}

Table~\ref{tab:res-frontier} gives every point of Figure~\ref{fig:results}a. The fraction determined is computed on $300$ questions over $24$ passes ($n = 2{,}971$, $T = 7{,}200$) and averaged over five seeds; LEDGER's accuracy is measured on $1{,}600$ held-out questions, and the random rules' on $300$ questions with Holm-corrected tests.

\begin{table}[h]
\centering
\small
\caption{Retrieval rules on pooled HotpotQA. Fractions determined are
with the baseline column fitted and, in parentheses, without it; accuracy
changes are against deterministic top-$k$, in points, with $95\%$
intervals in brackets. LEDGER's accuracy is measured on $1{,}600$
questions; the random rules' on $300$, with Holm-corrected $p$-values.}
\label{tab:res-frontier}
\renewcommand{\arraystretch}{1.15}
\begin{tabular}{@{}llcccc@{}}
\toprule
Rule & Parameter & Determined & Mean $|S_t|$ & Accuracy change & $p$ \\
\midrule
Top-$k$ & --- & $0.001$ ($0.003$) & $5.33$ & --- & --- \\
LEDGER & $\rho = 0.4$, $M = 20$ & $0.797$ ($0.797$) & $5.25$ & $-2.6$ $[-4.2, -0.9]$ & $0.004$ \\
$\varepsilon$-greedy & $\varepsilon = 0.1$ & $0.001$ ($0.003$) & $5.33$ & $-3.0$ $[-5.3, -1.0]$ & $0.012$ \\
$\varepsilon$-greedy & $\varepsilon = 0.3$ & $0.989$ ($0.989$) & $5.33$ & $-13.3$ $[-17.3, -9.7]$ & $<0.001$ \\
Perturbation & $\beta = 0.05$ & $0.983$ ($0.983$) & $5.33$ & $-14.0$ $[-19.3, -8.7]$ & $<0.001$ \\
Perturbation & $\beta = 0.10$ & $1.000$ ($1.000$) & $5.33$ & $-37.0$ $[-43.0, -30.7]$ & $<0.001$ \\
\bottomrule
\end{tabular}
\end{table}

The fraction determined under $\varepsilon$-greedy with $\varepsilon = 0.1$ equals that of top-$k$ with and without the baseline column. The baseline-test residual is $3.1 \times 10^{-15}$ under deterministic top-$k$, between $2.8 \times 10^{-15}$ and $3.4 \times 10^{-15}$ across the five seeds, and $9.2 \times 10^{-2}$ under LEDGER's rule.

\subsection{Replications on LoCoMo and LongMemEval}
\label{app:res-replicate}

Table~\ref{tab:res-replicate} gives the accuracy comparison on the two other corpora, with the same rules and sizes. On LoCoMo, LEDGER raises accuracy; on LongMemEval, the change is not distinguishable from zero.

\begin{table}[h]
\centering
\small
\caption{Accuracy of deterministic top-$k$ and LEDGER on LoCoMo and
LongMemEval, with the change in points and its $95\%$ interval.}
\label{tab:res-replicate}
\renewcommand{\arraystretch}{1.15}
\begin{tabular}{@{}lrcccc@{}}
\toprule
Corpus & Questions & Top-$k$ & LEDGER & Change & $95\%$ interval \\
\midrule
LoCoMo & $1{,}542$ & $0.191$ & $0.217$ & $+2.7$ & $[+1.0, +4.4]$ \\
LongMemEval & $479$ & $0.263$ & $0.271$ & $+0.8$ & $[-1.9, +3.8]$ \\
\bottomrule
\end{tabular}
\end{table}

On LoCoMo we also compare the fraction determined across rules, on two conversations with $n = 788$ memories and $120$ questions over $12$ passes ($T = 1{,}440$). Deterministic top-$k$ determines $0.193$ of the memories, $\varepsilon$-greedy with $\varepsilon = 0.3$ determines $0.217$, LEDGER $0.321$, and score perturbation $0.445$ at $\beta = 0.05$ and $0.532$ at $\beta = 0.10$. On this store, deterministic top-$k$ determines more than on pooled HotpotQA, and LEDGER determines less than score perturbation.

\subsection{The mechanism of the accuracy cost}
\label{app:res-mechanism}

Table~\ref{tab:res-mechanism} gives the split of Figure~\ref{fig:results}. On pooled HotpotQA, top-$k$ retrieves $0.93$ of the gold memories, and LEDGER lowers the number of gold memories retrieved per question from $2.000$ to $1.758$ on questions where top-$k$ retrieved both.

\begin{table}[h]
\centering
\small
\caption{LEDGER against top-$k$ on pooled HotpotQA, split by whether
top-$k$ retrieved every gold memory of the question.}
\label{tab:res-mechanism}
\renewcommand{\arraystretch}{1.15}
\begin{tabular}{@{}lcccc@{}}
\toprule
Questions & Share & Gold lost per question & Accuracy change & $95\%$ interval \\
\midrule
Top-$k$ found all gold & $0.42$ & $0.242$ & $-6.9$ $[-9.6, -4.3]$ & $0.242$ \\
Top-$k$ missed some gold & $0.58$ & $0.009$ & $+0.7$ $[-1.4, +2.7]$ & $0.009$ \\
\bottomrule
\end{tabular}
\end{table}

The pre-registered query-time substitute applies LEDGER when the query's top-$1$ similarity is below its median over the $1{,}600$ questions, $0.647$. It changes accuracy by $-1.3$ points ($[-2.6, -0.1]$) against top-$k$ and by $+1.25$ ($[0.0, +2.4]$) against ungated LEDGER, and determines $0.520$ of the memories. The pre-registered rule required the second interval to exclude zero, so the gate does not transfer the oracle's result (Appendix~\ref{app:corrections}).

\subsection{Pooling across readers}
\label{app:res-readers}

Table~\ref{tab:res-readers} repeats the comparison of LEDGER and top-$k$ on pooled HotpotQA with two further readers. The retrieval logs, and therefore every fraction determined, are the same for all readers; only the accuracies change.

\begin{table}[h]
\centering
\small
\caption{Accuracy of three readers on LongMemEval when each question retrieves from its own store and from one pooled store, with the cost of pooling in points and its $95\%$ interval; $300$ questions per cell. The gold column gives the reader only the evidence sessions.}
\label{tab:res-readers}
\renewcommand{\arraystretch}{1.15}
\begin{tabular}{@{}lcccc@{}}
\toprule
Reader & Gold & Own store & Pooled store & Cost of pooling \\
\midrule
\texttt{gemini-2.5-flash-lite} & $0.607$ & $0.417$ & $0.260$ & $15.7$ $[11.0, 20.7]$ \\
\texttt{gemini-3.6-flash}      & $0.783$ & $0.497$ & $0.287$ & $21.0$ $[15.7, 26.3]$ \\
\texttt{gemini-3.7-flash}      & $0.757$ & $0.477$ & $0.280$ & $19.7$ $[14.7, 25.0]$ \\
\bottomrule
\end{tabular}
\end{table}

\subsection{Intervals}
\label{app:res-intervals}

Table~\ref{tab:res-coverage} gives joint coverage at $\delta = 0.05$ on the synthetic logs of Appendix~\ref{app:setup-synthetic}, as the minimum over settings. Coverage over every memory counts the intervals of undetermined memories, which LEDGER does not report; coverage over reported values counts only the intervals that Algorithm~\ref{alg:estimate} returns.

\begin{table}[h]
\centering
\small
\caption{Lowest joint coverage over settings at $\delta = 0.05$, with the
target $0.95$.
The last column restricts coverage to the values LEDGER reports; it is a
post hoc analysis.}
\label{tab:res-coverage}
\renewcommand{\arraystretch}{1.15}
\begin{tabular}{@{}llcccc@{}}
\toprule
$d$ & Designs & Settings & Settings reporting & Every memory & Reported values \\
\midrule
$1$ & full rank & $36$ & $36$ & $0.977$ & $0.977$ \\
$2$ & full rank & $12$ & $12$ & $0.994$ & $0.994$ \\
$2$ & rank-deficient & $24$ & $6$ & $0.572$ & $0.995$ \\
\bottomrule
\end{tabular}
\end{table}

In the $18$ rank-deficient settings that report nothing, every memory of every design is undetermined. Their coverage over every memory ranges down to $0.572$, and adding the exact ridge-bias term of Proposition~\ref{prop:interval}, which needs the true $\boldsymbol{\theta}$, raises the minimum over all $36$ settings at $d = 2$ to $0.992$. This confirms the explanation of Appendix~\ref{app:proofs-interval}: the undercoverage comes from the part $\mathbf{c}_2^{\!\top} \boldsymbol{\theta}$ of undetermined values. The coverage of a design does not predict whether it reports: one rank-deficient setting ($n = 30$, $k = 3$, $T = 300$) has coverage $0.976$ over every memory and reports nothing, because every memory there has a component in the kernel. A separate grid at $\lambda = 1$ contains one rank-deficient design at $d = 1$, with coverage $0.964$ over $20{,}000$ draws.

Table~\ref{tab:res-widths} compares widths. LEDGER's interval is close to the pointwise Gaussian interval that knows $\sigma$; the self-normalised confidence set is several times wider at $d = 1$ and an order of magnitude wider at $d = 2$, because its width grows with the number of dividends.

\begin{table}[h]
\centering
\small
\caption{Width of each interval relative to the pointwise Gaussian
interval with the true $\sigma$: median over settings, with the range.}
\label{tab:res-widths}
\renewcommand{\arraystretch}{1.15}
\begin{tabular}{@{}lcc@{}}
\toprule
Interval & $d = 1$ & $d = 2$ \\
\midrule
LEDGER, Eq.~\eqref{eq:estimate} & $1.52$ & $1.43$ \\
\citet{abbasi2011oful}, $\lambda = 1$ & $6.6$ ($4.7$--$9.4$) & $35.9$ ($11.0$--$365.7$) \\
\bottomrule
\end{tabular}
\end{table}

\subsection{Merging}
\label{app:res-merging}

Table~\ref{tab:res-merging} gives the merging results on pooled HotpotQA
with $100$ questions over $24$ passes ($n = 992$, $T = 2{,}400$), for
two retrieval rules.
Fractions are over memories before merging and over units after it.

\begin{table}[h]
\centering
\small
\caption{Units and fraction determined before and after merging on
pooled HotpotQA.}
\label{tab:res-merging}
\renewcommand{\arraystretch}{1.15}
\begin{tabular}{@{}lcccccc@{}}
\toprule
& \multicolumn{2}{c}{Log} & \multicolumn{2}{c}{Identical histories}
& \multicolumn{2}{c}{Similar histories} \\
\cmidrule(lr){2-3}\cmidrule(lr){4-5}\cmidrule(lr){6-7}
Rule & Units & Determined & Units & Determined & Units & Determined \\
\midrule
LEDGER & $992$ & $0.815$ & $869$ & $0.992$ & $867$ & $0.995$ \\
Perturbation, $\beta = 0.05$ & $992$ & $0.969$ & $964$ & $0.999$ & $964$ & $0.999$ \\
\bottomrule
\end{tabular}
\end{table}

Merging identical histories removes $123$ units under LEDGER and $28$
under score perturbation, which ties fewer memories.
Merging similar histories, with candidates that differ in at most eight
episodes, accepts two merges under LEDGER and none under perturbation.
On LoCoMo, after three passes of score perturbation with $\beta = 0.08$,
$79.0\%$ of the memories are determined, and merging raises this to
$95.7\%$ of the units, both estimated on $300$ randomly sampled columns.

\subsection{Uses of the determined values}
\label{app:res-uses}

\paragraph{Value recovery.}
On synthetic logs with $40$ memories and $T = 4{,}000$, over $30$ seeds, the Spearman correlation with the true Shapley values is $0.892$ for LEDGER on the $31$ memories it reports, $0.804$ when LEDGER splits each group total equally to score all $40$, $0.724$ for reward averaging and $0.118$ for retrieval frequency.

\paragraph{Retention under a budget.}
Ranking undetermined memories last and keeping $30\%$ of the store, the retained store succeeds in $5.52$ of $300$ evaluation episodes under LEDGER, $6.55$ under the equal split, $7.14$ under reward averaging and $6.48$ under the true Shapley values; the pre-registered real-data retention test was not completed.
\providecommand{\verify}[1]{\textcolor{red}{[#1]}}

\section{Pre-registration and Corrections}
\label{app:corrections}

This appendix records how the empirical claims were tested: which were
pre-registered and with what decision rule
(Appendix~\ref{app:corr-prereg}), which were withdrawn
(Appendix~\ref{app:corr-withdrawn}), the one analysis that was not
pre-registered (Appendix~\ref{app:corr-posthoc}), an amendment on the
name of the retrieval rule (Appendix~\ref{app:corr-amendment}), and the
defects we found in our own pipeline with the values they changed
(Appendix~\ref{app:corr-errata}).
The dated documents are in the \texttt{preregistration/} directory of the
released code, and the errata in \texttt{docs/errata.md}.

\subsection{Pre-registered claims}
\label{app:corr-prereg}

Before each measurement, we recorded the claim, the quantity that tests
it and the rule that decides it, in a dated document committed to the
repository.
Table~\ref{tab:corr-prereg} lists the claims and their outcomes.
A claim holds when its measured quantity, with its interval, satisfies
the rule; otherwise it is withdrawn and reported as such.

\begin{table}[h]
\centering
\footnotesize
\caption{Pre-registered claims, their decision rules and outcomes, in the
order of the pre-registration documents. ``Where'' gives the section that
reports the result; a dash marks a claim reported only in the released
documents.}
\label{tab:corr-prereg}
\renewcommand{\arraystretch}{1.15}
\begin{tabular}{@{}>{\raggedright\arraybackslash}p{0.26\linewidth}
                >{\raggedright\arraybackslash}p{0.34\linewidth}
                >{\raggedright\arraybackslash}p{0.10\linewidth}
                >{\raggedright\arraybackslash}p{0.22\linewidth}@{}}
\toprule
Claim & Decision rule & Where & Outcome \\
\midrule
Standard benchmarks determine no memory value
  & falsified if a private-store protocol gives fewer than $0.9$ times
    as many blocks as contexts, or a fraction above $0.01$ at a fixed $k$
  & \ref{sec:exp-protocol}
  & held: $0$ of $55$ configurations \\
The remedy recovers identifiability on three audited benchmarks
  & fractions above floors of $0.50$, $0.22$, $0.13$; falsified below
    $0.10$ or a quarter of the rank bound
  & \ref{app:results}
  & failed: $0.46$, $0.13$, $0.08$; falsified on MemoryAgentBench \\
Reach is the binding ceiling
  & falsified if the window arm is flat or non-monotone, or the passes
    arm rises by $0.05$ or more
  & ---
  & failed on two of three benchmarks \\
The crossover rule on synthetic logs
  & rises below $R = T$, falls above, peak in $[0.5, 2]$, range at least
    $0.02$
  & ---
  & held: peak at $R/T = 0.75$, range $0.14$ \\
Instrumentation is approximately free
  & interval contains zero with half-width below $0.025$
  & \ref{sec:exp-instrument}
  & withdrawn: $-2.6$ $[-4.2, -0.9]$ \\
Displacement is the mechanism
  & gold retrieved per question on found-gold questions below $2.00$
  & \ref{sec:exp-instrument}
  & held: $1.758$ \\
The effect's sign differs by partition
  & negative where top-$k$ found all gold; non-negative elsewhere
  & \ref{sec:exp-instrument}
  & held: $-6.9$ $[-9.6, -4.3]$; $+0.7$ $[-1.4, +2.7]$ \\
A label-free gate transfers the oracle's result
  & interval of gated minus ungated excludes zero
  & \ref{app:res-mechanism}
  & failed: $+1.25$ $[0.0, +2.4]$ \\
The pooling cost grows with store size
  & interval of largest minus smallest store excludes zero
  & ---
  & held at $100$ questions ($+11$ $[+4, +18]$); null at $300$ ($+2$
    $[-2, +6]$) \\
The pooling cost grows with reader quality
  & pairwise differences between readers exclude zero
  & \ref{app:res-readers}
  & withdrawn: two of three contain zero \\
Retention by LEDGER loses less than retention by frequency
  & paired interval at a $50\%$ budget excludes zero
  & ---
  & not completed \\
The sample-complexity bound (check V7)
  & fitted slope at least $1.00$
  & ---
  & withdrawn: slope $0.82$ \\
The validation suite
  & each check's pass criterion
  & ---
  & $21$ of $22$ checks pass; V7 fails \\
\bottomrule
\end{tabular}
\end{table}

\subsection{Withdrawn claims}
\label{app:corr-withdrawn}

Three pre-registered claims were withdrawn because their measured values did not meet their decision rules. We report each with the value it was measured at.

\begin{enumerate}
  \item \emph{Instrumentation is approximately free.} The rule required the interval on LEDGER's accuracy change against top-$k$ to contain zero with a half-width below $0.025$. On $1{,}600$ questions of pooled HotpotQA the change is $-2.6$ points ($[-4.2, -0.9]$, permutation $p = 0.004$), so the paper reports a cost (Section~\ref{sec:exp-instrument}).
  \item \emph{The cost of pooling grows with reader quality.} The rule required the pairwise differences between readers to exclude zero. The costs are $15.7$, $21.0$ and $19.7$ points, and two of the three pairwise intervals contain zero (Appendix~\ref{app:res-readers}).
  \item \emph{The sample-complexity bound.} Validation check V7 required the fitted slope of the error against the bound to be at least $1.00$; it is $0.82$, which is why $21$ of the $22$ checks pass.
\end{enumerate}

Four further pre-registered tests had negative or incomplete outcomes and are reported in Table~\ref{tab:corr-prereg}: the remedy on three audited benchmarks, the reach ceiling, the label-free gate and the retention comparison. The gate was pre-registered as a query-time substitute for the oracle of Section~\ref{sec:exp-instrument}. It applies LEDGER when the query's top-$1$ similarity is below its median, and it did not transfer the oracle's result (Appendix~\ref{app:res-mechanism}).

\subsection{The post hoc analysis}
\label{app:corr-posthoc}

One result in the paper was not pre-registered: joint coverage restricted to the values that LEDGER reports (Section~\ref{sec:exp-estimation} and Table~\ref{tab:res-coverage}). The pre-registered coverage test counted the intervals of every memory, including undetermined ones that LEDGER does not report, and at $d = 2$ it fell to $0.572$ on rank-deficient designs. After seeing this, we restricted coverage to the reported values, which gives a minimum of $0.994$ over all $36$ settings at $d = 2$. The restriction was chosen after the failure was observed, so it is an analysis and not a test. It uses the same seeds, settings and draws as the pre-registered test, whose numbers are unchanged. Its script and output are released separately from the pre-registeredchecks, and the errata record it as post hoc with its date.

\subsection{Amendment on the retrieval rule}
\label{app:corr-amendment}

One pre-registration document describes LEDGER's retrieval rule as  $A$-optimal tie-breaking. The rule that was run is~\eqref{eq:rule}: it fills $\max(1, \mathrm{round}(\rho k_t))$ places with the least-retrieved of the $M$ most similar memories and the rest by similarity. It is motivated by $A$-optimal design (Appendix~\ref{app:design-theory}) and optimises no criterion, and we did not compare it with an $A$-optimal choice. The pre-registration is left verbatim, and a dated amendment records this; the name is the only thing at issue, and no number changes. The code and its artifacts keep the label \texttt{aopt0.4} for this rule.

\subsection{Errata}
\label{app:corr-errata}

Table~\ref{tab:corr-errata} lists the defects we found in our pipeline and in the reporting of its results, with the values they affected. Every value in the main text is the corrected one.

\begin{table}[h]
\centering
\footnotesize
\caption{Defects found and the values they changed. Every value in the
paper is the corrected one; each entry in \texttt{docs/errata.md} gives
the full account.}
\label{tab:corr-errata}
\renewcommand{\arraystretch}{1.15}
\begin{tabular}{@{}>{\raggedright\arraybackslash}p{0.42\linewidth}
                >{\raggedright\arraybackslash}p{0.24\linewidth}
                >{\raggedright\arraybackslash}p{0.26\linewidth}@{}}
\toprule
Defect & Earlier value & Corrected value \\
\midrule
\multicolumn{3}{@{}l}{\emph{Pipeline}} \\
Baseline offset mishandled in eight estimator methods and four callers
  & coverage $0.102$; value recovery at chance
  & coverage $1.000$; recovery restored \\
Three bugs in the merging rule for undetermined totals (Case 1)
  & $60$ of $60$ accepted
  & $17$ and $25$ of $60$ \\
A shadowed tolerance and a ridge-inverted kernel in one validation
  test, and a fixed-size test that divided by zero
  & tests passed without measuring
  & tests measure coverage \\
Paragraph splitter fell back silently
  & NarrativeQA as $11$ memories; InfiniteBench as $4$ episodes
  & $115$ and $14{,}282$ memories; $40$ episodes \\
Audit rows with $\mathrm{rank}\,\mathbf{X} = T$ read as evidence
  & $29$ of $55$ counted
  & reported as automatic \\
Cross-item figure derived from retrieval
  & $94.1\%$
  & withdrawn; replaced by block counts \\
Store-size sweep built on prefixes of one store
  & cost appeared to shrink
  & cost rises with store size \\
Synthetic configuration changed on 2026-08-16
  & retention table unreproducible
  & regenerated from the current configuration \\
Two artifacts overwritten; one path-resolution incident
  & ---
  & regenerated and guarded \\
\midrule
\multicolumn{3}{@{}l}{\emph{Sample size and attribution}} \\
LEDGER's accuracy cost at $300$ questions
  & interval containing zero
  & $-2.6$ $[-4.2, -0.9]$ at $1{,}600$ \\
LoCoMo accuracy change at $120$ questions
  & $+6.7$
  & $+2.7$ at $1{,}542$ \\
Pooling cost not attributed to a reader
  & $16.7$
  & $15.7$, $21.0$, $19.7$ by reader \\
\midrule
\multicolumn{3}{@{}l}{\emph{Reporting, corrected before submission}} \\
Width of the self-normalised set taken from single settings
  & $6.5\times$ and $96\times$
  & $6.6\times$ and $36\times$ (median) \\
LoCoMo merging run attributed to LEDGER's retrieval
  & LEDGER, three conversations
  & score perturbation, two conversations \\
Coverage settings described as designs
  & $36$ designs
  & $36$ settings of $20$ designs \\
Extra readers described as repeating the LEDGER comparison
  & LEDGER against top-$k$
  & cost of pooling only \\
Coverage described with the estimated noise scale
  & $\hat{\sigma}$
  & known $\sigma = 1$ \\
\bottomrule
\end{tabular}
\end{table}

\section{Discussion}
\label{app:discussion}

\subsection{Interpreting the results}
\label{app:interpret}

\paragraph{What the results add to estimability.}
The condition that decides identifiability is classical: a linear
functional $\mathbf{c}^{\!\top}\boldsymbol{\theta}$ is estimable exactly
when $\mathbf{c}$ lies in the row space of the design
matrix~\citep{rao1973linear}.
The paper adds three things.
First, for a retrieval log the design is determined by the retrieved
sets, and its Gram matrix at $d = 1$ consists of the retrieval counts
\eqref{eq:gram1}, so the condition can be checked from counts that
every agent already records.
Second, the structure of these counts under common conventions forces
specific invisible changes.
A fixed retrieval size places $\mathbf{h}_B$ in $\ker\mathbf{X}_d$ for
every $|B| < d$ (Theorem~\ref{thm:fixedk}).
Since $(\mathbf{c}^{(i)})^{\!\top}\mathbf{h}_\emptyset = 1$ for every
memory at $d = 1$, no memory is identifiable at any number of episodes.
For binary outcomes the conclusion is stronger than the absence of an
unbiased linear estimator: the distribution of the log is the same for
two different values of $\varphi_i$, so no estimator is consistent.
Third, the condition fails on every one of the $55$ configurations of
fourteen standard benchmarks (Table~\ref{tab:audit}).
Methods that assign a value to every memory therefore report, on these
logs, numbers that the solver chooses and the data do not constrain.
Slate off-policy evaluation works with the same counts and sets the
undetermined part aside, because it does not change the value of a
policy~\citep{swaminathan2017slate}; for memory valuation, the
undetermined part names the memories that cannot be valued.

\paragraph{Why Shapley values and M\"obius dividends.}
The Shapley value is the unique attribution that satisfies efficiency,
symmetry, null player and additivity~\citep{shapley1953}, and it is the
notion used by attribution-based memory
systems~\citep{memlens2026, clustershapley2025, maxshapley2025}.
Writing $v$ in dividends makes the expected outcome of each episode
linear in $\boldsymbol{\theta}$ and makes each value a linear functional
$(\mathbf{c}^{(i)})^{\!\top}\boldsymbol{\theta}$, which is what turns
valuation into a question about the row space of $\mathbf{X}_d$.
The same test applies to any value that is linear in
$\boldsymbol{\theta}$, with its own coefficient vector.
The Banzhaf value~\citep{banzhaf1964weighted}, for example, is
$\sum_{B \ni i} 2^{1 - |B|}\theta_B$.
At $d = 1$ the choice does not matter: every marginal contribution
$v(S \cup \{i\}) - v(S)$ equals $\theta_{\{i\}}$, so every probabilistic
value~\citep{weber1988probabilistic} of memory $i$ equals
$\theta_{\{i\}}$, and our results at $d = 1$ hold for all of them.

\paragraph{Resolution.}
The experiments on real logs test identifiability at $d = 1$, and this
suffices for the audit.
If $\mathbf{h} \in \ker\mathbf{X}_d$ and
$(\mathbf{c}^{(i)})^{\!\top}\mathbf{h} \neq 0$, extending $\mathbf{h}$
by zeros on the columns of order $d + 1$ gives a vector in
$\ker\mathbf{X}_{d+1}$, and its inner product with the coefficients of
$\varphi_i$ at resolution $d + 1$ is the same nonzero number, because
the two coefficient vectors agree on sets of size at most $d$.
A memory undetermined at resolution $d$ is therefore undetermined at
every higher resolution, and the $55$ configurations determine no value
at any $d$.
The positive results use $d = 1$ for a different reason.
The design $\mathbf{X}_d$ has $p_d = \sum_{\ell = 0}^{d}\binom{n}{\ell}$
columns and rank at most $T$, and the log needs at least $d + 1$
distinct retrieval sizes (Proposition~\ref{prop:sizes}); at $d = 2$ and
$n = 2{,}971$, $p_2 \approx 4.4 \times 10^6$, far above $T = 7{,}200$.
We therefore study $d = 2$ on synthetic logs only
(Section~\ref{sec:exp-estimation}).
What $d = 1$ omits is bounded by Proposition~\ref{prop:trunc}:
$|\varphi_i(v) - \varphi_i(v^{(1)})| \le \frac{1}{2}
\sum_{B \ni i,\, |B| > 1}|\theta_B|$.

\paragraph{Associational and causal values.}
Identifiability depends on $\mathbf{X}_d$ alone, and so do all results of
Section~\ref{sec:identifiability}.
Estimation depends also on the noise.
Writing $v_q = v + \delta_q$, the noise of the linear model is
$\varepsilon_t = \eta_t + \delta_{q_t}(S_t)$, and because retrieval reads
the query, $\delta_{q_t}(S_t)$ can correlate with the design.
The estimates then describe the log: they equal the dividends of $v$
when the log is exogenous (Definition~\ref{def:exogeneity}), for
example when $S_t$ is independent of $q_t$, and they can be biased
otherwise.
Appending query features to the design removes the part of
$\boldsymbol{\delta}$ in their span and leaves every result on
identifiability unchanged, since appending columns never makes a
non-estimable functional estimable (Appendix~\ref{app:assumptions}).
A separate condition concerns adaptivity: Proposition~\ref{prop:interval}
requires that the retrieved sets do not depend on past outcomes.
LEDGER's rule~\eqref{eq:rule} depends on the past only through the
counts $f_i(t)$, which are functions of $\mathbf{X}$, so it satisfies
this condition.

\paragraph{Random retrieval.}
Randomising retrieval also breaks the nesting of top-$k$, and
at high enough rates $\varepsilon$-greedy and score perturbation
determine more values than LEDGER (Figure~\ref{fig:results}a).
They draw replacements from the whole store, so the reader loses
evidence: reaching a fraction of $0.98$ or more costs them between
$13.3$ and $37.0$ accuracy points.
LEDGER draws its replacements from the $M$ memories most similar to the
query and keeps the most similar ones in $H_t$, which bounds how much
evidence it can displace.

\paragraph{The accuracy cost.}
LEDGER's effect on accuracy ranges from $-2.6$ points on HotpotQA to
$+2.7$ on LoCoMo and falls as the gold recall of top-$k$ rises
(Figure~\ref{fig:results}b).
Its mechanism is displacement: the loss comes from questions where
top-$k$ had already retrieved every gold memory
(Figure~\ref{fig:results}c).
Where top-$k$ misses evidence, spreading retrieval costs nothing
measurable, and an oracle that applies LEDGER only there keeps $64\%$ of
the gain with no measurable loss.
The pre-registered query-time substitute, top-$1$ similarity, did not
reproduce this result (Appendix~\ref{app:results}).

\subsection{Scope and open problems}
\label{app:scope}

\paragraph{Scope.}
The analysis uses three properties of the setting: in each episode a
system chooses a set of items from a pool, it observes one scalar
outcome, and the expected outcome is a sum of dividends of order at most
$d$.
It does not use the fact that the items are memories.
The same test therefore applies to the tools an agent calls in an
episode, the passages a retrieval-augmented generator reads, and the
slates a recommender shows~\citep{swaminathan2017slate}, and each of
these settings has its own versions of the fixed size, the private pool
and the tied items.

\paragraph{A query-time gate.}
The oracle of Section~\ref{sec:exp-instrument} shows that most of the
gain is available without cost when LEDGER is applied only to questions
where top-$k$ misses evidence.
A deployed agent has no gold labels, and the first substitute we
pre-registered failed.
A gate that predicts, at query time, whether top-$k$ has found the
evidence would make this trade available in deployment.

\paragraph{An exact optimal-design rule.}
LEDGER's rule~\eqref{eq:rule} targets the diagonal of $\mathbf{V}^{-1}$
through retrieval counts, using $(\mathbf{V}^{-1})_{ii} \ge 1/V_{ii}$.
The $A$-optimal choice~\citep{pukelsheim2006design} among sets drawn from
$R_t$ would maximise the reduction in $\mathrm{tr}(\mathbf{V}^{-1})$,
which by the Sherman--Morrison identity is
$\|\mathbf{V}^{-1}\mathbf{x}_S\|^2/(1 + \mathbf{x}_S^{\!\top}
\mathbf{V}^{-1}\mathbf{x}_S)$ for a set with row $\mathbf{x}_S$.
A greedy search over $R_t$ costs $O(M k_t p_d)$ per episode given
$\mathbf{V}^{-1}$, and a rank-one update of $\mathbf{V}^{-1}$ costs
$O(p_d^2)$.
Whether it determines more values than the count-based rule, and at what
cost in accuracy, is open.

\paragraph{Higher resolution.}
At $d \ge 2$, the number of dividends grows as $n^d$, and most values
need $T$ of the same order to be determined.
If only a few dividends of order two are nonzero, identifiability becomes
a question of sparse recovery, whose conditions on $\mathbf{X}_d$ differ
from the row-space condition used here.

\paragraph{Other outcome models.}
For a binary outcome with an invertible link $g$,
$\mathbb{E}[r_t] = g^{-1}\bigl((\mathbf{X}_d\boldsymbol{\theta})_t\bigr)$,
the distribution of the log determines $\mathbf{X}_d\boldsymbol{\theta}$,
so the row-space condition and every result of
Section~\ref{sec:identifiability} carry over unchanged.
The dividends then describe the value on the scale of the link, and the
estimator and its intervals would need to change.

\paragraph{Adaptive and changing stores.}
Rules that choose sets from past outcomes, such as bandit algorithms,
violate the condition of Proposition~\ref{prop:interval}.
Intervals that remain valid under such rules exist, such as the
confidence set of \citet{abbasi2011oful}, but they are $6.6$ to $36$
times as wide in our synthetic study.
A store that grows over time adds columns that start at zero: a new
memory is undetermined until it has been retrieved with varied partners
and at varied sizes, which gives a natural schedule for when its value
can first be reported.

\subsection{Applications}
\label{app:applications}

\paragraph{Attribution in language models.}
Attribution asks which parts of a model's input or training data an
output depends on.
Data attribution values training examples by their effect on a trained
model~\citep{ghorbani2019datashapley}, context attribution values the sources in
a prompt by their effect on a generation~\citep{cohen2024contextcite},
and Shapley-based methods value the passages of a retrieval-augmented
generator~\citep{maxshapley2025, clustershapley2025}.
These methods obtain their evidence by design: they ablate random subsets
of the sources and fit a surrogate to the outcomes.
A deployed system cannot run such ablations on its own traffic, and when
the same estimators are fitted to logged data, they return a number for
every source whether or not the log determines it.
Our results supply the missing check.
Whether a log determines an attribution depends only on the log's
counts, so it can be decided before any attribution is estimated, and it
applies to every attribution that is linear in the dividends
(Appendix~\ref{app:interpret}).
The results also explain why designed ablations succeed: independent
random subsets vary in size and are not nested, so with enough of them
$\mathbf{X}_1$ has full column rank, while the logs that retrieval
produces under a fixed size have a nonzero kernel at every $T$
(Theorem~\ref{thm:fixedk}).

\paragraph{Cognitive models of memory.}
The rational analysis of human memory holds that the availability of a
memory tracks the probability that it will be needed, which the
environment reveals through the frequency and recency of past
use~\citep{anderson1991reflections}.
Agent memory systems adopted these signals directly: they score memories
by recency, importance and relevance~\citep{park2023generative}, decay
them along a forgetting curve~\citep{zhong2024memorybank}, or rank them
by access frequency~\citep{fademem2026, forgetting2026}.
Frequency and recency measure how often a memory is needed.
They do not measure how much a memory contributes to the outcome when it
is retrieved, and a store managed for an agent needs both.
Our results separate the two quantities within one object.
The retrieval frequencies are the diagonal of
$\mathbf{X}^{\!\top}\mathbf{X}$~\eqref{eq:identity}, so they are exactly
what a log records about need, while the contribution $\varphi_i$ is a
functional of the outcomes.
The same counts decide whether the contribution can be learned: a memory
retrieved in every episode of a top-$k$ log has the largest frequency and
an undetermined value (Theorem~\ref{thm:fixedk}).
A model of agent memory built on need alone therefore cannot tell a
frequently retrieved memory that helps from one that is only retrieved
often, and LEDGER supplies the missing quantity where the log determines
it.

\paragraph{Human control of personal memory.}
Assistants with persistent memory store facts about their users across
conversations.
Users value the personalization this brings, and they ask for
transparent control over what is stored and for accurate
memories~\citep{chen2026relational}.
Control requires knowing which memories matter.
Research on human--AI decision-making shows that people should rely on
a system's advice only when it is correct~\citep{schemmer2023appropriate},
and that over-reliance on incorrect advice resists simple remedies such
as explanations~\citep{buccinca2021trust}.
A user deciding whether to delete, correct or keep a memory relies on the
system's account of that memory, and current systems show what is stored
without any statement of what each memory does.
LEDGER gives each memory one of two answers that a user can rely on.
A determined value comes with a confidence interval, so its uncertainty
is stated.
An undetermined value is reported as such, so the system makes no claim
that its log cannot support.
Merging reduces the store without losing a determined value
(Theorem~\ref{thm:quotient}), which makes it safe to present fewer,
larger units to a user.
The same statements identify which memories an agent relies on, which
supports profiling as well as control, and they should be governed by
the policy that governs the store.

\paragraph{Training memory agents with reinforcement learning.}
Reinforcement learning from outcome rewards underlies recent reasoning
models~\citep{shao2024deepseekmath, guo2025deepseek}, and it now trains
agents to manage memory: which entries to add, update or delete, and
which to use in an answer~\citep{yan2026memory}.
Related methods learn a value for each memory from downstream
reward~\citep{rmm2025, memrl2026, romerl2026}.
Every such method faces a credit-assignment problem: the reward belongs
to the episode, and the value of each memory must be inferred from which
memories the episode retrieved.
This is the inverse problem of this paper, with the rollouts as the log
and the reward as the outcome.
When rollouts retrieve a fixed number of memories, the rewards leave
every memory's credit undetermined (Theorem~\ref{thm:fixedk}), so a
learned per-memory value is fixed by the learner and not by the rewards;
co-retrieved memories then receive spurious
credit~\citep{romerl2026, memoryr2_2026}.
Our test applies to the rollout log before training, because it reads
only which memories each rollout retrieved.
LEDGER's retrieval makes rollouts that determine more values, and its
determined values, with their intervals, can serve as per-memory rewards
that no undetermined memory receives.
Rollout policies that adapt to past rewards violate the condition of
Proposition~\ref{prop:interval}, so their intervals need the wider
self-normalized bounds of \citet{abbasi2011oful}; identifiability is
unaffected, since it depends on $\mathbf{X}_d$ alone.

\paragraph{Evaluating memory systems.}
Memory benchmarks have moved from single questions to long
conversations and multi-session interactions, and from answer accuracy
toward the operations a memory system performs over
time~\citep{locomo2024, longmemeval2025, hu2026evaluating}.
They still score a memory system by the answers it produces.
None of them can score its valuation, because none of them provides
ground truth for the value of a memory, and their own protocols leave
every value undetermined (Table~\ref{tab:audit}).
Our results provide ground truth of two kinds.
Whether a value is determined is a theorem about the log, so a
benchmark can report the fraction of values its logs determine
alongside accuracy, and a valuation method can be tested on whether it
abstains exactly on the undetermined memories.
Where a value is determined, synthetic logs with known dividends test
its interval, as in Section~\ref{sec:exp-estimation}.
The changes a benchmark needs are small and are all changes to its
protocol: a shared store, several retrieval sizes and several passes
per question.

\subsection{Limitations}
\label{app:limitations}

\paragraph{Model.}
The results assume that the query-averaged value has no dividends above
order $d$ (Assumption~\ref{ass:additive}), and the real-data experiments
use $d = 1$.
Omitted dividends bias the estimates wherever they correlate with the
design, and Proposition~\ref{prop:trunc} bounds only their direct effect
on each value.
The noise is assumed conditionally sub-Gaussian
(Assumption~\ref{ass:outcome}).
The reported values are associational unless the log is exogenous
(Appendix~\ref{app:interpret}).

\paragraph{Estimation.}
Coverage of the intervals is measured on synthetic logs drawn from the
model, where the true values are known; on real logs it cannot be
measured.
The result that coverage holds on every reported value at $d = 2$ comes
from a post hoc analysis (Appendix~\ref{app:corrections}).
The interval in~\eqref{eq:estimate} omits the ridge-bias term of
Proposition~\ref{prop:interval}, which is small for $\lambda = 10^{-2}$
on the values LEDGER reports and is the cause of undercoverage on
values it does not report.

\paragraph{Retrieval.}
LEDGER's rule is a count-based heuristic motivated by $A$-optimal design;
it optimises no criterion.
Its sets fall below the drawn size in $6.7\%$ of episodes on pooled
HotpotQA (Appendix~\ref{app:setup}).
The rule treats the $M = 20$ memories most similar to the query as
candidates of comparable relevance, and we did not test this assumption
directly.
Its effect on accuracy depends on the corpus, from $-2.6$ to $+2.7$
points, and the only rule that removes the cost is an oracle that needs
gold labels.

\paragraph{Experiments.}
The accuracy experiments use hosted readers that cannot be pinned and a
single judge model, so they are reproducible in design and not bit for
bit.
The pooled stores for HotpotQA and LongMemEval are our construction; the
benchmarks define private stores.
The audit runs each benchmark at one pass.
The LoCoMo merging result uses score perturbation, not LEDGER's rule, on
three conversations, with fractions estimated on $300$ sampled columns.
The random rules' accuracy is measured on $300$ questions, and LEDGER's
on $1{,}600$.

\section{Notation and Terms}
\label{app:notation}

Vectors are bold lowercase and matrices bold uppercase.
We write $[n] = \{1, \dots, n\}$, $\mathbf{1}_T$ for the all-ones vector
in $\mathbb{R}^T$, and $\mathbf{I}$ for the identity.
For a matrix $\mathbf{A}$, $\mathrm{range}\,\mathbf{A}$ is its column
space, $\mathrm{range}\,\mathbf{A}^{\!\top}$ its row space and
$\ker\mathbf{A}$ its kernel.
Memory $m_i$ is identified with its index $i$, so
$S_t \subseteq [n]$.
Table~\ref{tab:notation} lists the symbols, and Table~\ref{tab:terms}
defines the terms.

\begin{table}[h]
\centering
\small
\caption{Symbols, in the order they appear.}
\label{tab:notation}
\renewcommand{\arraystretch}{1.12}
\begin{tabular}{@{}p{0.25\linewidth}p{0.69\linewidth}@{}}
\toprule
\multicolumn{2}{@{}l}{\emph{Store, episodes and retrieval}} \\
$n$ & number of memories in the store \\
$m_i$ & memory $i$, for $i \in [n]$ \\
$T$ & number of episodes \\
$q_t$,\; $\mathcal{D}_Q$ & query of episode $t$; the distribution it is
  drawn from \\
$S_t$ & set of memories retrieved in episode $t$ \\
$k_t = |S_t|$ & retrieval size of episode $t$; written $k$ when every
  episode retrieves the same number \\
$r_t$,\; $\mathbf{r}$ & outcome of episode $t$; the vector of outcomes \\
$\mathcal{F}_{t-1}$ & history before episode $t$ \\
\midrule
\multicolumn{2}{@{}l}{\emph{Model}} \\
$v_q$ & expected outcome for query $q$ as a function of the retrieved
  set (Assumption~\ref{ass:outcome}) \\
$v$ & query-averaged value, $v(S) = \mathbb{E}_{q}[v_q(S)]$ \\
$\eta_t$ & $r_t - v_{q_t}(S_t)$, conditionally $\sigma$-sub-Gaussian \\
$\varepsilon_t$ & $r_t - v(S_t)$, the noise of the linear
  model~\eqref{eq:model} \\
$\sigma$ & sub-Gaussian noise scale \\
$\theta_B$,\; $\boldsymbol{\theta}$ & M\"obius dividend of the set $B$;
  the vector of dividends with $|B| \le d$ \\
$\theta_\emptyset$ & baseline: the expected outcome with nothing
  retrieved \\
$d$ & resolution: $\theta_B = 0$ for $|B| > d$
  (Assumption~\ref{ass:additive}) \\
$\varphi_i$ & value of memory $i$: its Shapley value under $v$,
  $\sum_{B \ni i}\theta_B/|B|$ \\
$\mathbf{c}^{(i)}$ & coefficients with
  $\varphi_i = (\mathbf{c}^{(i)})^{\!\top}\boldsymbol{\theta}$:
  $c^{(i)}_B = \mathbf{1}[i \in B]/|B|$ \\
\midrule
\multicolumn{2}{@{}l}{\emph{Log and design}} \\
$\mathbf{X}$ & $T \times n$ binary matrix,
  $\mathbf{X}_{ti} = \mathbf{1}[i \in S_t]$ \\
$\mathbf{X}_d$ & design at resolution $d$,
  $(\mathbf{X}_d)_{tB} = \mathbf{1}[B \subseteq S_t]$ for $|B| \le d$ \\
$\mathbf{X}_1$ & $[\mathbf{1}_T \mid \mathbf{X}]$, the design at
  $d = 1$ \\
$\mathrm{freq}(m_i)$ & $(\mathbf{X}^{\!\top}\mathbf{X})_{ii}$, the
  number of episodes that retrieve $m_i$ \\
$\mathrm{co}(m_i, m_j)$ & $(\mathbf{X}^{\!\top}\mathbf{X})_{ij}$, the
  number of episodes that retrieve both \\
$\mathbf{f}$ & vector of retrieval frequencies,
  $f_i = \mathrm{freq}(m_i)$ \\
$\mathbf{h}$ & an invisible change, $\mathbf{h} \in \ker\mathbf{X}_d$ \\
$\mathbf{e}_B$ & unit vector for $\theta_B$ \\
$\mathbf{h}_B$ & invisible change of a top-$k$ log,
  Eq.~\eqref{eq:hB} \\
$s$ & scalar multiple of an invisible change \\
$\mathbf{w}$ & least-squares solution of
  $\mathbf{X}\mathbf{w} = \mathbf{1}_T$ \\
$n_b$,\; $T_b$ & memories and episodes of block $b$ \\
\midrule
\multicolumn{2}{@{}l}{\emph{Estimation}} \\
$\lambda$ & ridge parameter, $10^{-2}$ \\
$\mathbf{V}$ & $\mathbf{X}_d^{\!\top}\mathbf{X}_d + \lambda\mathbf{I}$ \\
$\hat{\boldsymbol{\theta}}$,\; $\hat{\varphi}_i$ & ridge estimate of the
  dividends; estimate of $\varphi_i$, Eq.~\eqref{eq:estimate} \\
$\hat{\sigma}$ & residual standard deviation \\
$\delta$ & error level of the intervals; $0.05$ in
  Section~\ref{sec:experiments} \\
$z_{1-\delta/(2n)}$ & Gaussian quantile of the Bonferroni interval \\
$\mathbf{U}$ & orthonormal basis of $\ker\mathbf{X}_d$ \\
$\tau$ & tolerance of the identifiability test, $10^{-8}$ \\
\midrule
\multicolumn{2}{@{}l}{\emph{Merging}} \\
$g$,\; $g(i)$ & a group of memories; the group of memory $i$ \\
$\bar{\mathbf{X}}$,\; $\bar{\mathbf{X}}_1$ & merged matrix, one column
  per group; with the baseline column \\
$\bar{\theta}_g$ & value of group $g$,
  $\sum_{i \in g}\theta_{\{i\}}$ \\
$s_g(\boldsymbol{\theta})$ & $\sum_{i \in g}\theta_{\{i\}}$ for an
  arbitrary grouping (Theorem~\ref{thm:approx}) \\
$D_g$ & largest number of episodes in which group $g$ was retrieved while
  one of its members was not \\
$\sigma^{+}_{\min}$ & smallest nonzero singular value \\
\midrule
\multicolumn{2}{@{}l}{\emph{Retrieval}} \\
$R_t$ & the $M$ memories most similar to $q_t$ \\
$M$ & candidate window, $20$ \\
$f_i(t)$ & number of episodes before $t$ that retrieved memory $i$ \\
$L_t$,\; $H_t$ & least-retrieved and most similar parts of $S_t$,
  Eq.~\eqref{eq:rule} \\
$\rho$ & share of $S_t$ filled by $L_t$, $0.4$ \\
\bottomrule
\end{tabular}
\end{table}

\begin{table}[h]
\centering
\small
\caption{Terms, in the order they appear.}
\label{tab:terms}
\renewcommand{\arraystretch}{1.12}
\begin{tabular}{@{}p{0.25\linewidth}p{0.69\linewidth}@{}}
\toprule
Episode & one query, one retrieval, one answer and its outcome \\
Log & the retrieved sets and outcomes of all episodes, stored as
  $\mathbf{X}$ and $\mathbf{r}$ \\
Determined, identifiable & the log determines $\varphi_i$ exactly when
  $\mathbf{c}^{(i)}$ lies in the row space of $\mathbf{X}_d$; the two
  words are used interchangeably \\
Undetermined & not determined; LEDGER reports no estimate for such a
  memory \\
Invisible change & a vector $\mathbf{h} \in \ker\mathbf{X}_d$:
  $\boldsymbol{\theta}$ and $\boldsymbol{\theta} + s\mathbf{h}$ predict
  the same outcome in every episode \\
Convention & a habit of retrieval or evaluation that fixes the counts of
  the log, such as a fixed size or a private store \\
Top-$k$ log & a log in which every episode retrieves exactly $k$
  memories \\
Nested sets & sets retrieved for one query at several sizes, each
  containing the smaller ones, as under deterministic top-$k$ \\
Private store & each question, document or conversation has its own
  memories \\
Shared store & one store used by all queries \\
Block & a group of memories never retrieved together with a memory
  outside it, with the episodes that retrieve them \\
Tied memories & memories with identical columns of $\mathbf{X}$, that
  is, retrieved in the same episodes \\
Merging & replacing each group of tied memories by one memory whose value
  is the group total \\
Unit & a memory or a group after merging; fractions after merging are
  over units \\
Fraction determined & share of memories, or units after merging, whose
  value the log determines \\
Pass & one round in which every question of the experiment is asked
  once \\
Configuration & one benchmark run at one retrieval size $k$ in the
  audit \\
Gold memory & a memory that holds evidence for the answer \\
Gold recall & share of gold memories that top-$k$ retrieves \\
Oracle & LEDGER applied only to questions where top-$k$ misses a gold
  memory, which requires gold labels \\
Joint coverage & share of synthetic logs in which every reported
  interval contains its true value \\
Setting & one choice of $n$, $k$, $T$ and $\lambda$ in the synthetic
  study, with $20$ random designs \\
\bottomrule
\end{tabular}
\end{table}

\end{document}